\pdfoutput=1

\documentclass[11pt]{article}

\usepackage[preprint]{acl}

\usepackage{times}
\usepackage{latexsym}
\usepackage{bbm}
\usepackage{todonotes}
\usepackage{footnote}
\makesavenoteenv{figure}
\usepackage[T1]{fontenc}
\usepackage[utf8]{inputenc}
\usepackage{microtype}
\usepackage{inconsolata}
\usepackage{graphicx}
\usepackage{xspace}
\usepackage{hyperref}       
\usepackage{url}            
\usepackage{booktabs}       
\usepackage{amsfonts}       
\usepackage{nicefrac}       
\usepackage{xcolor}         
\usepackage{multirow}
\usepackage{bbding}
\usepackage{arydshln}
\usepackage{amsmath}
\usepackage{csquotes}
\usepackage{colortbl}
\usepackage[most]{tcolorbox}
\usepackage{listings}
\usepackage[dvipsnames]{xcolor}
\usepackage{adjustbox}
\usepackage{enumitem}
\usepackage{tabularx}
\usepackage{mathtools}
\usepackage{fontawesome5}
\usepackage{placeins}

\usepackage{algorithm,algpseudocode}
\usepackage{caption}

\newcommand{\ours}{{MIC}}
\newcommand{\benchmark}{{MIC-Bench}}

\newcommand{\rh}[1]{\textcolor{blue}{#1}}

\title{MIC: Explaining Image--Claim Inconsistencies in AI-Generated Multimodal Misinformation}

\author{
  \textbf{Ruihong Zeng}\textsuperscript{\textnormal{1}}\thanks{%
  Work done during an internship at UKP Lab.},
  \textbf{Jonathan Tonglet}\textsuperscript{\textnormal{2,3,4}},
  \textbf{Preslav Nakov}\textsuperscript{\textnormal{1}},
  \textbf{Iryna Gurevych}\textsuperscript{\textnormal{1,2}} \\
  \textsuperscript{1}Mohamed bin Zayed University of Artificial Intelligence, UAE \\
  \textsuperscript{2}Ubiquitous Knowledge Processing Lab (UKP Lab), Department of Computer Science, \\
  TU Darmstadt and National Research Center for Applied Cybersecurity ATHENE, Germany \\
  \textsuperscript{3}Department of Electrical Engineering,
  KU Leuven, Belgium \\
  \textsuperscript{4}Department of Computer Science,
  KU Leuven, Belgium
}

\begin{document}
\maketitle

\begin{abstract}
Claims paired with AI-generated images are a rapidly growing form of misinformation.
Existing automated fact-checking (AFC) methods mainly treat this as a provenance problem, detecting low-level synthesis artifacts to decide whether an image is AI-generated.
However, such methods do not verify what human fact-checkers often check: whether an image's content is consistent with the context implied by its accompanying claim.
To address this gap, we introduce \ours{}~(\textbf{M}ultimodal \textbf{I}nconsistency \textbf{C}hecking), an AFC framework that assists human fact-checkers by detecting AI-generated multimodal misinformation and explaining inconsistencies using world knowledge. 
\ours{} first uses supervised fine-tuning (SFT) for task adaptation and then applies Group Relative Policy Optimization (GRPO) to directly optimize component-level verifiable rewards for verdict prediction, inconsistency type classification, visual evidence description, and world-knowledge explanation.
We further introduce \benchmark{}, a benchmark comprising 8,812 image--claim instances derived from 4,406 claims, where each claim is paired with an authentic image and an AI-generated counterpart that introduces a controlled contextual inconsistency.
Compared with SFT alone, GRPO further improves Macro-F1 by 4.67 and 4.11 points in the in-distribution and out-of-distribution settings, respectively, while also improving the semantic similarity of visual evidence descriptions and world-knowledge explanations to reference annotations.
Our code and data are available at \url{https://github.com/UKPLab/arxiv2026-mic}.

\end{abstract}

\section{Introduction}
\label{sec:intro}

\begin{figure}
    \centering
    \includegraphics[width=0.99\linewidth]{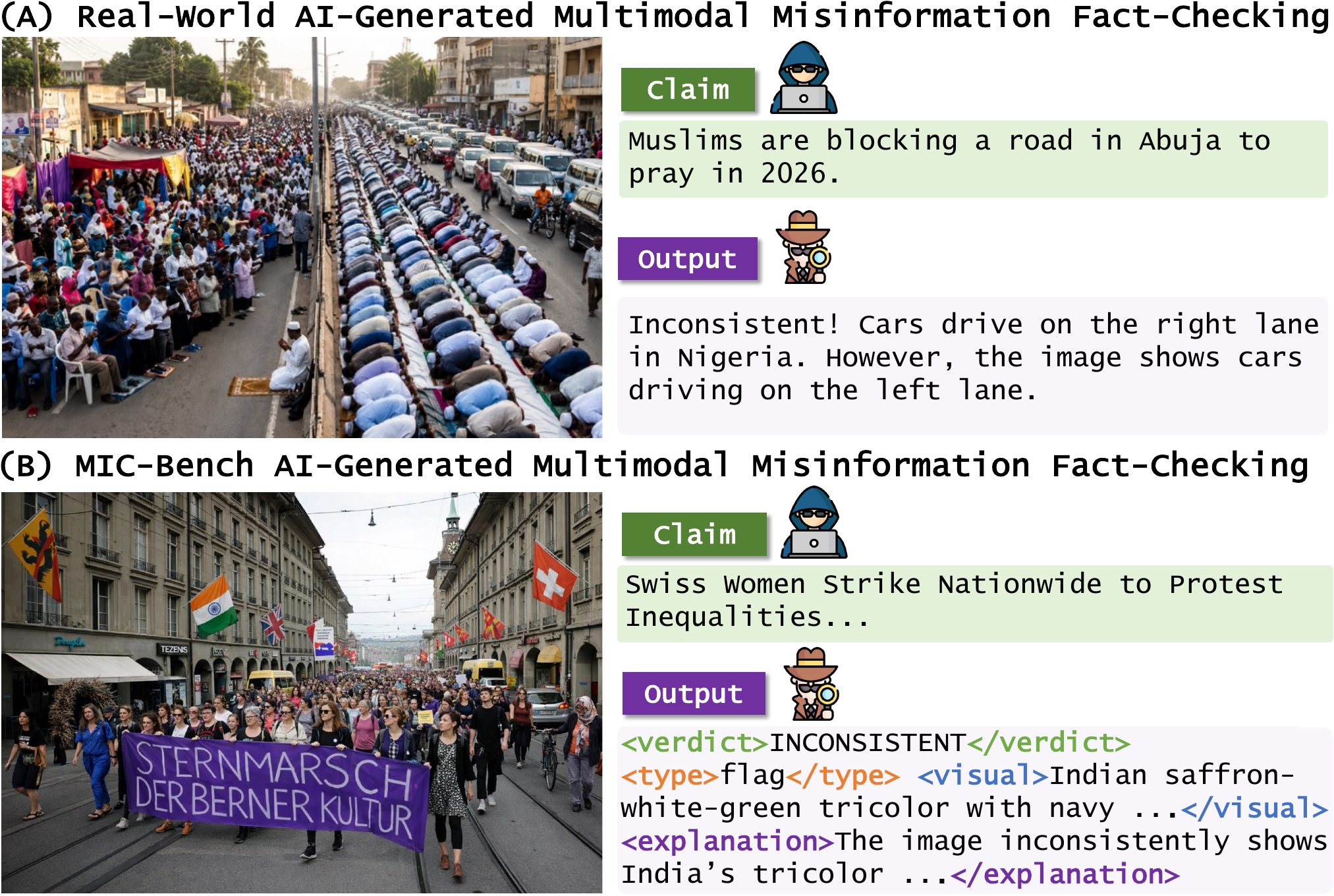}
    \caption{
    (A) A real-world example of AI-generated multimodal misinformation fact-checked by a human expert~\citep{awosoro2026dubawa}.
    (B) A \benchmark{} example of AI-generated multimodal misinformation fact-checked by \ours{} using visual evidence and world knowledge.
    }
    \label{fig:trigger_fig}
\end{figure}

AI-generated images paired with claims are an increasingly pressing form of misinformation~\cite{dufour2024ammeba,drolsbach2025characterizing}. Journalists and fact-checkers have developed practical verification strategies over time, many of which rely on identifying inconsistencies between the image and its claimed context~\citep{gijn2025detectingai}. Figure~\ref{fig:trigger_fig}~(A) illustrates a real-world example in which AI-generated multimodal misinformation is exposed by a human fact-checker using knowledge of Nigerian driving rules~\citep{awosoro2026dubawa}.

While photorealistic images can be rapidly generated and disseminated alongside false claims~\cite{wack2024political,cazzamatta2025ai}, verifying them often requires hours for human fact-checkers~\cite{gijn2025detectingai}. To support this process, prior work has focused on detecting AI-generated images through synthesis artifacts, such as subtle pixel-level irregularities in texture or noise patterns~\cite{karageorgiou2025any,gye2025reducing,xu2025fakeshield}.
Recent methods also explain detection results by describing detected artifacts~\cite{zhou2025aigi,jifakexplain}.
However, these methods face two key limitations. First, synthesis artifacts become increasingly subtle as generative models improve, placing detectors in a continual arms race~\cite{liu2024evolving,xu2025fully,pei2024deepfake}. 
Second, even when these methods detect that an image is AI-generated, they do not identify whether a specific visual element conflicts with the context implied by the accompanying claim~\citep{gijn2025detectingai}.
OOC detection studies authentic images paired with misleading captions~\citep{qi2024sniffer,tonglet-etal-2025-cove}, while AI-generated news detection distinguishes real from generated news~\citep{huang2024miragenews,zhang2025generation}.
Neither directly addresses cases where an AI-generated image broadly depicts the claimed event but contains a localized, context-dependent inconsistency.
Identifying and explaining such inconsistencies requires localizing the relevant visual evidence and applying world knowledge about the claimed context, which remains time-consuming for human fact-checkers.

To address this gap, we study \emph{claim-grounded visual inconsistency detection} for AI-generated multimodal misinformation.
Given an image--claim pair, the model predicts whether the pair is consistent.
When an inconsistency is detected, the model also identifies its type, describes the relevant visual evidence, and uses world knowledge to explain why this evidence conflicts with the claim.

To tackle this challenge, we propose \ours{}, a multimodal large language model~(MLLM) for structured and explainable verification of AI-generated multimodal misinformation. Rather than producing an unconstrained binary verdict or free-form rationale, \ours{} decomposes verification into four structured outputs: (\emph{i})~a consistency verdict, (\emph{ii})~an inconsistency type, (\emph{iii})~a visual evidence description, and (\emph{iv})~a world-knowledge explanation. To align generation with these objectives, \ours{} adopts a two-stage training framework. 
First, SFT teaches the model to produce structured reasoning traces that identify the relevant visual evidence and explain how it supports or contradicts the claim context.
Next, GRPO refines the model using verifiable rewards over the four task components, jointly improving verdict accuracy, inconsistency type prediction, visual evidence description, and knowledge-based explanation generation. This formulation transforms explanation generation from unconstrained free-form reasoning into a set of grounded and verifiable objectives.

To support training and evaluation, we introduce \benchmark{}, a benchmark of 8,812 image--claim instances derived from 4,406 real news image--claim pairs, where each claim is paired with one authentic image and one AI-generated counterpart containing a single, visually plausible inconsistency drawn from one of nine inconsistency types rooted in the practices and accumulated experience of journalists and fact-checkers~\citep{gijn2025detectingai}.
In addition to binary verdicts, \benchmark{} provides structured annotations for inconsistency types, visual evidence descriptions, and world-knowledge explanations, enabling fine-grained evaluation of both prediction accuracy and explanation quality. 
Our experiments against eleven strong baselines show that \ours{} substantially improves Macro-F1 while generating explanations that more closely match the reference annotations across both in-distribution and out-of-distribution settings, and outperforms all baselines on images from an unseen generator.

We make the following contributions:

\begin{itemize}[itemsep=2pt, topsep=2pt]
    \item We formulate \emph{claim-grounded visual inconsistency detection} as a specific setting of image--claim consistency verification, focusing on AI-generated images that broadly depict the claimed event but contain a localized visual detail conflicting with the geographic, cultural, or temporal context implied by the claim.
    \item We introduce \benchmark{}, a benchmark of 8,812 image--claim instances, where each claim is paired with an authentic image and an AI-generated counterpart containing a single controlled inconsistency, annotated with its type, a visual evidence description, and a world-knowledge explanation.
    \item We propose \ours{}, which uses GRPO to directly optimize component-level verifiable rewards for verdict prediction, inconsistency type classification, visual evidence description, and world-knowledge explanation.
    
\end{itemize}

\section{Related Work}
\label{sec:related}

\begin{figure*}[!t]
    \centering
    \includegraphics[width=0.99\linewidth]{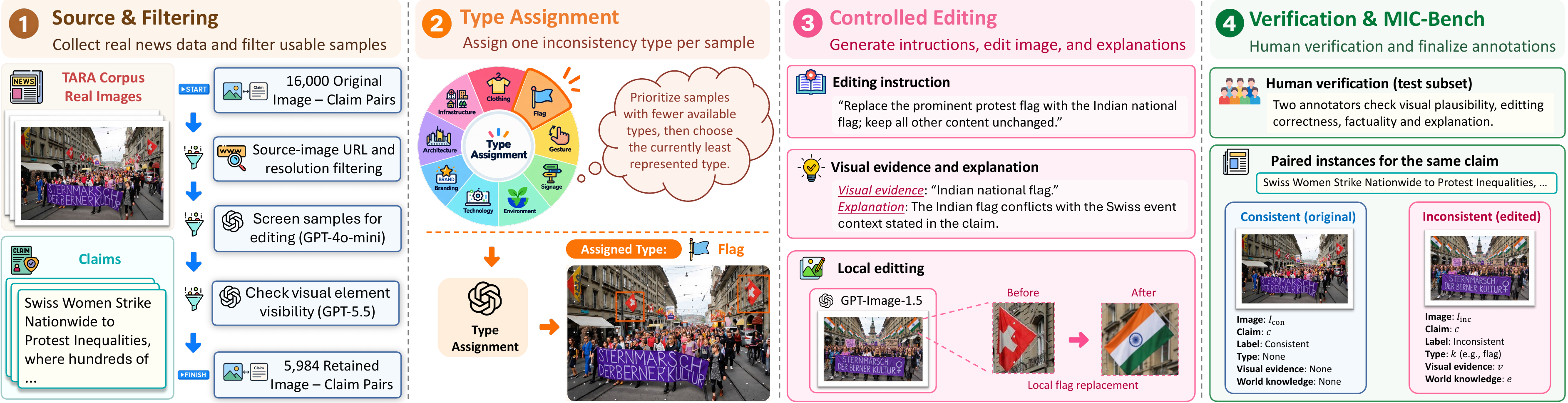}
    \caption{Overview of our \benchmark{} construction pipeline.}
    \label{fig:dataset_construction}
\end{figure*}

\paragraph{AI-Generated Image Detection.}
Detecting whether an image is AI-generated has been extensively studied as generative models evolved from GANs to diffusion-based systems~\cite{xu2025fully,pei2024deepfake}. Early approaches relied on CNN-based image-domain cues and frequency-domain artifacts~\cite{wang2020cnn,frank2020leveraging}, while more recent methods incorporate CLIP-based representations~\cite{tan2025c2p,keita2025deeclip}, commonsense reasoning over physically implausible content~\cite{tan2025semantic,yang2025heie}, and natural language rationales to explain the detection decisions~\cite{zhou2025aigi,jifakexplain}. Other work has focused on provenance signals and content authentication rather than passive detection~\cite{rosenthol2022c2pa,gowal2025synthid}. Collectively, these approaches provide valuable signals for human fact-checking. However, most of them primarily address whether an image is AI-generated, rather than whether the depicted visual content is consistent with the context implied by the claim.

\paragraph{Multimodal AFC.}
AFC aims to support human fact-checkers by automating parts of their workflow~\cite{nakov2021automated,guo2022survey, akhtar-etal-2023-multimodal, geng2025m4fc}.
Within multimodal AFC, prior work has studied out-of-context (OOC) misinformation, where authentic images are paired with claims that misrepresent their event, location, or time~\citep{qi2024sniffer,10944123,tonglet-etal-2025-cove}.
For AI-generated news, MiRAGeNews generates fictional captions and corresponding images, framing detection as distinguishing real from generated image--caption pairs~\citep{huang2024miragenews}.
For explainable verification, Fact-R1~\cite{DBLP:conf/nips/ZhangLZCSLYLZ25} studies video misinformation detection and its FakeVV dataset constructs out-of-context misinformation by replacing entities in news titles while retaining the original videos.
In \benchmark{}, we instead keep each real-event claim fixed and pair it with an authentic source image and an AI-generated counterpart containing a single localized visual inconsistency.
Our setting targets visual details that conflict with the geographic, cultural, or temporal context implied by the claim, while the image broadly depicts the claimed event.

\section{The \benchmark{} Dataset}
\label{sec:dataset_construction}

While AI-generated multimodal misinformation is growing rapidly, real-world fact-checked instances remain too limited and sparsely annotated to support supervised training~\cite{geng2025m4fc}.
Existing image--claim misinformation benchmarks do not jointly provide an authentic image and a controlled edited counterpart paired with the same real-event claim, together with fine-grained annotations of the inconsistency type, visual evidence, and world-knowledge explanation.
We therefore construct \benchmark{}, a synthetic dataset based on real news image--caption pairs from TARA~\cite{fu2022there}.
We treat each caption as a claim $c$ and its image as the consistent source image $I_{\text{con}}$.
Appendix~\ref{appendix:benchmark_comparison} provides a detailed comparison with related benchmarks.

\subsection{Inconsistency Taxonomy}
\label{subsec:source_data_and_taxonomy}

Prior work emphasizes designing AFC from the perspective of professional fact-checkers~\citep{nakov2021automated}.
Accordingly, to define the scope of inconsistencies represented in \benchmark{}, we introduce a taxonomy of nine types grounded in journalism fact-checking guidelines~\citep{gijn2025detectingai}.
Each selected type involves a visual element that can be localized and modified through a controlled edit, but judging its consistency with the claim requires relevant world knowledge about the event and its geographic, cultural, or temporal context.
As summarized in Table~\ref{tab:taxonomy}, the taxonomy includes clothing, flag, gesture, signage, architecture, infrastructure, technology, branding, and environment, with representative examples provided in Appendix~\ref{subsec:type_examples}. 

\subsection{Dataset Construction}
\label{subsec:dataset_construction}
As shown in Figure~\ref{fig:dataset_construction}, our pipeline introduces one controlled inconsistency into each selected source image.
We first use GPT-4o-mini~\cite{openai2024gpt4omini} to select image–claim pairs with visible elements that can be edited to create an inconsistency with the claim’s context. We then use GPT-5.5 to check whether the visual elements associated with each of the nine inconsistency types are clearly visible, without inferring their presence from the claim.
We then assign one target type per retained pair using a constrained-first and rarest-first strategy to balance taxonomy coverage.
For the selected type, GPT-5.5~\cite{openai2026gpt55} generates a detailed edit instruction, an initial visual evidence description $v_0$, and an initial world-knowledge explanation $e_0$.
GPT-Image-1.5~\cite{openai2025gptimage15} then generates the inconsistent counterpart $I_{\text{inc}}$ by modifying only the target visual element while preserving the rest of the image.
This design controls \emph{what} is edited, \emph{which inconsistency type} is introduced, and \emph{why} it contradicts the claimed context.

\begin{table}[!t]
\centering
\resizebox{\columnwidth}{!}{%
\begin{tabular}{@{}l p{8cm}@{}}
\toprule
\textbf{Inconsistency type} & \textbf{Representative visual elements} \\
\midrule
clothing       & Uniforms, traditional clothing, religious attire \\
\midrule
flag           & National flags, regional flags, protest flags \\
\midrule
gesture        & Cultural greetings, public-speech gestures \\
\midrule
signage        & Storefront signs, government signs, street signs \\
\midrule
architecture   & Religious buildings, civic landmarks, arches, facades \\
\midrule
infrastructure & Road signs, transit signs, traffic signals \\
\midrule
technology     & Consumer electronics, broadcast devices \\
\midrule
branding       & Brand logos, campaign materials, event marks \\
\midrule
environment    & Vegetation, weather, seasonal cues, landscapes \\
\bottomrule
\end{tabular}}
\caption{The nine inconsistency types in \benchmark{}. }
\label{tab:taxonomy}
\end{table}

\subsection{Annotation Schema}
\label{subsec:annotation_schema}

Each claim $c$ is paired with a source image $I_{\text{con}}$ and an edited image $I_{\text{inc}}$.
The inconsistent image is annotated with $(k, v, e)$, where $k$ is an inconsistency type from $\mathcal{K}$, the set of the nine inconsistency types, $v$ describes the manipulated visual element, and $e$ explains why the edit contradicts the claimed context. The consistent image uses \texttt{None} for these fields.
To support structured reasoning supervision, we provide two rationales for each claim: $r_{\text{con}}$ for why $I_{\text{con}}$ is consistent with $c$, and $r_{\text{inc}}$ for why $I_{\text{inc}}$ contradicts $c$.
These rationales are initialized by GPT-5.5 and human-verified to remain grounded in the image and consistent with the final annotations.
For test-set quality control, two annotators determine whether to keep each candidate based on visual realism, edit fidelity, and the factual correctness of the explanation.
They achieved 97.4\% agreement on whether each candidate should be kept or rejected, with Krippendorff's $\alpha=0.81$~\cite{hayes2007answering}.
Retained examples were revised when necessary to ensure that their rationales remained grounded in the image and consistent with $(k, v, e)$.

The final dataset comprises 4,406 claims, each paired with one consistent source image and one AI-generated inconsistent counterpart, yielding 8,812 image--claim instances in total.
We use a temporal split and report two test settings: \emph{in-distribution replacement}, where the visual elements introduced by the edit also appear in the training data, and \emph{out-of-distribution replacement}, where the introduced visual elements are held out from training.
Details are provided in Appendix~\ref{appendix:dataset_details}.

\section{Problem Formulation}
\label{sec:problem_formulation}

Given a news image $I$ and a claim $c$, the goal of \textit{claim-grounded visual inconsistency detection} is to determine whether the image is consistent with the context implied by the claim, yielding a verdict $y \in \{\textsc{consistent}, \textsc{inconsistent}\}$. 
For consistent cases, the inconsistency-specific outputs are set to \texttt{None}.
For inconsistent cases, the model additionally predicts: (\emph{i})~an inconsistency type $k \in \mathcal{K}$ from the taxonomy defined in Section~\ref{subsec:source_data_and_taxonomy}, (\emph{ii})~a description $v$ of the visual element for the inconsistency, and (\emph{iii})~an explanation $e$ that grounds the inconsistency in relevant world knowledge.

\section{Method}
\label{sec:method}

\subsection{Overall Framework}
\label{subsec:overall_framework}

\ours{} trains an MLLM through two stages:
(\emph{i}) \ours{} first applies \textbf{SFT}~\cite{guo2025deepseek} to teach the MLLM the structured output schema and initialize its ability to generate explanations grounded in world knowledge.
(\emph{ii}) The SFT-initialized model is then refined using \textbf{GRPO}~\cite{shao2024deepseekmath}, where rule-based verifiable rewards are used to improve the reliability of inconsistency detection and explanation generation.

To support structured generation and reward computation, \ours{} serializes each model output into five fixed fields: \texttt{<think>}, \texttt{<verdict>}, \texttt{<type>}, \texttt{<visual>}, and \texttt{<explanation>}. The \texttt{<think>} field contains the reasoning trace $r$, i.e., $r_{\text{con}}$ for consistent image--claim pairs and $r_{\text{inc}}$ for inconsistent pairs.
The remaining fields correspond to the structured prediction $(y, k, v, e)$. This separation makes the reasoning process and verifiable outputs independently parseable, enabling per-field supervision during SFT and per-field reward optimization during GRPO.

\subsection{SFT}
\label{subsec:sft_stage}
We perform SFT using the reference reasoning traces and annotations introduced in Section~\ref{subsec:annotation_schema}.
For each training instance, the target sequence contains a reference reasoning trace in the \texttt{<think>} field, using $r_{\text{con}}$ for consistent image--claim pairs and $r_{\text{inc}}$ for inconsistent pairs, followed by the structured targets $(y,k,v,e)$ in the remaining fields. 
For consistent instances, inconsistency-specific fields are filled with \texttt{None}. 
For parameter-efficient adaptation, we use LoRA~\cite{hulora} by inserting low-rank adaptation modules into the linear layers of the vision encoder, the vision--language projector, and the language model, while keeping the pretrained weights frozen.

\subsection{Reward Functions Design}
\label{subsec:reward_function_design}

We use a format reward $R_{\text{fmt}}$ following previous work~\cite{guo2025deepseek} and define four task-specific rewards for verdict prediction ($R_{\text{ver}}$), inconsistency type classification ($R_{\text{type}}$), visual evidence description ($R_{\text{desc}}$), and world-knowledge explanation ($R_{\text{exp}}$).

Given an image--claim pair $(I,c)$, the policy model $\pi_\theta$ generates a structured response $o$, which is parsed by $P(\cdot)$ into predictions $(\hat{y}, \hat{k}, \hat{v}, \hat{e}) = P(o)$. For inconsistent examples, the overall reward is computed as a weighted combination of the five reward components:
\begin{align}
R(I,c,o) =
&\ \lambda_{\text{fmt}} R_{\text{fmt}}(o)
+ \lambda_{\text{ver}} R_{\text{ver}}(o) \nonumber  \\
&+ \lambda_{\text{type}} R_{\text{type}}(o)
+ \lambda_{\text{desc}} R_{\text{desc}}(o) \nonumber   \\
&+ \lambda_{\text{exp}} R_{\text{exp}}(o), 
\end{align}
where $\lambda_{\text{fmt}}$, $\lambda_{\text{ver}}$, $\lambda_{\text{type}}$, $\lambda_{\text{desc}}$, and $\lambda_{\text{exp}}$ denote the corresponding reward weights. For consistent examples, the overall reward includes only $R_{\text{fmt}}$ and $R_{\text{ver}}$, excluding inconsistency-specific rewards from the weighted sum.

\paragraph{Format Reward.}
To ensure reliable reward extraction, we use a binary format reward that assigns a score 1 when the generated response can be parsed into $P(o)$ and contains all required fields, and 0 otherwise.

\paragraph{Accuracy-based Rewards.}
$R_{\text{ver}}$ and $R_{\text{type}}$ are binary accuracy rewards for the predicted verdict $\hat{y}$ and predicted inconsistency type $\hat{k}$, respectively, assigning 1 to correct predictions and 0 otherwise.

\paragraph{Embedding-based Semantic Rewards.}
We define $R_{\text{desc}}$ and $R_{\text{exp}}$ as the cosine similarity between the generated visual evidence description $\hat{v}$ or world-knowledge explanation $\hat{e}$ and ground-truth annotation, computed using Qwen3-Embedding-0.6B~\cite{zhang2025qwen3}.

\subsection{GRPO Optimization}
\label{subsec:grpo_optimization}

After SFT, we further train the model using GRPO~\cite{shao2024deepseekmath}. 
For each image--claim pair $(I,c)$, the model samples a group of $G$ structured responses $\{o_i\}_{i=1}^{G}$. 
Each response is assigned a reward according to the reward function defined in Section~\ref{subsec:reward_function_design}, i.e., $\rho_i = R(I,c,o_i)$.
These response-level rewards can therefore be used directly for policy optimization.

GRPO computes a group-relative advantage score $A_i$ by normalizing the rewards within the sampled response group:
\begin{align}
A_i =
\frac{
\rho_i - \mathrm{mean}(\{\rho_1,\ldots,\rho_G\})
}{
\mathrm{std}(\{\rho_1,\ldots,\rho_G\}) + \epsilon
},
\end{align}
where $\epsilon$ is a small constant introduced for numerical stability. This group-relative normalization favors higher-reward responses within the sampled group without requiring a separate value model.
We further compare GRPO with DPO~\citep{rafailov2023direct} in Appendix~\ref{subsec:training_strategy_comparison} and find that GRPO performs better overall.

\section{Experiments}
\label{sec:exp_setup}

\subsection{Baselines}
\label{subsec:baseline_methods}
We compare \ours{} against various baselines: (i) a generic vision-language encoder, CLIP; (ii) multimodal misinformation detectors, SNIFFER~\citep{qi2024sniffer} and MiRAGe~\cite{huang2024miragenews}; (iii) closed-source VLM, GPT-5.4-mini~\cite{openai2026gpt54mini}; and (iv) open-source VLMs, including Qwen2.5-VL-3B-Instruct~\citep{bai1others}, Qwen3-VL variants~\citep{bai2025qwen3}, LLaVA-OneVision-7B~\citep{li2024llava}, and InternVL3-8B~\citep{zhu2025internvl3}.

\begin{table*}[!t]
  \centering
  \resizebox{0.85\textwidth}{!}{%
  \begin{tabular}{l rrrr rrrr}
  \toprule
  & \multicolumn{4}{c}{\textbf{In-Distribution Replacement}} & \multicolumn{4}{c}{\textbf{Out-of-Distribution Replacement}} \\
    \cmidrule(lr){2-5} \cmidrule(lr){6-9}
    \textbf{Model} & \textbf{Macro-F1} & \textbf{TypeAcc} & \textbf{VisualSim} & \textbf{ExplSim}
    & \textbf{Macro-F1} & \textbf{TypeAcc} & \textbf{VisualSim} & \textbf{ExplSim} \\
  \midrule
    CLIP                  & 39.91 &  -- &  -- &  --   & 40.72 &  -- &  -- &  --   \\
    MiRAGe                           & 53.13 &  -- &  -- &  --   & 51.70 &  -- &  -- &  --   \\
    SNIFFER                          & 46.69 &  -- &  -- & 10.66 & 44.07 &  -- &  -- &  8.49 \\
    GPT-5.4-mini                     & 53.89 & 11.52 & 20.14 & 17.88 & 53.00 & 12.18 & 19.43 & 17.02 \\
    Qwen2.5-VL-3B-Instruct                    & 40.47 & 15.79 & 13.83 & 29.11 & 38.98 & 13.32 & 11.83 & 28.08 \\
    Qwen3-VL-4B-Instruct                      & 58.90 & 19.06 & 23.16 & 25.24 & 59.44 & 19.63 & 24.18 & 27.18 \\
    Qwen3-VL-8B-Instruct                      & 59.95 & 23.33 & 26.74 & 28.55 & 58.14 & 19.77 & 25.13 & 26.88 \\
    Qwen3-VL-30B-A3B                 & 63.96 & 32.01 & 37.95 & 37.42 & 63.14 & 33.81 & 38.45 & 36.61 \\
    LLaVA-OneVision-7B               & 56.01 &  9.25 & 13.97 & 18.65 & 52.06 &  7.59 & 10.43 & 16.45 \\
    InternVL3-8B                     & 55.52 & 13.66 & 18.24 & 22.65 & 55.56 & 14.18 & 17.82 & 22.51 \\
    Qwen3-VL-8B-Thinking             & 59.79 & 40.40 & 43.41 & 46.19 & 59.09 & 37.54 & 42.64 & 46.28 \\
    \midrule
    \ours{}                          & \textbf{94.88} & \textbf{89.76} & \textbf{83.04} & \textbf{82.54} & \textbf{90.26} & \textbf{78.65} &
    \textbf{72.09} & \textbf{75.78} \\
  \bottomrule
  \end{tabular}%
  }
  \caption{Performance (in \%) of \ours{} and baselines on \benchmark{}.
  \textit{In-Distribution Replacement} uses edited-in visual elements seen during training, whereas \textit{Out-of-Distribution Replacement} uses edited-in visual elements held out from training.
  The best performance per dataset per metric is marked in boldface.}
  \label{tab:main_results}
\end{table*}

\subsection{Implementation Details}

We implement \ours{} in PyTorch~\citep{paszke2019pytorch} and conduct all experiments on 8 AMD MI210 GPUs, each with 64\,GB of memory.
During SFT, we adopt Qwen3-VL-4B-Instruct~\citep{bai2025qwen3} as the backbone model and use LoRA adapters with rank $r=64$ and scaling factor $\alpha=128$. The model is trained for 3 epochs using AdamW with a learning rate of $2\times10^{-4}$ and a batch size of 8.
For GRPO training, we initialize the model from the SFT checkpoint and optimize it using a batch size of 8 with 8 sampled rollouts per prompt. More implementation details are provided in Appendix~\ref{app:experimental_setup}.

\subsection{Evaluation Metrics}
\label{subsec:evaluation_metrics}

We report Macro-F1 over \textsc{consistent} and \textsc{inconsistent} verdicts.
For inconsistent examples, \emph{TypeAcc} measures inconsistency-type accuracy, while \emph{VisualSim} and \emph{ExplSim} measure the semantic similarity of generated visual evidence and world-knowledge explanations to their respective references.
Following Section~\ref{subsec:reward_function_design}, both similarity metrics are computed using Qwen3-Embedding-0.6B~\cite{zhang2025qwen3}.
We assess robustness across embedding models in Appendix~\ref{subsec:alt_embedding} and report reference-based \emph{VisualJudge} and \emph{ExplJudge} scores in Appendix~\ref{subsec:llm_judge_evaluation}.

\begin{figure}[!t]
    \centering
    \includegraphics[width=0.99\linewidth]{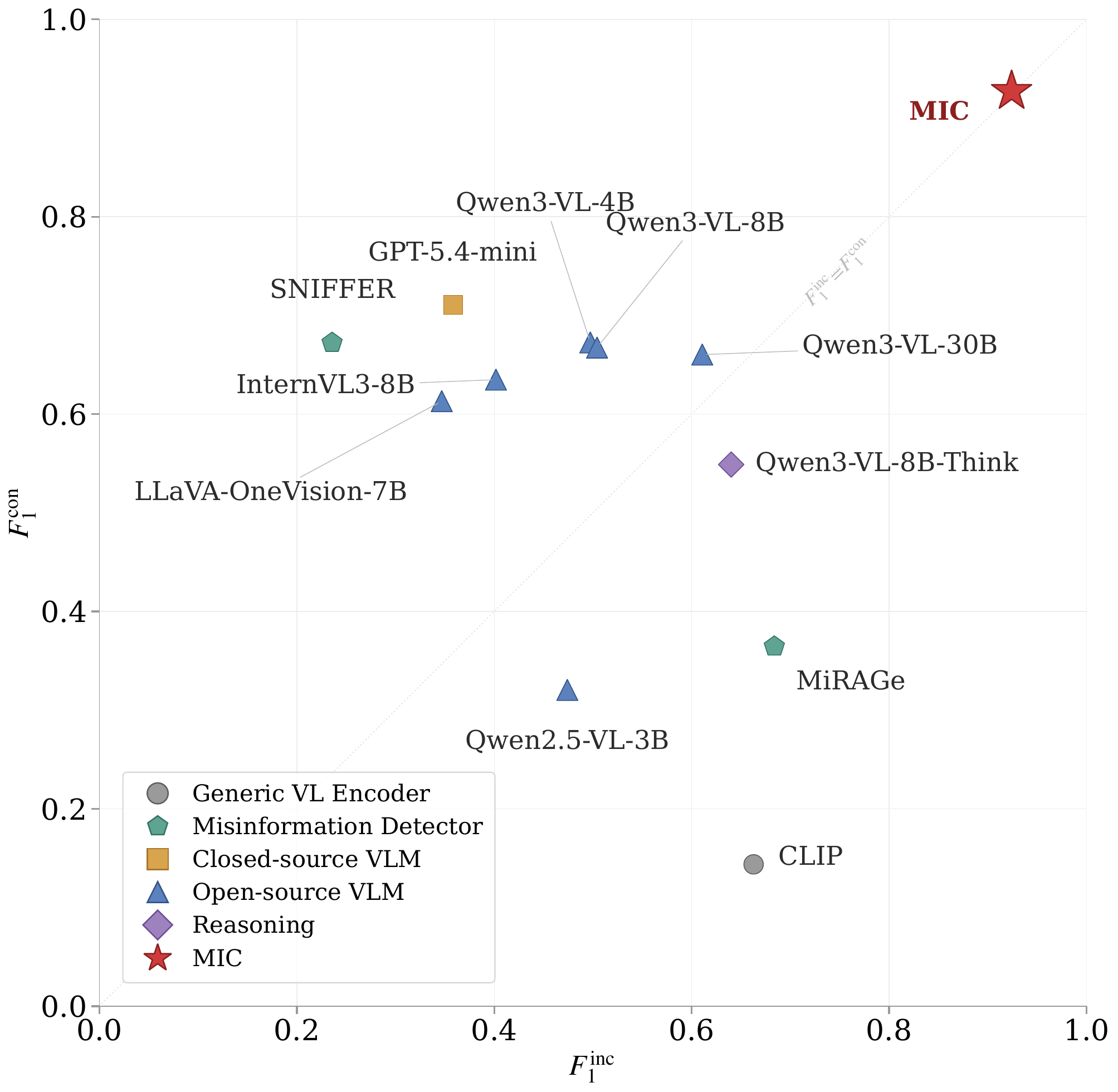}
    \caption{Verdict prediction performance on the test set.}
    \label{fig:verdict_classification}
\end{figure}

\subsection{Main Results}
\label{subsec:main_results}
We provide our main experimental results in this section.
Additional results are in Appendix~\ref{sec:additional_experimental_results}.

\paragraph{\ours{} achieves a higher Macro-F1 in verdict classification.}
Figure~\ref{fig:verdict_classification} compares the per-class F1 scores of \ours{} and representative baselines. 
$F_1^{\mathrm{inc}}$ measures inconsistency detection, while $F_1^{\mathrm{con}}$ measures the ability to correctly identify genuine claim-consistent images. 
Existing methods tend to favor one class over the other, suggesting a mismatch between their original assumptions and claim-grounded visual inconsistency detection. 
For example, SNIFFER transfers poorly because out-of-context misinformation assumes authentic images with misleading contexts, while \benchmark{} contains localized AI-generated edits that create claim-grounded inconsistencies.
MiRAGe over-predicts inconsistency, yielding high $F_1^{\mathrm{inc}}$ but low $F_1^{\mathrm{con}}$, whereas GPT-5.4-mini shows the opposite bias and misses many subtle inconsistencies. 
Overall, \ours{} achieves high F1 on both classes, leading to a stronger Macro-F1 and indicating that task-specific supervision and rewards better capture claim-grounded visual inconsistencies.

\paragraph{\ours{} substantially improves fine-grained inconsistency detection.}
Table~\ref{tab:main_results} shows that \ours{} substantially outperforms all baselines across both verdict prediction and fine-grained explanation metrics.
While several large VLM baselines, such as Qwen3-VL-30B-A3B and Qwen3-VL-8B-Thinking, achieve moderate Macro-F1, their much lower \emph{TypeAcc}, \emph{VisualSim}, and \emph{ExplSim} indicate that they struggle to identify the specific inconsistent element and explain why it contradicts the claim context.
In contrast, \ours{} achieves strong performance on both the in-distribution and out-of-distribution splits, suggesting that task-specific supervision and verifiable rewards better align the model with the structured verification process.
The performance gap remains large even when edited-in visual elements are held out from training, indicating generalization beyond memorized patterns.

\section{Analysis and Discussion}

\begin{figure}[!t]
    \centering
    \includegraphics[width=0.99\linewidth]{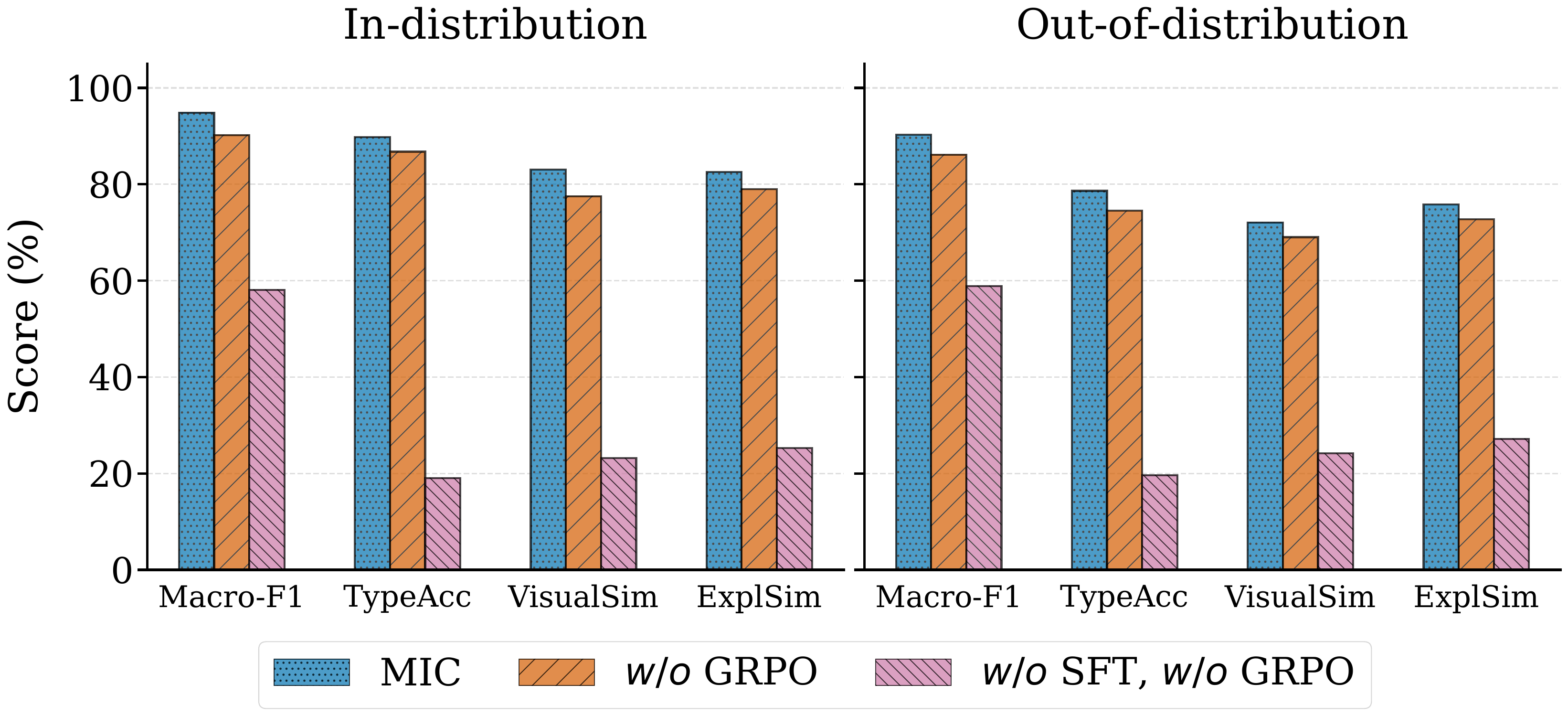}
    \caption{Ablation study of the \ours{} training stages.}
    \label{fig:ablation_study}
\end{figure}

\paragraph{Ablation Study.}
\label{subsec:ablation_study}
To analyze the contribution of each training stage, we compare the full \ours{} framework with an SFT-only model and the base MLLM. As shown in Figure~\ref{fig:ablation_study}, the base MLLM performs poorly across all metrics, with \emph{TypeAcc} dropping from $89.76\%$ to approximately $20\%$ on the in-distribution split. This shows that general multimodal pretraining alone is insufficient for fine-grained claim-grounded inconsistency detection.
SFT closes most of this gap, indicating that SFT is the main driver of structured output generation and task adaptation. Adding GRPO further improves \emph{TypeAcc}, \emph{VisualSim}, and \emph{ExplSim} across both test splits, suggesting that verifiable reward optimization enhances explanation grounding beyond likelihood-based supervision.

\begin{figure}
    \centering
    \includegraphics[width=0.99\linewidth]{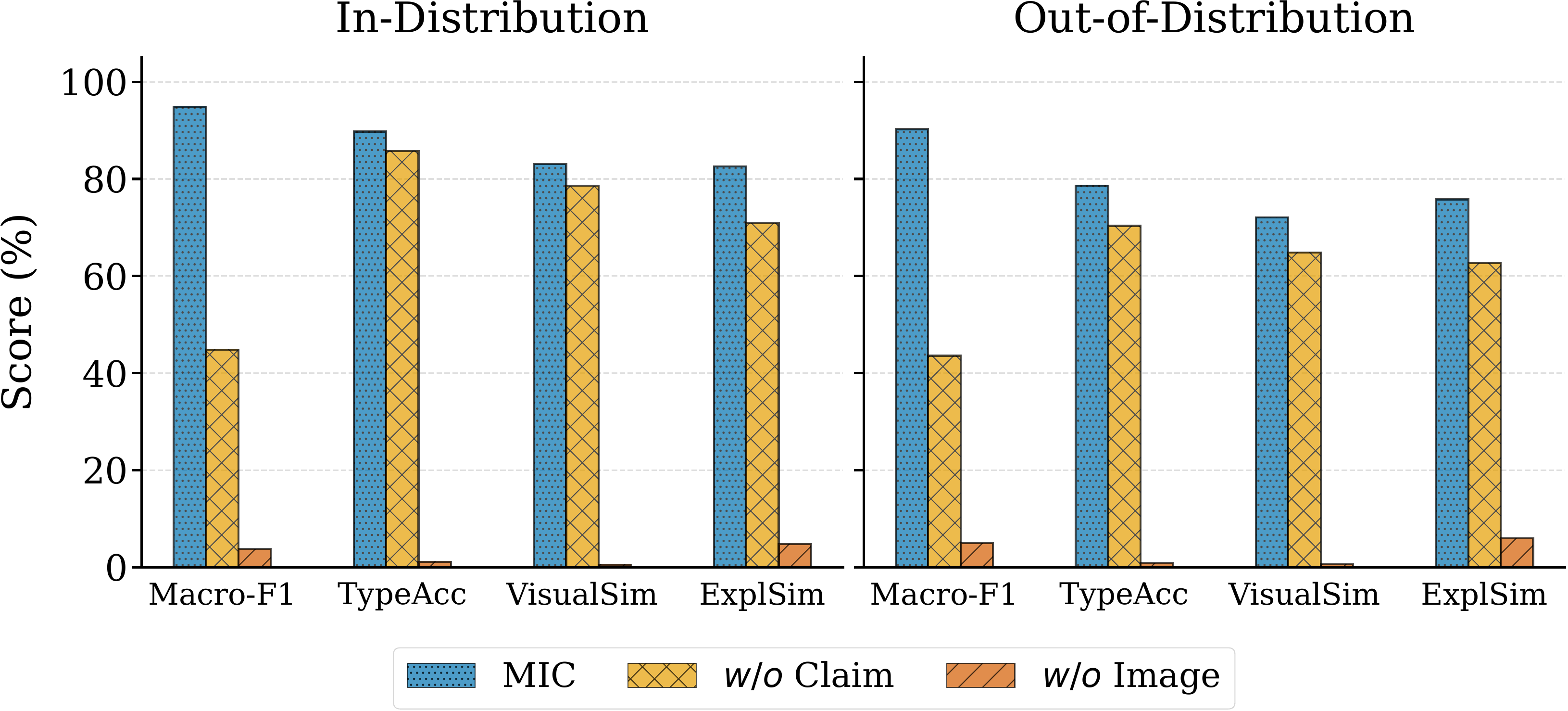}
    \caption{Impact of visual and textual inputs of \ours{}.}
    \label{fig:modality_perturbation}
\end{figure}

\paragraph{Impact of Visual and Textual Inputs.}
\label{subsec:modality_perturbation_study}
To investigate whether \ours{} relies on both modalities and whether \benchmark{} contains caption-only shortcuts, we evaluate two input-perturbation settings: removing the claim and removing the image. 
As shown in Figure~\ref{fig:modality_perturbation}, removing the claim degrades performance, especially on Macro-F1 and \emph{ExplSim}, indicating that textual context is necessary for determining whether a visual element contradicts the claimed event. 
Removing the image by replacing it with a black canvas leads to a more severe drop, particularly on \emph{TypeAcc}, \emph{VisualSim}, and \emph{ExplSim}. 
This suggests that the model cannot reliably infer inconsistencies from the claim alone, providing evidence against caption-only shortcuts in the data-generation process and confirming that visual evidence is essential for grounded prediction.

\begin{figure}[!t]
    \centering
    \includegraphics[width=0.99\linewidth]{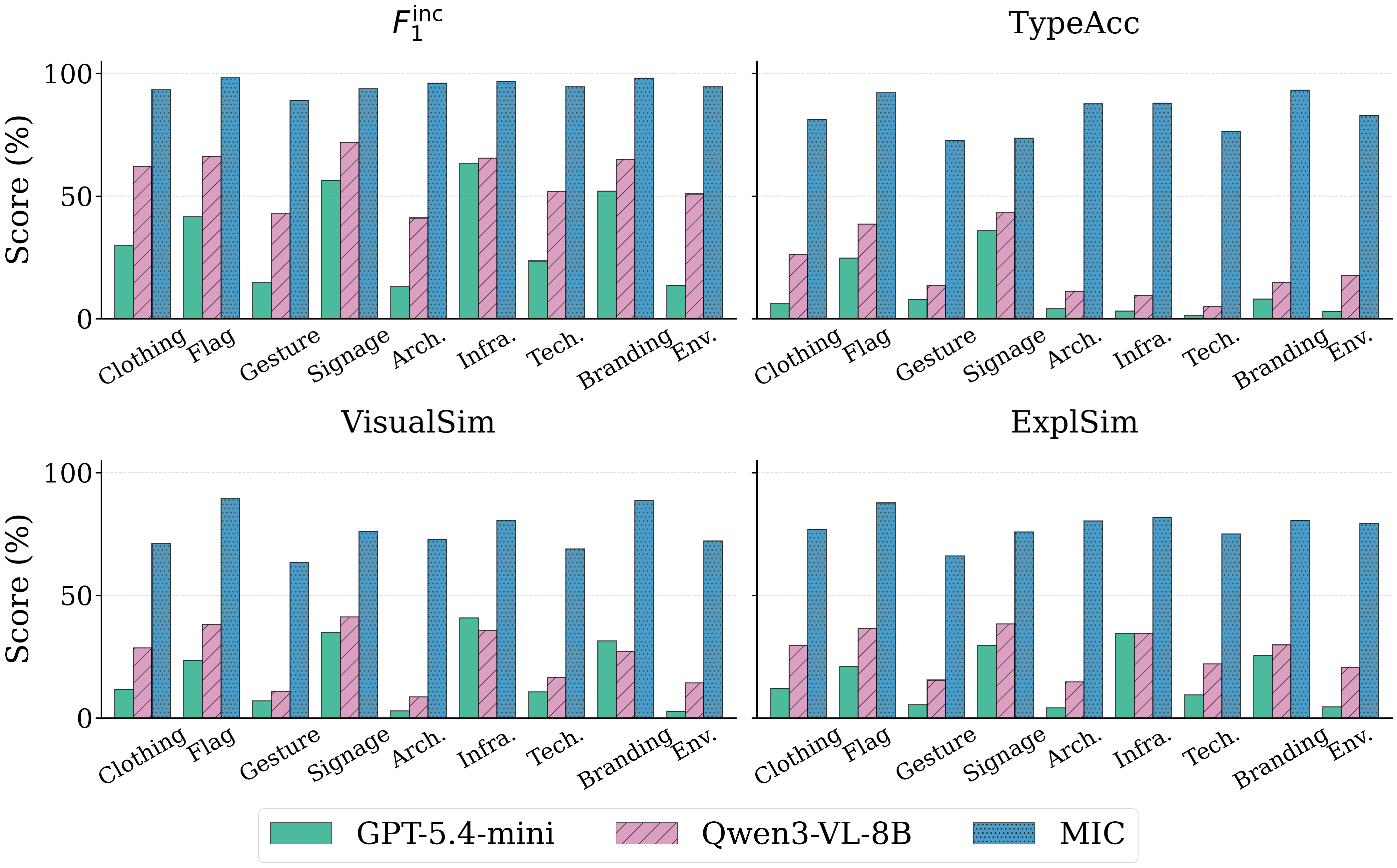}
    \caption{Fine-grained performance of \ours{} across different inconsistency types.}
    \label{fig:type_performance}
\end{figure}

\paragraph{Performance across Inconsistency Types.}
\label{subsec:type_analysis}

To evaluate the fine-grained capabilities of \ours{}, we analyze performance across the nine inconsistency types in Figure~\ref{fig:type_performance}.
\ours{} consistently outperforms GPT-5.4-mini and Qwen3-VL-8B-Instruct across all categories, achieving the best \emph{TypeAcc} for every inconsistency type.
This indicates that \ours{} can distinguish among different inconsistency types, rather than only detecting that a mismatch exists.
The gains are especially large for \emph{VisualSim} and \emph{ExplSim}, suggesting that baseline models often struggle to describe the manipulated visual element or provide a grounded world-knowledge explanation.
Overall, these results show that \ours{} captures type-specific inconsistency patterns while grounding its predictions in relevant world knowledge.

\begin{figure}[!t]
    \centering
    \includegraphics[width=0.99\linewidth]{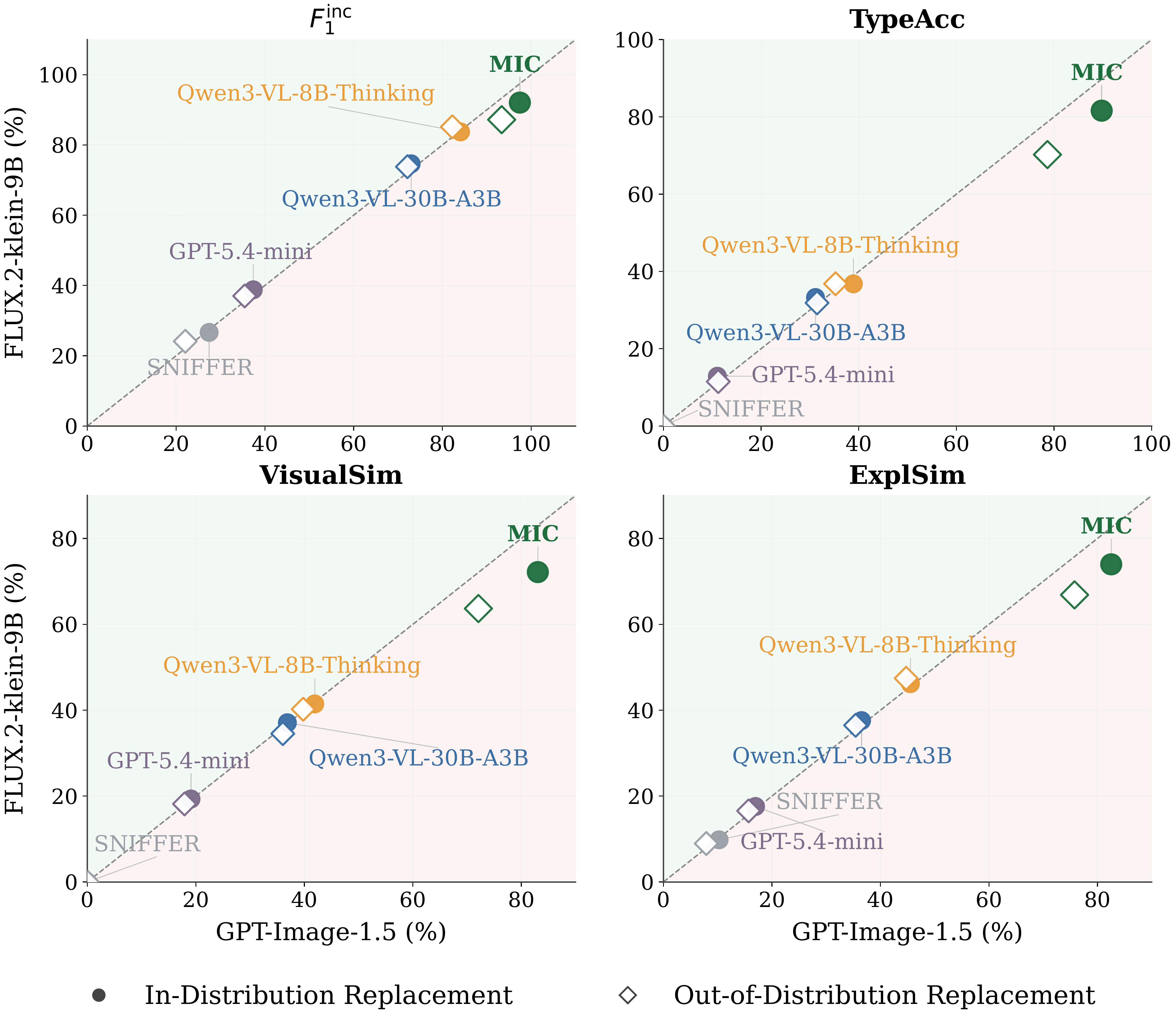}
    \caption{Generalization performance of \ours{} on images generated by FLUX.2-klein-9B.}
    \label{fig:flux2_performance}
\end{figure}

\paragraph{Generalization to Unseen Image Generators.}
\label{subsec:generator_generalization}
To evaluate generalization to unseen image generators, we test \ours{} and representative baselines on images synthesized by FLUX.2-klein-9B~\cite{blackforestlabs2026flux2klein} using the same edit instructions and evaluation protocol as \benchmark{}.
As shown in Figure~\ref{fig:flux2_performance}, \ours{} outperforms all baselines on FLUX-generated images, especially on \emph{TypeAcc}, \emph{VisualSim}, and \emph{ExplSim}.
At the same time, \ours{} shows a noticeable drop when evaluated on FLUX-generated images compared with GPT-generated images.
This suggests that task-specific fine-tuning improves inconsistency detection, but may also make the model partially adapted to the visual distribution of the training generator. 
Nevertheless, \ours{} still substantially outperforms all baselines on FLUX-generated images.

\begin{figure}
    \centering
    \includegraphics[width=0.99\linewidth]{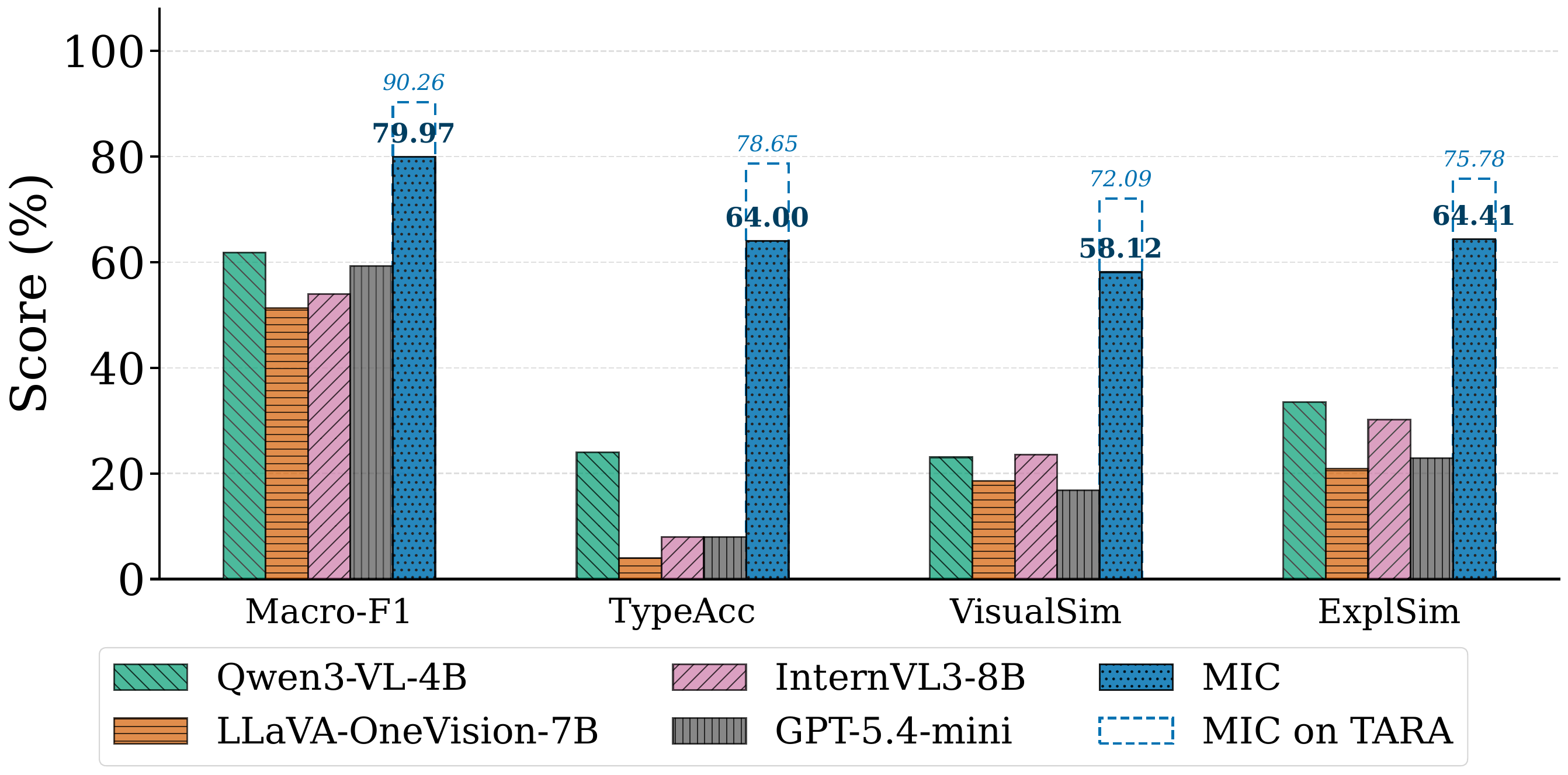}
    \caption{Generalization performance of \ours{} on recent 2026 news images.}
    \label{fig:recent_news}
\end{figure}

\paragraph{Generalization to Recent News Images.}
\label{subsec:recent_news_generalization}
To evaluate generalization to emerging events, we collect real-world news images from 25 recent events in 2026 using Wikimedia Commons and Wikipedia, and construct an additional benchmark following the same pipeline as \benchmark{}.
Representative examples are provided in Appendix~\ref{sec:recent_news_examples}.
As shown in Figure~\ref{fig:recent_news}, \ours{} outperforms all baselines, indicating that its claim-grounded inconsistency patterns transfer to emerging events.
However, the smaller gains on \emph{TypeAcc}, \emph{VisualSim}, and \emph{ExplSim} suggest that fine-grained reasoning remains challenging for events outside current MLLMs' parametric knowledge.
This motivates extending \ours{} with retrieval to support evolving-event reasoning.
Appendix~\ref{subsec:ai_generated_shortcut} further shows that \ours{} does not rely solely on whether an image appears AI-generated.

\begin{table}[t]
\centering\small
\begin{tabular}{l c}
\toprule
\textbf{Failure mode} & \textbf{Count} \\
\midrule
Missed cultural edit & 11 \\
Missed object or scene edit & 4 \\
Evidence describes feature not in image & 5 \\
Evidence describes unedited element & 1 \\
Wrong country or brand attribution & 9 \\
\bottomrule
\end{tabular}
\caption{Manual analysis of 30 \ours{} failure cases from the \benchmark{} test splits.}
\label{tab:failure_modes_summary}
\end{table}

\paragraph{Error Analysis.}
\label{subsec:error_analysis}

To characterize the error patterns of \ours{}, we manually analyze 30 failure cases sampled from the \benchmark{} test splits.
Table~\ref{tab:failure_modes_summary} summarizes the failure categories, with examples in Appendix~\ref{sec:error_examples}.
Most errors involve missed cultural, object, or scene-level edits that appear plausible in isolation but conflict with the claim context, highlighting the need for reasoning beyond object recognition.
Others involve unsupported visual evidence or incorrect country/brand attribution, where \ours{} identifies the right region but assigns the wrong contextual origin.
These errors motivate future work on evidence-grounded explanations for subtle geopolitical, cultural, and temporal inconsistencies.

\section{Conclusion and Future Work}
\label{sec:conclusion}

We presented \ours{}, an explainable framework for detecting AI-generated multimodal misinformation through image--claim consistency verification.
Our work studies \emph{claim-grounded visual inconsistency detection} as a specific setting of image--claim consistency verification, focusing on localized visual details that conflict with the context implied by the claim.
Using SFT and GRPO, \ours{} jointly predicts verdicts, inconsistency types, visual evidence descriptions, and world-knowledge explanations.

We also introduced \benchmark{}, a benchmark of 8,812 image--claim instances derived from 4,406 claims, with fine-grained annotations across nine inconsistency types.
GRPO improves verdict prediction and explanation similarity to reference annotations over SFT alone.
Further evaluations support transfer to held-out replacement entities, unseen image generators, and controlled tests based on recent news images.

Future work will incorporate external evidence retrieval for evolving world knowledge, extend the framework to broader forms of multimodal misinformation, and explore deployment in professional fact-checking workflows.


\section*{Limitations}
\label{sec:limitations}

Our work has three limitations: 

First, \benchmark{} focuses on a single controlled inconsistency per AI-generated image, enabling isolated evaluation of the introduced edit.
Although our two-inconsistency evaluation provides initial evidence of transfer beyond single-inconsistency training, evaluating more complex combinations of inconsistencies remains a direction for future work.

Second, \benchmark{} is limited to nine inconsistency categories and therefore does not cover all forms of world-knowledge inconsistency that may arise in real-world AI-generated news imagery. 
The current taxonomy is grounded in GIJN fact-checking guidelines and captures a practically relevant subset of such inconsistencies. 
Extending the taxonomy to broader and more diverse forms of inconsistency, ideally in collaboration with journalists and fact-checkers, remains an important direction for future work.

Third, like other MLLM-based approaches, \ours{} draws on the parametric knowledge of its underlying model, and its coverage is therefore tied to the model's training data.
This knowledge may be incomplete or outdated, particularly for recent events.
Future work will investigate retrieving background facts from text sources such as Wikipedia to support image--claim consistency judgments.

\section*{Ethics and Broader Impact}

This work aims to support journalists and professional fact-checkers by improving the verification of AI-generated multimodal misinformation through structured and interpretable explanations grounded in world knowledge. By moving beyond binary provenance prediction, \ours{} is designed to provide more transparent and actionable verification signals that can better support real-world fact-checking workflows.

At the same time, research on AI-generated misinformation detection raises broader ethical considerations. Although \benchmark{} is intended to advance defensive research, the same image-editing pipelines used to construct the dataset could potentially be misused to create deceptive or manipulative content outside research settings. To reduce this risk, \benchmark{} is built exclusively from publicly available news image--claim pairs from TARA, all edited images are explicitly labeled as AI-generated, and the dataset is released solely for research and benchmarking purposes.

Our work also has important limitations related to knowledge reliability and cultural coverage. Since \ours{} relies on the parametric knowledge of the underlying MLLM, its explanations may occasionally contain factual inaccuracies, outdated information, or culturally biased assumptions. Such risks may be amplified for underrepresented regions, languages, and visual cultures. Consequently, the outputs of \ours{} should not be treated as definitive factual judgments, but rather as decision-support signals that require human verification.

Human annotation was conducted by trained annotators for verification and explanation tasks and did not involve the collection of private or sensitive personal data. We emphasize that \ours{} is intended to assist rather than replace professional fact-checkers. Any deployment in public-facing verification systems should therefore include expert review, transparency about model limitations, and appropriate safeguards against overreliance on automated predictions.

\section*{Acknowledgments}
This work has been funded by the LOEWE initiative (Hesse, Germany) within the emergenCITY center (Grant Number: LOEWE/1/12/519/03/05.001(0016)/72), and the German Federal Ministry of Research, Technology and Space and the Hessian Ministry of Higher Education, Research, Science and the Arts through their joint support of the National Research Center for Applied Cybersecurity ATHENE. The figures have been designed using resources from Flaticon.com.

\bibliography{references}

@inproceedings{fu2022there,
    title = "There{'}s a Time and Place for Reasoning Beyond the Image",
    author = "Fu, Xingyu  and
      Zhou, Ben  and
      Chandratreya, Ishaan  and
      Vondrick, Carl  and
      Roth, Dan",
    editor = "Muresan, Smaranda  and
      Nakov, Preslav  and
      Villavicencio, Aline",
    booktitle = "Proceedings of the 60th Annual Meeting of the Association for Computational Linguistics (Volume 1: Long Papers)",
    month = may,
    year = "2022",
    address = "Dublin, Ireland",
    publisher = "Association for Computational Linguistics",
    url = "https://aclanthology.org/2022.acl-long.81/",
    doi = "10.18653/v1/2022.acl-long.81",
    pages = "1138--1149",
}

@article{shao2024deepseekmath,
  author       = {Zhihong Shao and
                  Peiyi Wang and
                  Qihao Zhu and
                  Runxin Xu and
                  Junxiao Song and
                  Mingchuan Zhang and
                  Y. K. Li and
                  Y. Wu and
                  Daya Guo},
  title        = {DeepSeekMath: Pushing the Limits of Mathematical Reasoning in Open
                  Language Models},
  journal      = {CoRR},
  volume       = {abs/2402.03300},
  year         = {2024},
  url          = {https://doi.org/10.48550/arXiv.2402.03300},
  doi          = {10.48550/ARXIV.2402.03300},
  eprinttype   = {arXiv},
  eprint       = {2402.03300},
  bibsource    = {dblp computer science bibliography, https://dblp.org}
}

@article{guo2025deepseek,
  author       = {DeepSeek{-}AI},
  title        = {DeepSeek-R1: Incentivizing Reasoning Capability in LLMs via Reinforcement
                  Learning},
  journal      = {CoRR},
  volume       = {abs/2501.12948},
  year         = {2025},
  url          = {https://doi.org/10.48550/arXiv.2501.12948},
  doi          = {10.48550/ARXIV.2501.12948},
  eprinttype   = {arXiv},
  eprint       = {2501.12948},
  bibsource    = {dblp computer science bibliography, https://dblp.org}
}

@inproceedings{qi2024sniffer,
  author       = {Peng Qi and
                  Zehong Yan and
                  Wynne Hsu and
                  Mong{-}Li Lee},
  title        = {Sniffer: Multimodal Large Language Model for Explainable Out-of-Context
                  Misinformation Detection},
  booktitle    = {{IEEE/CVF} Conference on Computer Vision and Pattern Recognition,
                  {CVPR} 2024, Seattle, WA, USA, June 16-22, 2024},
  pages        = {13052--13062},
  publisher    = {{IEEE}},
  year         = {2024},
  url          = {https://doi.org/10.1109/CVPR52733.2024.01240},
  doi          = {10.1109/CVPR52733.2024.01240},
  bibsource    = {dblp computer science bibliography, https://dblp.org}
}

@inproceedings{radford2021learning,
  author       = {Alec Radford and
                  Jong Wook Kim and
                  Chris Hallacy and
                  Aditya Ramesh and
                  Gabriel Goh and
                  Sandhini Agarwal and
                  Girish Sastry and
                  Amanda Askell and
                  Pamela Mishkin and
                  Jack Clark and
                  Gretchen Krueger and
                  Ilya Sutskever},
  editor       = {Marina Meila and
                  Tong Zhang},
  title        = {Learning Transferable Visual Models From Natural Language Supervision},
  booktitle    = {Proceedings of the 38th International Conference on Machine Learning,
                  {ICML} 2021, 18-24 July 2021, Virtual Event},
  series       = {Proceedings of Machine Learning Research},
  pages        = {8748--8763},
  publisher    = {{PMLR}},
  year         = {2021},
  url          = {http://proceedings.mlr.press/v139/radford21a.html},
  bibsource    = {dblp computer science bibliography, https://dblp.org}
}

@inproceedings{
jifakexplain,
title={FakeXplain: {AI}-Generated Image Detection via Human-Aligned Grounded Reasoning},
author={Yikun Ji and Yan Hong and Qi Fan and jun lan and Huijia Zhu and Weiqiang Wang and Liqing Zhang and Jianfu Zhang},
booktitle={The Fourteenth International Conference on Learning Representations},
year={2026},
url={https://openreview.net/forum?id=UcpTOa8OnG}
}

@inproceedings{
hulora,
title={Lo{RA}: Low-Rank Adaptation of Large Language Models},
author={Edward J Hu and yelong shen and Phillip Wallis and Zeyuan Allen-Zhu and Yuanzhi Li and Shean Wang and Lu Wang and Weizhu Chen},
booktitle={International Conference on Learning Representations},
year={2022},
url={https://openreview.net/forum?id=nZeVKeeFYf9}
}

@article{zhang2025qwen3,
  author       = {Yanzhao Zhang and
                  Mingxin Li and
                  Dingkun Long and
                  Xin Zhang and
                  Huan Lin and
                  Baosong Yang and
                  Pengjun Xie and
                  An Yang and
                  Dayiheng Liu and
                  Junyang Lin and
                  Fei Huang and
                  Jingren Zhou},
  title        = {Qwen3 Embedding: Advancing Text Embedding and Reranking Through Foundation
                  Models},
  journal      = {CoRR},
  volume       = {abs/2506.05176},
  year         = {2025},
  url          = {https://doi.org/10.48550/arXiv.2506.05176},
  doi          = {10.48550/ARXIV.2506.05176},
  eprinttype   = {arXiv},
  eprint       = {2506.05176},
  bibsource    = {dblp computer science bibliography, https://dblp.org}
}

@article{geng2025m4fc,
  title={{M4FC}: A multimodal, multilingual, multicultural, multitask real-world fact-checking dataset},
  author={Geng, Jiahui and Tonglet, Jonathan and Gurevych, Iryna},
  journal={arXiv preprint arXiv:2510.23508},
  year={2025}
}

@inproceedings{tonglet-etal-2025-cove,
    title = "{COVE}: {CO}ntext and {VE}racity prediction for out-of-context images",
    author = "Tonglet, Jonathan  and
      Thiem, Gabriel  and
      Gurevych, Iryna",
    editor = "Chiruzzo, Luis  and
      Ritter, Alan  and
      Wang, Lu",
    booktitle = "Proceedings of the 2025 Conference of the Nations of the Americas Chapter of the Association for Computational Linguistics: Human Language Technologies (Volume 1: Long Papers)",
    month = apr,
    year = "2025",
    address = "Albuquerque, New Mexico",
    publisher = "Association for Computational Linguistics",
    url = "https://aclanthology.org/2025.naacl-long.102/",
    doi = "10.18653/v1/2025.naacl-long.102",
    pages = "2029--2049",
    ISBN = "979-8-89176-189-6"
}

@inproceedings{wang2020cnn,
  author       = {Sheng{-}Yu Wang and
                  Oliver Wang and
                  Richard Zhang and
                  Andrew Owens and
                  Alexei A. Efros},
  title        = {CNN-Generated Images Are Surprisingly Easy to Spot... for Now},
  booktitle    = {2020 {IEEE/CVF} Conference on Computer Vision and Pattern Recognition,
                  {CVPR} 2020, Seattle, WA, USA, June 13-19, 2020},
  pages        = {8692--8701},
  publisher    = {Computer Vision Foundation / {IEEE}},
  year         = {2020},
  url          = {https://openaccess.thecvf.com/content\_CVPR\_2020/html/Wang\_CNN-Generated\_Images\_Are\_Surprisingly\_Easy\_to\_Spot...\_for\_Now\_CVPR\_2020\_paper.html},
  doi          = {10.1109/CVPR42600.2020.00872},
  bibsource    = {dblp computer science bibliography, https://dblp.org}
}

@inproceedings{frank2020leveraging,
  author       = {Joel Frank and
                  Thorsten Eisenhofer and
                  Lea Sch{\"{o}}nherr and
                  Asja Fischer and
                  Dorothea Kolossa and
                  Thorsten Holz},
  title        = {Leveraging Frequency Analysis for Deep Fake Image Recognition},
  booktitle    = {Proceedings of the 37th International Conference on Machine Learning,
                  {ICML} 2020, 13-18 July 2020, Virtual Event},
  series       = {Proceedings of Machine Learning Research},
  pages        = {3247--3258},
  publisher    = {{PMLR}},
  year         = {2020},
  url          = {http://proceedings.mlr.press/v119/frank20a.html},
  bibsource    = {dblp computer science bibliography, https://dblp.org}
}

@article{gowal2025synthid,
  author       = {Sven Gowal and
                  Rudy Bunel and
                  Florian Stimberg and
                  David Stutz and
                  Guillermo Ortiz{-}Jim{\'{e}}nez and
                  Christina Kouridi and
                  Mel Vecer{\'{\i}}k and
                  Jamie Hayes and
                  Sylvestre{-}Alvise Rebuffi and
                  Paul Bernard and
                  Chris Gamble and
                  Mikl{\'{o}}s Z. Horv{\'{a}}th and
                  Fabian Kaczmarczyck and
                  Alex Kaskasoli and
                  Aleksandar Petrov and
                  Ilia Shumailov and
                  Meghana Thotakuri and
                  Olivia Wiles and
                  Jessica Yung and
                  Zahra Ahmed and
                  Victor Martin and
                  Simon Rosen and
                  Christopher Savcak and
                  Armin Senoner and
                  Nidhi Vyas and
                  Pushmeet Kohli},
  title        = {SynthID-Image: Image watermarking at internet scale},
  journal      = {CoRR},
  volume       = {abs/2510.09263},
  year         = {2025},
  url          = {https://doi.org/10.48550/arXiv.2510.09263},
  doi          = {10.48550/ARXIV.2510.09263},
  eprinttype   = {arXiv},
  eprint       = {2510.09263},
  bibsource    = {dblp computer science bibliography, https://dblp.org}
}

@inproceedings{zhou2025aigi,
  author       = {Ziyin Zhou and
                  Yunpeng Luo and
                  Yuanchen Wu and
                  Ke Sun and
                  Jiayi Ji and
                  Ke Yan and
                  Shouhong Ding and
                  Xiaoshuai Sun and
                  Yunsheng Wu and
                  Rongrong Ji},
  title        = {Aigi-Holmes: Towards Explainable and Generalizable AI-Generated Image
                  Detection via Multimodal Large Language Models},
  booktitle    = {{IEEE/CVF} International Conference on Computer Vision, {ICCV} 2025,
                  Honolulu, HI, USA, October 19-25, 2025},
  pages        = {18746--18758},
  publisher    = {{IEEE}},
  year         = {2025},
  url          = {https://doi.org/10.1109/ICCV51701.2025.01742},
  doi          = {10.1109/ICCV51701.2025.01742},
  bibsource    = {dblp computer science bibliography, https://dblp.org}
}

@article{pei2024deepfake,
  author       = {Gan Pei and
                  Jiangning Zhang and
                  Menghan Hu and
                  Zhenyu Zhang and
                  Chengjie Wang and
                  Yunsheng Wu and
                  Guangtao Zhai and
                  Jian Yang and
                  Dacheng Tao},
  title        = {Deepfake Generation and Detection: {A} Benchmark and Survey},
  journal      = {{ACM} Comput. Surv.},
  volume       = {58},
  number       = {11},
  pages        = {273:1--273:41},
  year         = {2026},
  url          = {https://doi.org/10.1145/3801962},
  doi          = {10.1145/3801962},
  bibsource    = {dblp computer science bibliography, https://dblp.org}
}

@article{xu2025fully,
    author  = {Qijie Xu and Can Wang and Jiawei Chen and Siwei Lyu and Defang Chen},
    title   = {Fully AI-Generated Image Detection: Definition, Recent Advances and Challenges},
    journal = {CoRR},
    volume  = {abs/2502.19716},
    year    = {2025},
    url     = {https://arxiv.org/abs/2502.19716v2}
    }

@article{liu2024evolving,
  title={Evolving from single-modal to multi-modal facial deepfake detection: Progress and challenges},
  author={Liu, Ping and Tao, Qiqi and Zhou, Joey Tianyi},
  journal={arXiv preprint arXiv:2406.06965},
  year={2024},
  url={https://arxiv.org/abs/2406.06965}
}

@inproceedings{keita2025deeclip,
  author       = {Mamadou Keita and
                  Wassim Hamidouche and
                  Hessen Bougueffa Eutamene and
                  Abdelmalik Taleb{-}Ahmed and
                  Abdenour Hadid},
  editor       = {Jacques Blanc{-}Talon and
                  Patrice Delmas and
                  Hiroki Takahashi and
                  Minami Yasuhiro},
  title        = {DeeCLIP: {A} Robust and Generalizable Transformer-Based Framework
                  for Detecting AI-Generated Images},
  booktitle    = {Advanced Concepts for Intelligent Vision Systems - 22nd International
                  Conference, {ACIVS} 2025, Tokyo, Japan, July 28-30, 2025, Proceedings},
  series       = {Lecture Notes in Computer Science},
  pages        = {146--158},
  publisher    = {Springer},
  year         = {2025},
  url          = {https://doi.org/10.1007/978-3-032-07343-3\_12},
  doi          = {10.1007/978-3-032-07343-3\_12},
  bibsource    = {dblp computer science bibliography, https://dblp.org}
}

@article{tan2025semantic,
  author       = {Chuangchuang Tan and
                  Xiang Ming and
                  Jinglu Wang and
                  Renshuai Tao and
                  Bin Li and
                  Yunchao Wei and
                  Yao Zhao and
                  Yan Lu},
  title        = {Semantic Visual Anomaly Detection and Reasoning in AI-Generated Images},
  journal      = {CoRR},
  volume       = {abs/2510.10231},
  year         = {2025},
  url          = {https://doi.org/10.48550/arXiv.2510.10231},
  doi          = {10.48550/ARXIV.2510.10231},
  eprinttype   = {arXiv},
  eprint       = {2510.10231},
  bibsource    = {dblp computer science bibliography, https://dblp.org}
}

@inproceedings{xu2025fakeshield,
  author       = {Zhipei Xu and
                  Xuanyu Zhang and
                  Runyi Li and
                  Zecheng Tang and
                  Qing Huang and
                  Jian Zhang},
  title        = {FakeShield: Explainable Image Forgery Detection and Localization via
                  Multi-modal Large Language Models},
  booktitle    = {The Thirteenth International Conference on Learning Representations,
                  {ICLR} 2025, Singapore, April 24-28, 2025},
  publisher    = {OpenReview.net},
  year         = {2025},
  url          = {https://openreview.net/forum?id=pAQzEY7M03},
  bibsource    = {dblp computer science bibliography, https://dblp.org}
}

@inproceedings{yang2025heie,
  author       = {Fan Yang and
                  Ru Zhen and
                  Jianing Wang and
                  Yanhao Zhang and
                  Haoxiang Chen and
                  Haonan Lu and
                  Sicheng Zhao and
                  Guiguang Ding},
  title        = {{HEIE:} MLLM-Based Hierarchical Explainable {AIGC} Image Implausibility
                  Evaluator},
  booktitle    = {{IEEE/CVF} Conference on Computer Vision and Pattern Recognition,
                  {CVPR} 2025, Nashville, TN, USA, June 11-15, 2025},
  pages        = {3856--3866},
  publisher    = {Computer Vision Foundation / {IEEE}},
  year         = {2025},
  url          = {https://openaccess.thecvf.com/content/CVPR2025/html/Yang\_HEIE\_MLLM-Based\_Hierarchical\_Explainable\_AIGC\_Image\_Implausibility\_Evaluator\_CVPR\_2025\_paper.html},
  doi          = {10.1109/CVPR52734.2025.00365},
  bibsource    = {dblp computer science bibliography, https://dblp.org}
}

@inproceedings{tan2025c2p,
  author       = {Chuangchuang Tan and
                  Renshuai Tao and
                  Huan Liu and
                  Guanghua Gu and
                  Baoyuan Wu and
                  Yao Zhao and
                  Yunchao Wei},
  editor       = {Toby Walsh and
                  Julie Shah and
                  Zico Kolter},
  title        = {{C2P-CLIP:} Injecting Category Common Prompt in {CLIP} to Enhance
                  Generalization in Deepfake Detection},
  booktitle    = {Thirty-Ninth {AAAI} Conference on Artificial Intelligence, Thirty-Seventh
                  Conference on Innovative Applications of Artificial Intelligence,
                  Fifteenth Symposium on Educational Advances in Artificial Intelligence,
                  {AAAI} 2025, Philadelphia, PA, USA, February 25 - March 4, 2025},
  pages        = {7184--7192},
  publisher    = {{AAAI} Press},
  year         = {2025},
  url          = {https://doi.org/10.1609/aaai.v39i7.32772},
  doi          = {10.1609/AAAI.V39I7.32772},
  bibsource    = {dblp computer science bibliography, https://dblp.org}
}

@inproceedings{rosenthol2022c2pa,
  title={C2pa: the world’s first industry standard for content provenance (conference presentation)},
  author={Rosenthol, Leonard},
  booktitle={Applications of Digital Image Processing XLV},
  volume={12226},
  pages={122260P},
  year={2022},
  organization={SPIE},
  url={https://www.spiedigitallibrary.org/conference-proceedings-of-spie/12226/122260P/C2PA-the-worlds-first-industry-standard-for-content-provenance/10.1117/12.2632021.short}
}

@article{guo2022survey,
    title = "A Survey on Automated Fact-Checking",
    author = "Guo, Zhijiang  and
      Schlichtkrull, Michael  and
      Vlachos, Andreas",
    editor = "Roark, Brian  and
      Nenkova, Ani",
    journal = "Transactions of the Association for Computational Linguistics",
    volume = "10",
    year = "2022",
    address = "Cambridge, MA",
    publisher = "MIT Press",
    url = "https://aclanthology.org/2022.tacl-1.11/",
    doi = "10.1162/tacl_a_00454",
    pages = "178--206"
}

@inproceedings{huang2024miragenews,
    title = "{M}i{RAG}e{N}ews: Multimodal Realistic {AI}-Generated News Detection",
    author = "Huang, Runsheng  and
      Dugan, Liam  and
      Yang, Yue  and
      Callison-Burch, Chris",
    editor = "Al-Onaizan, Yaser  and
      Bansal, Mohit  and
      Chen, Yun-Nung",
    booktitle = "Findings of the Association for Computational Linguistics: EMNLP 2024",
    month = nov,
    year = "2024",
    address = "Miami, Florida, USA",
    publisher = "Association for Computational Linguistics",
    url = "https://aclanthology.org/2024.findings-emnlp.959/",
    doi = "10.18653/v1/2024.findings-emnlp.959",
    pages = "16436--16448"
}

@inproceedings{zhang2025generation,
    title = "From Generation to Detection: A Multimodal Multi-Task Dataset for Benchmarking Health Misinformation",
    author = "Zhang, Zhihao  and
      Zhang, Yiran  and
      Zhou, Xiyue  and
      Huang, Liting  and
      Razzak, Imran  and
      Nakov, Preslav  and
      Naseem, Usman",
    editor = "Christodoulopoulos, Christos  and
      Chakraborty, Tanmoy  and
      Rose, Carolyn  and
      Peng, Violet",
    booktitle = "Findings of the Association for Computational Linguistics: EMNLP 2025",
    month = nov,
    year = "2025",
    address = "Suzhou, China",
    publisher = "Association for Computational Linguistics",
    url = "https://aclanthology.org/2025.findings-emnlp.1316/",
    doi = "10.18653/v1/2025.findings-emnlp.1316",
    pages = "24245--24260",
    ISBN = "979-8-89176-335-7"
}

@article{dufour2024ammeba,
  author       = {Nicholas Dufour and
                  Arkanath Pathak and
                  Pouya Samangouei and
                  Nikki Hariri and
                  Shashi Deshetti and
                  Andrew Dudfield and
                  Christopher Guess and
                  Pablo Hern{\'{a}}ndez Escayola and
                  Bobby Tran and
                  Mevan Babakar and
                  Christoph Bregler},
  title        = {AMMeBa: {A} Large-Scale Survey and Dataset of Media-Based Misinformation
                  In-The-Wild},
  journal      = {CoRR},
  volume       = {abs/2405.11697},
  year         = {2024},
  url          = {https://doi.org/10.48550/arXiv.2405.11697},
  doi          = {10.48550/ARXIV.2405.11697},
  eprinttype   = {arXiv},
  eprint       = {2405.11697},
  bibsource    = {dblp computer science bibliography, https://dblp.org}
}

@article{cazzamatta2025ai,
  title={AI-generated misinformation: A case study on emerging trends in fact-checking practices across Brazil, Germany, and the United Kingdom},
  author={Cazzamatta, Regina and Sar{\i}sakalo{\u{g}}lu, Aynur},
  journal={Emerging Media},
  volume={3},
  number={2},
  pages={214--251},
  year={2025},
  publisher={SAGE Publications Sage UK: London, England},
  url={https://journals.sagepub.com/doi/full/10.1177/27523543251344971}
}

@article{wack2024political,
  author       = {Morgan Wack and
                  Kayla Duskin and
                  Damian Hodel},
  title        = {Political Fact-Checking Efforts are Constrained by Deficiencies in
                  Coverage, Speed, and Reach},
  journal      = {CoRR},
  volume       = {abs/2412.13280},
  year         = {2024},
  url          = {https://doi.org/10.48550/arXiv.2412.13280},
  doi          = {10.48550/ARXIV.2412.13280},
  eprinttype   = {arXiv},
  eprint       = {2412.13280},
  bibsource    = {dblp computer science bibliography, https://dblp.org}
}

@article{drolsbach2025characterizing,
  author       = {Chiara Drolsbach and
                  Nicolas Pr{\"{o}}llochs},
  title        = {Characterizing AI-Generated Misinformation on Social Media},
  journal      = {CoRR},
  volume       = {abs/2505.10266},
  year         = {2025},
  url          = {https://doi.org/10.48550/arXiv.2505.10266},
  doi          = {10.48550/ARXIV.2505.10266},
  eprinttype   = {arXiv},
  eprint       = {2505.10266},
  bibsource    = {dblp computer science bibliography, https://dblp.org}
}

@inproceedings{karageorgiou2025any,
  author       = {Dimitrios Karageorgiou and
                  Symeon Papadopoulos and
                  Ioannis Kompatsiaris and
                  Efstratios Gavves},
  title        = {Any-Resolution AI-Generated Image Detection by Spectral Learning},
  booktitle    = {{IEEE/CVF} Conference on Computer Vision and Pattern Recognition,
                  {CVPR} 2025, Nashville, TN, USA, June 11-15, 2025},
  pages        = {18706--18717},
  publisher    = {Computer Vision Foundation / {IEEE}},
  year         = {2025},
  url          = {https://openaccess.thecvf.com/content/CVPR2025/html/Karageorgiou\_Any-Resolution\_AI-Generated\_Image\_Detection\_by\_Spectral\_Learning\_CVPR\_2025\_paper.html},
  doi          = {10.1109/CVPR52734.2025.01743},
  bibsource    = {dblp computer science bibliography, https://dblp.org}
}

@inproceedings{gye2025reducing,
  author       = {Seoyeon Gye and
                  Junwon Ko and
                  Hyounguk Shon and
                  Minchan Kwon and
                  Junmo Kim},
  title        = {Reducing the Content Bias for AI-generated Image Detection},
  booktitle    = {{IEEE/CVF} Winter Conference on Applications of Computer Vision, {WACV}
                  2025, Tucson, AZ, USA, February 26 - March 6, 2025},
  pages        = {399--408},
  publisher    = {{IEEE}},
  year         = {2025},
  url          = {https://doi.org/10.1109/WACV61041.2025.00049},
  doi          = {10.1109/WACV61041.2025.00049},
  bibsource    = {dblp computer science bibliography, https://dblp.org}
}

@misc{awosoro2026dubawa,
  author       = {Sunday Awosoro},
  title        = {Photo of Muslims worshipping on Abuja road, AI-generated},
  year         = {2026},
  month        = mar,
  day          = {30},
  howpublished = {Dubawa},
  url          = {https://dubawa.org/photo-of-muslims-worshipping-on-abuja-road-ai-generated},
  note         = {Accessed: 2026-05-12}
}

@misc{gijn2025detectingai,
  author       = {Henk van Ess},
  title        = {Reporter’s Guide to Detecting AI-Generated Content},
  year         = {2025},
  month        = sep, 
  day          = {1},
  url          = {https://gijn.org/resource/guide-detecting-ai-generated-content},
  howpublished = {Global Investigative Journalism Network},
  note         = {Accessed: 2026-05-12}
}

@misc{blackforestlabs2026flux2klein,
    author       = {{Black Forest Labs}},
    title        = {{FLUX.2 [klein]}: Towards Interactive Visual Intelligence},
    year         = {2026},
    month        = jan,
    day          = {15},
    howpublished = {Black Forest Labs Blog},
    url = {https://bfl.ai/blog/flux2-klein-towards-interactive-visual-intelligence},
    note         = {Accessed: 2026-05-12}
    }

@INPROCEEDINGS{10944123,
  author       = {Stefanos{-}Iordanis Papadopoulos and
                  Christos Koutlis and
                  Symeon Papadopoulos and
                  Panagiotis C. Petrantonakis},
  title        = {Similarity Over Factuality: Are we Making Progress on Multimodal Out-of-Context
                  Misinformation Detection?},
  booktitle    = {{IEEE/CVF} Winter Conference on Applications of Computer Vision, {WACV}
                  2025, Tucson, AZ, USA, February 26 - March 6, 2025},
  pages        = {5041--5050},
  publisher    = {{IEEE}},
  year         = {2025},
  url          = {https://doi.org/10.1109/WACV61041.2025.00544},
  doi          = {10.1109/WACV61041.2025.00544},
  bibsource    = {dblp computer science bibliography, https://dblp.org}
}

@inproceedings{akhtar-etal-2023-multimodal,
    title = "Multimodal Automated Fact-Checking: A Survey",
    author = "Akhtar, Mubashara  and
      Schlichtkrull, Michael  and
      Guo, Zhijiang  and
      Cocarascu, Oana  and
      Simperl, Elena  and
      Vlachos, Andreas",
    editor = "Bouamor, Houda  and
      Pino, Juan  and
      Bali, Kalika",
    booktitle = "Findings of the Association for Computational Linguistics: EMNLP 2023",
    month = dec,
    year = "2023",
    address = "Singapore",
    publisher = "Association for Computational Linguistics",
    url = "https://aclanthology.org/2023.findings-emnlp.361/",
    doi = "10.18653/v1/2023.findings-emnlp.361",
    pages = "5430--5448"
}

@inproceedings{nakov2021automated,
  author       = {Preslav Nakov and
                  David P. A. Corney and
                  Maram Hasanain and
                  Firoj Alam and
                  Tamer Elsayed and
                  Alberto Barr{\'{o}}n{-}Cede{\~{n}}o and
                  Paolo Papotti and
                  Shaden Shaar and
                  Giovanni Da San Martino},
  editor       = {Zhi{-}Hua Zhou},
  title        = {Automated Fact-Checking for Assisting Human Fact-Checkers},
  booktitle    = {Proceedings of the Thirtieth International Joint Conference on Artificial
                  Intelligence, {IJCAI} 2021, Virtual Event / Montreal, Canada, 19-27
                  August 2021},
  pages        = {4551--4558},
  publisher    = {ijcai.org},
  year         = {2021},
  url          = {https://doi.org/10.24963/ijcai.2021/619},
  doi          = {10.24963/IJCAI.2021/619},
  bibsource    = {dblp computer science bibliography, https://dblp.org}
}

@article{bai2025qwen3,
  author       = {Qwen Team},
  title        = {Qwen3-VL Technical Report},
  journal      = {CoRR},
  volume       = {abs/2511.21631},
  year         = {2025},
  url          = {https://doi.org/10.48550/arXiv.2511.21631},
  doi          = {10.48550/ARXIV.2511.21631},
  eprinttype   = {arXiv},
  eprint       = {2511.21631},
  bibsource    = {dblp computer science bibliography, https://dblp.org}
}

@article{li2024llava,
  author       = {Bo Li and
                  Yuanhan Zhang and
                  Dong Guo and
                  Renrui Zhang and
                  Feng Li and
                  Hao Zhang and
                  Kaichen Zhang and
                  Peiyuan Zhang and
                  Yanwei Li and
                  Ziwei Liu and
                  Chunyuan Li},
  title        = {LLaVA-OneVision: Easy Visual Task Transfer},
  journal      = {Trans. Mach. Learn. Res.},
  volume       = {2025},
  year         = {2025},
  url          = {https://openreview.net/forum?id=zKv8qULV6n},
  bibsource    = {dblp computer science bibliography, https://dblp.org}
}

@article{zhu2025internvl3,
  author       = {Jinguo Zhu and
                  Weiyun Wang and
                  Zhe Chen and
                  Zhaoyang Liu and
                  Shenglong Ye and
                  Lixin Gu and
                  Hao Tian and
                  Yuchen Duan and
                  Weijie Su and
                  Jie Shao and
                  Zhangwei Gao and
                  Erfei Cui and
                  Xuehui Wang and
                  Yue Cao and
                  Yangzhou Liu and
                  Xingguang Wei and
                  Hongjie Zhang and
                  Haomin Wang and
                  Weiye Xu and
                  Hao Li and
                  Jiahao Wang and
                  Nianchen Deng and
                  Songze Li and
                  Yinan He and
                  Tan Jiang and
                  Jiapeng Luo and
                  Yi Wang and
                  Conghui He and
                  Botian Shi and
                  Xingcheng Zhang and
                  Wenqi Shao and
                  Junjun He and
                  Yingtong Xiong and
                  Wenwen Qu and
                  Peng Sun and
                  Penglong Jiao and
                  Han Lv and
                  Lijun Wu and
                  Kaipeng Zhang and
                  Huipeng Deng and
                  Jiaye Ge and
                  Kai Chen and
                  Limin Wang and
                  Min Dou and
                  Lewei Lu and
                  Xizhou Zhu and
                  Tong Lu and
                  Dahua Lin and
                  Yu Qiao and
                  Jifeng Dai and
                  Wenhai Wang},
  title        = {InternVL3: Exploring Advanced Training and Test-Time Recipes for Open-Source
                  Multimodal Models},
  journal      = {CoRR},
  volume       = {abs/2504.10479},
  year         = {2025},
  url          = {https://doi.org/10.48550/arXiv.2504.10479},
  doi          = {10.48550/ARXIV.2504.10479},
  eprinttype   = {arXiv},
  eprint       = {2504.10479},
  bibsource    = {dblp computer science bibliography, https://dblp.org}
}

@article{bai1others,
  author       = {Shuai Bai and
                  Keqin Chen and
                  Xuejing Liu and
                  Jialin Wang and
                  Wenbin Ge and
                  Sibo Song and
                  Kai Dang and
                  Peng Wang and
                  Shijie Wang and
                  Jun Tang and
                  Humen Zhong and
                  Yuanzhi Zhu and
                  Ming{-}Hsuan Yang and
                  Zhaohai Li and
                  Jianqiang Wan and
                  Pengfei Wang and
                  Wei Ding and
                  Zheren Fu and
                  Yiheng Xu and
                  Jiabo Ye and
                  Xi Zhang and
                  Tianbao Xie and
                  Zesen Cheng and
                  Hang Zhang and
                  Zhibo Yang and
                  Haiyang Xu and
                  Junyang Lin},
  title        = {Qwen2.5-VL Technical Report},
  journal      = {CoRR},
  volume       = {abs/2502.13923},
  year         = {2025},
  url          = {https://doi.org/10.48550/arXiv.2502.13923},
  doi          = {10.48550/ARXIV.2502.13923},
  eprinttype   = {arXiv},
  eprint       = {2502.13923},
  bibsource    = {dblp computer science bibliography, https://dblp.org}
}

@article{paszke2019pytorch,
  author       = {Adam Paszke and
                  Sam Gross and
                  Francisco Massa and
                  Adam Lerer and
                  James Bradbury and
                  Gregory Chanan and
                  Trevor Killeen and
                  Zeming Lin and
                  Natalia Gimelshein and
                  Luca Antiga and
                  Alban Desmaison and
                  Andreas K{\"{o}}pf and
                  Edward Z. Yang and
                  Zachary DeVito and
                  Martin Raison and
                  Alykhan Tejani and
                  Sasank Chilamkurthy and
                  Benoit Steiner and
                  Lu Fang and
                  Junjie Bai and
                  Soumith Chintala},
  editor       = {Hanna M. Wallach and
                  Hugo Larochelle and
                  Alina Beygelzimer and
                  Florence d'Alch{\'{e}}{-}Buc and
                  Emily B. Fox and
                  Roman Garnett},
  title        = {PyTorch: An Imperative Style, High-Performance Deep Learning Library},
  booktitle    = {Advances in Neural Information Processing Systems 32: Annual Conference
                  on Neural Information Processing Systems 2019, NeurIPS 2019, December
                  8-14, 2019, Vancouver, BC, Canada},
  pages        = {8024--8035},
  year         = {2019},
  url          = {https://proceedings.neurips.cc/paper/2019/hash/bdbca288fee7f92f2bfa9f7012727740-Abstract.html},
  bibsource    = {dblp computer science bibliography, https://dblp.org}
}

@inproceedings{ReimersG19,
  author       = {Nils Reimers and
                  Iryna Gurevych},
  editor       = {Kentaro Inui and
                  Jing Jiang and
                  Vincent Ng and
                  Xiaojun Wan},
  title        = {Sentence-BERT: Sentence Embeddings using Siamese BERT-Networks},
  booktitle    = {Proceedings of the 2019 Conference on Empirical Methods in Natural
                  Language Processing and the 9th International Joint Conference on
                  Natural Language Processing, {EMNLP-IJCNLP} 2019, Hong Kong, China,
                  November 3-7, 2019},
  pages        = {3980--3990},
  publisher    = {Association for Computational Linguistics},
  year         = {2019},
  url          = {https://doi.org/10.18653/v1/D19-1410},
  doi          = {10.18653/V1/D19-1410},
  bibsource    = {dblp computer science bibliography, https://dblp.org}
}

@inproceedings{WangYHYMW24,
  author       = {Liang Wang and
                  Nan Yang and
                  Xiaolong Huang and
                  Linjun Yang and
                  Rangan Majumder and
                  Furu Wei},
  editor       = {Lun{-}Wei Ku and
                  Andre Martins and
                  Vivek Srikumar},
  title        = {Improving Text Embeddings with Large Language Models},
  booktitle    = {Proceedings of the 62nd Annual Meeting of the Association for Computational
                  Linguistics (Volume 1: Long Papers), {ACL} 2024, Bangkok, Thailand,
                  August 11-16, 2024},
  pages        = {11897--11916},
  publisher    = {Association for Computational Linguistics},
  year         = {2024},
  url          = {https://doi.org/10.18653/v1/2024.acl-long.642},
  doi          = {10.18653/V1/2024.ACL-LONG.642},
  bibsource    = {dblp computer science bibliography, https://dblp.org}
}

@misc{openai2026gpt54mini,
  author       = {{OpenAI}},
  title        = {{GPT-5.4 mini}},
  year         = {2026},
  howpublished = {\url{https://developers.openai.com/api/docs/models/gpt-5.4-mini}},
}

@misc{openai2024gpt4omini,
  author = {{OpenAI}},
  title = {{GPT‑4o mini}: advancing cost-efficient intelligence},
  year = {2024},
  url = {https://openai.com/index/gpt-4o-mini-advancing-cost-efficient-intelligence/}
}

@misc{openai2026gpt55,
  author = {{OpenAI}},
  title = {Introducing {GPT-5.5}},
  year = {2026},
  url = {https://openai.com/index/introducing-gpt-5-5/}
}

@misc{openai2025gptimage15,
  author = {{OpenAI}},
  title = {The new {ChatGPT Images} is here},
  year = {2025},
  url = {https://openai.com/index/new-chatgpt-images-is-here/}
}

@article{rafailov2023direct,
  title={Direct preference optimization: Your language model is secretly a reward model},
  author={Rafailov, Rafael and Sharma, Archit and Mitchell, Eric and Manning, Christopher D and Ermon, Stefano and Finn, Chelsea},
  journal={Advances in neural information processing systems},
  volume={36},
  pages={53728--53741},
  year={2023}
}

@article{wei2022chain,
  title={Chain-of-thought prompting elicits reasoning in large language models},
  author={Wei, Jason and Wang, Xuezhi and Schuurmans, Dale and Bosma, Maarten and Xia, Fei and Chi, Ed and Le, Quoc V and Zhou, Denny and others},
  journal={Advances in neural information processing systems},
  volume={35},
  pages={24824--24837},
  year={2022}
}

@inproceedings{liu2021visual,
  title={Visual news: Benchmark and challenges in news image captioning},
  author={Liu, Fuxiao and Wang, Yinghan and Wang, Tianlu and Ordonez, Vicente},
  booktitle={Proceedings of the 2021 conference on empirical methods in natural language processing},
  pages={6761--6771},
  year={2021}
}

@inproceedings{tran2020transform,
  title={Transform and tell: Entity-aware news image captioning},
  author={Tran, Alasdair and Mathews, Alexander and Xie, Lexing},
  booktitle={Proceedings of the IEEE/CVF conference on computer vision and pattern recognition},
  pages={13035--13045},
  year={2020}
}

@inproceedings{DBLP:conf/nips/ZhangLZCSLYLZ25,
  author       = {Fanrui Zhang and
                  Dian Li and
                  Qiang Zhang and
                  Jun Chen and
                  Sinbadliu and
                  Junxiong Lin and
                  Jiahong Yan and
                  Jiawei Liu and
                  Zheng{-}Jun Zha},
  editor       = {Danielle Belgrave and
                  Cheng Zhang and
                  Laura N. Montoya and
                  Hsuan{-}Tien Lin and
                  Razvan Pascanu and
                  Piotr Koniusz and
                  Marzyeh Ghassemi and
                  Nancy Chen and
                  Iv{\'{a}}n Vladimir Meza Ru{\'{\i}}z and
                  Arturo Loaiza{-}Bonilla},
  title        = {Fact-R1: Towards Explainable Video Misinformation Detection with Deep
                  Reasoning},
  booktitle    = {Advances in Neural Information Processing Systems 38: Annual Conference
                  on Neural Information Processing Systems 2025, NeurIPS 2025, San Diego,
                  CA, USA, December 2-7, 2025 / Mexico City, Mexico, November 30 - December
                  5, 2025},
  year         = {2025},
  url          = {http://papers.nips.cc/paper\_files/paper/2025/hash/ee5bb72130c332c3d4bf8d231e617506-Abstract-Conference.html},
  bibsource    = {dblp computer science bibliography, https://dblp.org}
}

@article{yang2025qwen3,
  title={Qwen3 technical report},
  author={Yang, An and Li, Anfeng and Yang, Baosong and Zhang, Beichen and Hui, Binyuan and Zheng, Bo and Yu, Bowen and Gao, Chang and Huang, Chengen and Lv, Chenxu and others},
  journal={arXiv preprint arXiv:2505.09388},
  year={2025}
}

@article{hayes2007answering,
  title={Answering the call for a standard reliability measure for coding data},
  author={Hayes, Andrew F and Krippendorff, Klaus},
  journal={Communication methods and measures},
  volume={1},
  number={1},
  pages={77--89},
  year={2007},
  publisher={Taylor \& Francis}
}

\clearpage
\newpage

\appendix

\section{Dataset Details}
\label{appendix:dataset_details}

\subsection{Inconsistency Generation Pipeline}
\label{subsec:dataset_construction_pipeline}

\paragraph{Source corpus.}
We use the 16,000-image New York Times portion of TARA~\citep{fu2022there}, whose explicit time and location information supports the construction of context-dependent inconsistencies and temporally separated data splits.
In contrast, VisualNews~\citep{liu2021visual} and NYTimes800k~\citep{tran2020transform} primarily target news image captioning rather than spatiotemporal grounding.
Starting from these image--claim pairs, we construct the dataset through applicability filtering, target-type assignment, edit prompt generation, and image synthesis.

\paragraph{Applicability filtering.}
For each image--claim pair, we identify which taxonomy types can support a plausible world-knowledge inconsistency. A type $k$ is applicable to $(I_{\text{con}}, c)$ if its associated visual element is clearly visible in $I_{\text{con}}$ and the claim $c$ provides enough context for modifying that element to create an inconsistency. 
We first retain TARA examples with valid image URLs and a width and height of at least 1,024 pixels, yielding 13,507 pairs from the original 16,000 examples.
We then use GPT-4o-mini to screen each image–claim pair for editing suitability. A pair is retained if at least one inconsistency type has both a visible visual element and sufficient context in the claim or accompanying metadata for an edit to create a meaningful inconsistency. Next, GPT-5.5 checks whether the visual elements associated with each of the nine types are clearly visible and records the supporting visual evidence. This check requires direct observation of the image rather than assumptions based on the claim.

\paragraph{Type assignment.}
Each retained image--claim pair may support multiple applicable inconsistency types, while each edited sample is designed to contain only a single controlled inconsistency. We therefore assign exactly one target type $k$ to each pair using a greedy strategy that combines constrained-first ordering with rarest-first selection. Specifically, pairs with fewer applicable types are processed first to prioritize samples with limited assignment options. Within each pair, we then select the applicable type with the smallest accumulated sample count, encouraging a balanced distribution across the inconsistency taxonomy.

\paragraph{Edit prompt design and image generation.}
For each assigned tuple $(I_{\text{con}}, c, k)$, we use GPT-5.5 to generate edit instructions, an initial description of the edited visual element $v_0$, and an initial world-knowledge explanation $e_0$ describing the intended inconsistency. The edit instructions ensure that the resulting image remains visually plausible, that the modified element contradicts the claimed event while remaining coherent with the surrounding visual context, and that detecting the inconsistency requires contextual or world knowledge rather than reliance on obvious synthesis artifacts.
To reduce substitution bias, we use rotating geographic and contextual constraints instead of a fixed set of edited-in visual elements.
We generate edited images $I_{\text{inc}}$ using GPT-Image-1.5. For cross-generator evaluation, we additionally construct a FLUX.2-klein-9B~\cite{blackforestlabs2026flux2klein} test set using the same edit instructions and human verification procedure.
The resulting candidate tuples $(I_{\text{con}}, I_{\text{inc}}, c, k, v_0, e_0)$ are then passed to human annotation. All prompts are provided in Appendix~\ref{appendix:prompts}.

\subsection{Human Annotation}
\label{subsec:human_annotation}

\paragraph{Human verification.}
To ensure reliable evaluation, each test-set candidate tuple $(I_{\text{con}}, I_{\text{inc}}, c)$ and its preliminary annotations $(k, v_0, e_0)$ are independently reviewed by two human annotators.
Annotators evaluate three criteria: (\emph{i})~the visual realism of $I_{\text{inc}}$, (\emph{ii})~the fidelity of the edit to the intended modification $v_0$ while preserving non-target regions, and (\emph{iii})~the factual correctness of $e_0$ as a world-knowledge explanation of the inconsistency.
To assess factual correctness, annotators verify the key factual claims in $e_0$ against independent external evidence, prioritizing official sources and established fact-checking pages, with Wikipedia as a supplementary source.
We retain a candidate only when both annotators approve all criteria and discard all cases with conflicting judgments.
This verification process removes 132 candidate samples in total.
For retained examples, annotators further verify and revise the preliminary annotations to obtain the final structured annotations $(k, v, e)$.

\paragraph{Reasoning process generation.}
We further augment each candidate with Chain-of-Thought (CoT) rationales generated by GPT-5.5. The inconsistent rationale $r_{\text{inc},0}$ explains why $I_{\text{inc}}$ contradicts the claim $c$, while the consistent rationale $r_{\text{con},0}$ explains why $I_{\text{con}}$ remains consistent with the claim. Human annotators revise both rationales to ensure that they are grounded in the image and aligned with the final annotations $(k, v, e)$. 
The final \benchmark{} dataset contains 4,406 claims, each paired with a consistent source image $I_{\text{con}}$ and an inconsistent edited image $I_{\text{inc}}$.
Each claim also includes structured annotations $(k, v, e)$ for the inconsistent counterpart and rationales $(r_{\text{con}}, r_{\text{inc}})$ for both image versions.
Together, these yield 8,812 image--claim instances.

\subsection{Dataset Statistics}
\label{subsec:data_statistic}

\begin{table*}[!t]
\centering
\small
\resizebox{\textwidth}{!}{%
\begin{tabular}{@{}lrrrl@{}}
\toprule
\textbf{Stage} &
\textbf{Input} &
\textbf{Retained} &
\textbf{Removed} &
\textbf{Criterion} \\
\midrule
Source-image URL and resolution filtering
& 16,000
& 13,507
& 2,493
& Valid URL and resolution $\geq 1024 \times 1024$ \\

Sample screening for editing (GPT-4o-mini)
& 13,507
& 10,301
& 3,206
& At least one feasible inconsistency type \\

Visual element visibility check (GPT-5.5)
& 10,301
& 5,984
& 4,317
& Clearly visible target without caption-only inference \\

Edit generation and post-generation filtering
& 5,984
& 5,818
& 166
& Successful generation passing moderation and size checks \\

Human verification (test subset)
& 1,533
& 1,401
& 132
& Both annotators approve realism, edit fidelity, and the factual correctness of the explanation \\

Final split construction
& 5,818
& 4,406
& 1,412
& Temporal and entity-disjoint splits with the verified test set \\
\bottomrule
\end{tabular}%
}
\caption{Filtering statistics and decision criteria for the \benchmark{} construction pipeline.}
\label{tab:construction_funnel}
\end{table*}

\paragraph{Filtering statistic.}
Table~\ref{tab:construction_funnel} reports the number of pairs retained and removed at each construction stage together with the corresponding decision criteria.
The first four stages reduce 16,000 source pairs to 5,818 successfully generated candidates.
Human verification is applied separately to the 1,533 candidate test examples, retaining 1,401, while final split construction yields 4,406 claims.

\paragraph{Dataset composition.}
Table~\ref{tab:split_stats} summarizes the four data splits.
Each claim contributes two image--claim instances: the original image paired with the claim and labeled as \textsc{consistent}, and an edited image paired with the same claim and labeled as \textsc{inconsistent}.
Consequently, the 4,406 claims in \benchmark{} produce 4,406 consistent instances and 4,406 inconsistent instances, yielding a total of 8,812 image--claim instances.
Table~\ref{tab:data_stats} further reports the per-type statistics for the inconsistent portion of the dataset.
The nine inconsistency categories are kept approximately balanced, with per-type counts in the training set ranging from 244 to 315 samples.

\paragraph{Split construction.}
We construct the dataset splits using both temporal and replacement-entity separation.
For temporal separation, the training and validation sets contain claims from 2010 to 2017, while both test splits contain claims from 2018 to 2021.
For replacement-entity separation, the in-distribution test split uses edited-in visual elements observed during training but in unseen image contexts, whereas the out-of-distribution split uses a disjoint set of held-out replacement entities, such as previously unseen country flags or brand logos.
This setup disentangles generalization to novel image contexts from generalization to unseen semantic substitutions.

\paragraph{Introduced-element statistics.}
Table~\ref{tab:introduced_element_statistics} reports the distribution of introduced visual elements for each inconsistency type at the canonical entity level.
The distribution is long-tailed, with no single element accounting for more than $12.8\%$ of its type.
Across all types, the out-of-distribution split holds out 426 globally unique replacement entities, none of which appears in training.

\subsection{Inter-Annotator Agreement}
\label{subsec:iaa}
We compute inter-annotator agreement over the initial binary keep-or-reject decisions for the double-annotated test-set candidates.
The two annotators agree on 97.4\% of candidates, with Krippendorff's $\alpha=0.81$.
Only candidates approved by both annotators are retained, and their structured annotations are subsequently finalized.

\begin{table}[!t]
\centering
\small
\resizebox{\columnwidth}{!}{%
\begin{tabular}{lrrrr}
\toprule
\textbf{Split} & \textbf{\#Claims} & \textbf{\#Consistent} & \textbf{\#Inconsistent} & \textbf{Total} \\
\midrule
Train      & 2,561 & 2,561 & 2,561 & 5,122 \\
Validation        &    444  &    444  &    444  &    888  \\
Test (In-distribution)   &    703  &    703  &    703  & 1,406 \\
Test (Out-of-Distribution)  &    698  &    698  &    698  & 1,396 \\
\midrule
Total      & 4,406 & 4,406 & 4,406 & 8,812 \\
\bottomrule
\end{tabular}%
}
\caption{Dataset split statistics for \benchmark{}.}
\label{tab:split_stats}
\end{table}

\begin{table}[!t]
  \centering
  \small
  \resizebox{\columnwidth}{!}{%
  \begin{tabular}{lrrrr}
  \toprule
  \textbf{Inconsistency type} & \textbf{Train} & \textbf{Validation} & \textbf{Test (In-distribution)} & \textbf{Test (Out-of-Distribution)} \\
  \midrule
  clothing       & 295 & 59 &  68 &  73 \\
  flag           & 244 & 45 & 100 &  97 \\
  gesture        & 263 & 35 &  79 &  77 \\
  signage        & 263 & 52 &  88 &  89 \\
  architecture   & 293 & 59 &  80 &  77 \\
  infrastructure & 301 & 56 &  75 &  64 \\
  technology     & 315 & 44 &  60 &  71 \\
  branding       & 309 & 53 &  74 &  75 \\
  environment    & 278 & 41 &  79 &  75 \\
  \midrule
  Total          & 2561 & 444 & 703 & 698 \\
  \bottomrule
  \end{tabular}%
  }
  \caption{Distribution of inconsistency types in \benchmark{} across data splits.}
  \label{tab:data_stats}
\end{table}

\begin{table}[!t]
\centering
\resizebox{\columnwidth}{!}{%
\begin{tabular}{@{}lrrlr@{}}
\toprule
\textbf{Type} &
\textbf{Edits} &
\shortstack{\textbf{Distinct}\\\textbf{elements}} &
\shortstack{\textbf{Most frequent}\\\textbf{element (share)}} &
\shortstack{\textbf{OOD held-out}\\\textbf{entities}} \\
\midrule
Clothing       & 495 & 189 & Pakistani shalwar kameez ($3.8\%$) & 59 \\
Flag           & 486 & 48  & Chinese flag ($12.3\%$)            & 17 \\
Gesture        & 454 & 48  & Indian namaste ($12.8\%$)          & 22 \\
Signage        & 492 & 99  & Spanish text ($7.1\%$)             & 36 \\
Architecture   & 509 & 175 & Mughal architecture ($8.8\%$)      & 57 \\
Infrastructure & 496 & 245 & Pakistani motorway sign ($6.9\%$)  & 60 \\
Technology     & 490 & 338 & BMC Kirpi ($2.0\%$)                 & 61 \\
Branding       & 511 & 203 & Blockbuster Video ($8.2\%$)         & 67 \\
Environment    & 473 & 99  & Scandinavian birch ($12.1\%$)      & 48 \\
\bottomrule
\end{tabular}%
}
\caption{Introduced-element statistics at the canonical entity level for each inconsistency type.}
\label{tab:introduced_element_statistics}
\end{table}

\begin{table*}[!t]
\centering
\small
\resizebox{\textwidth}{!}{%
\begin{tabular}{@{}llllll@{}}
\toprule
\textbf{Benchmark} &
\textbf{Primary task} &
\textbf{Misinformation input} &
\textbf{Inconsistency structure} &
\textbf{Controlled edit} &
\textbf{Inconsistency annotations} \\
\midrule

MiRAGeNews
& Real vs.\ generated news
& Fictional claim + AI-generated image
& No controlled image--claim inconsistency
& No
& No \\

OOC benchmarks
& Consistent vs.\ inconsistent
& Different event/context + authentic image
& Broad contextual mismatch
& No
& No \\

\textbf{MIC-Bench}
& Consistency and explanation
& Same real-event claim + AI-generated image
& Localized inconsistency and broader event preserved
& Yes
& 9 types, visual evidence, and world-knowledge explanation \\

\bottomrule
\end{tabular}%
}
\caption{Comparison of \benchmark{} with related multimodal misinformation benchmarks.}
\label{tab:benchmark_comparison}
\end{table*}

\subsection{Benchmark Comparison}
\label{appendix:benchmark_comparison}
Table~\ref{tab:benchmark_comparison} summarizes the differences between \benchmark{} and related multimodal misinformation benchmarks.
MiRAGeNews pairs a fictional claim with an AI-generated image and frames the task as distinguishing real from generated image--caption pairs~\citep{huang2024miragenews}.
As the image is generated to depict the fictional claim it accompanies, the pair contains no controlled image–claim inconsistency. 
The misinformation lies in the fabrication of the pair as a whole rather than in a conflict between an AI-generated image and a real-event claim, and inconsistencies with world knowledge arise only incidentally during generation.
OOC benchmarks instead pair an authentic image with a claim that describes a different event, location, or time~\citep{qi2024sniffer,tonglet-etal-2025-cove}.
Although the image and real-event claim may share an entity or topic, the image originates from a different context, so the resulting mismatch is global rather than localized to a specific visual element. 
Neither setting provides an authentic source image together with a controlled edited counterpart that isolates the visual element responsible for the inconsistency, and neither annotates the inconsistency type, the supporting visual evidence, or the world-knowledge explanation.

\subsection{Examples by Inconsistency Types}
\label{subsec:type_examples}

Table~\ref{tab:type_visual_examples} presents one representative example for each inconsistency type in \benchmark{}. For each category, the table shows the original image and its corresponding AI-generated counterpart, together with the associated news claim. It additionally highlights the manipulated visual element responsible for the inconsistency and provides the corresponding world-knowledge explanation describing why the edited content contradicts the claimed geopolitical, cultural, temporal, or contextual setting. These examples illustrate the diversity of inconsistency patterns covered by \benchmark{} and demonstrate how visually plausible inconsistencies can create subtle yet semantically misleading image--claim pairs.

\newcommand{\typeexample}[5]{%
#1 &
\includegraphics[width=\linewidth]{figs/type_examples/#2.png} &
\includegraphics[width=\linewidth]{figs/type_examples/#2_edited.png} &
#3 &
#4 &
#5
}

\begin{table*}[!t]
\centering
\small
\renewcommand{\arraystretch}{1.12}
\resizebox{\textwidth}{!}{%
\begin{tabular}{
>{\raggedright\arraybackslash}m{2.1cm}
>{\centering\arraybackslash}m{3.1cm}
>{\centering\arraybackslash}m{3.1cm}
>{\raggedright\arraybackslash}m{4.7cm}
>{\raggedright\arraybackslash}m{4.7cm}
>{\raggedright\arraybackslash}m{5.6cm}
}
\toprule
\textbf{Inconsistent Type} & \textbf{Original Image} & \textbf{Edited Image} & \textbf{Claim} & \textbf{What Changed} & \textbf{Why Contradicts} \\
\midrule
\typeexample{clothing}{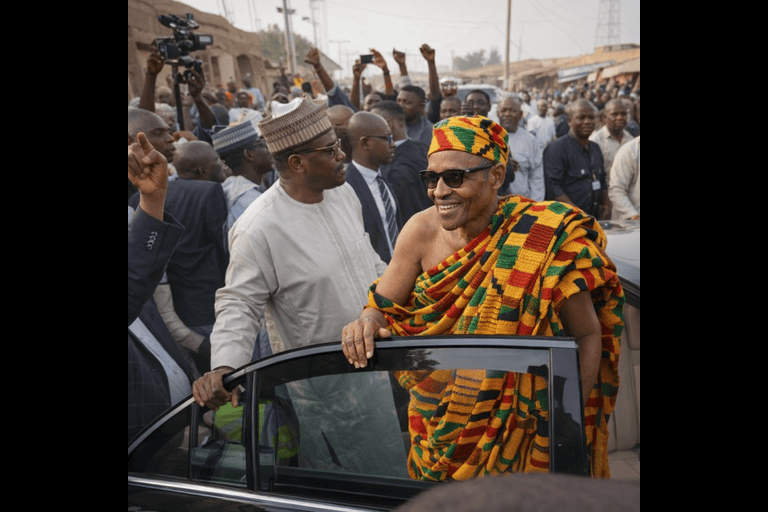}
{Muhammadu Buhari Wins Second Term as Nigeria's President. President Muhammadu Buhari of Nigeria after casting his vote in his hometown Daura on Saturday.}
{Buhari's light blue Nigerian traditional robe and patterned cap $\rightarrow$ Ghanaian Ashanti kente ceremonial garment.}
{The caption identifies Nigeria's president voting in his hometown, so dressing him in distinctly Ghanaian ceremonial clothing conflicts with the Nigerian electoral context.}\\
\midrule
\typeexample{flag}{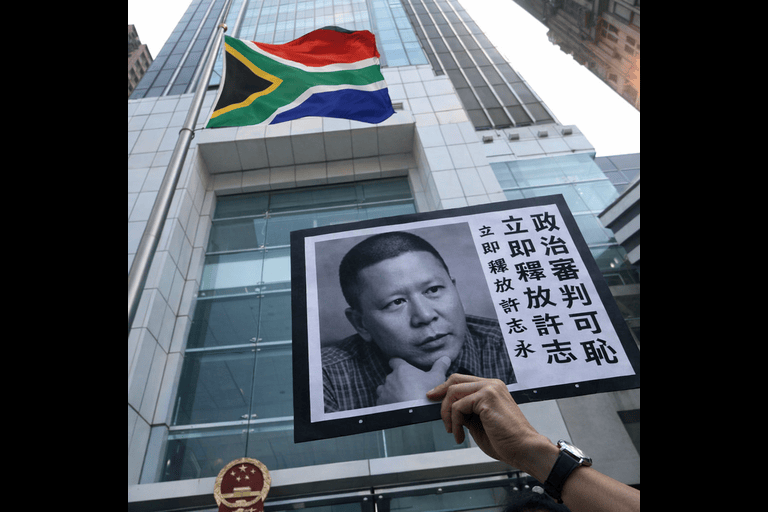}
{`Suffering and Hardship Belong to Me': A Voice From a Chinese Prison. A protester holding a photo of Xu Zhiyong outside the Chinese liaison office in Hong Kong in 2014.}
{Chinese five-star red flag $\rightarrow$ South African multicolor Y-design flag.}
{The caption identifies the scene as outside the Chinese liaison office in Hong Kong in 2014, where a Chinese national flag would be contextually expected, not South Africa's flag.}\\
\midrule
\typeexample{gesture}{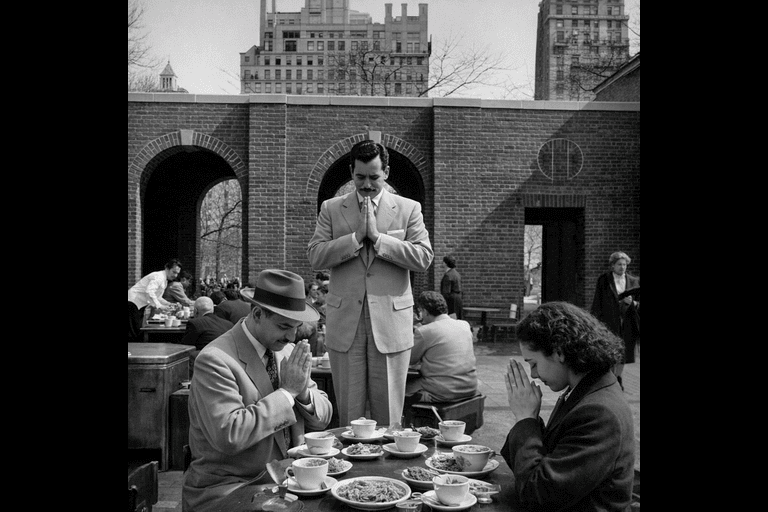}
{Retracing the Photographic Steps of a 1951 New York City Shoot. Central Park. April 2, 1951.}
{Man filming communal meal $\rightarrow$ man and nearby diners exchanging Indian namaste greeting.}
{A formal South Asian namaste-style greeting changes the scene from a 1951 Central Park social dining gathering into a culturally different ceremonial interaction that does not fit the caption's New York context.}\\
\midrule
\typeexample{signage}{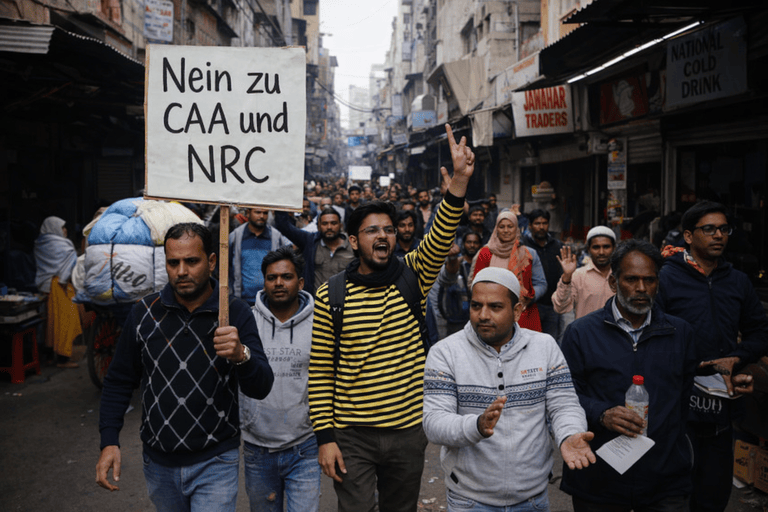}
{Gandhi Biographer Arrested as Protests Over Citizenship Law Sweep India. Protesters denouncing a new citizenship law near the historic Red Fort in New Delhi on Thursday.}
{English protest placard text ``No CAA. \& NRC.'' $\rightarrow$ German placard text ``Nein zu CAA und NRC''.}
{A German-language anti-CAA/NRC protest sign is inconsistent with a street protest described as taking place near the Red Fort in New Delhi, where such signage would ordinarily be in English, Hindi, or Urdu rather than German.}\\
\midrule
\typeexample{architecture}{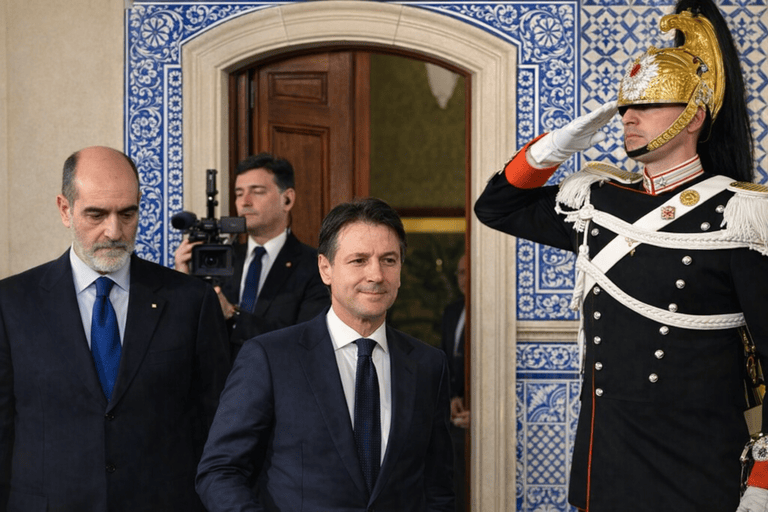}
{Italy's Populist Parties, on Precipice of Power, Fail to Form Government. Italy's prime minister-designate, Giuseppe Conte, center, leaving a meeting with Italy's president, Sergio Mattarella, on Sunday in Rome.}
{Italian official interior doorway/wall detailing $\rightarrow$ Brazilian Portuguese-colonial blue-and-white azulejo-tiled doorway surround with shallow colonial arch.}
{The caption places the scene in Rome at Italy's presidential residence, but the edited architecture implies a Brazilian Portuguese-colonial governmental interior instead of an Italian one.}\\
\midrule
\typeexample{infrastructure}{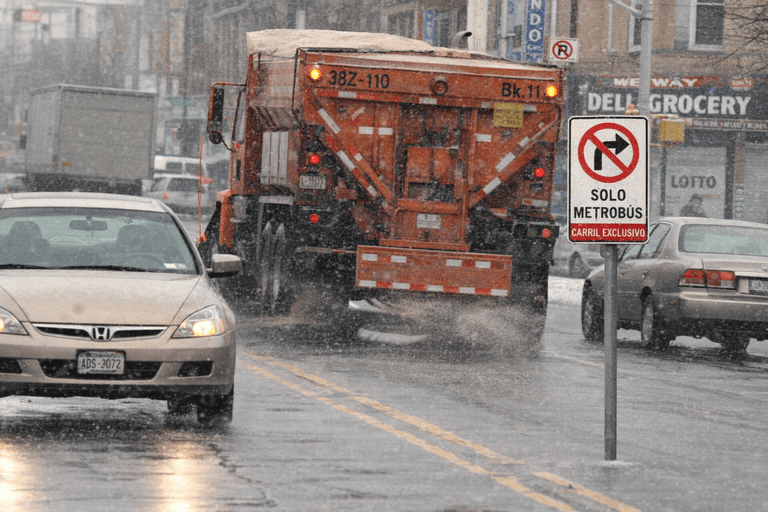}
{Snowstorm That Wasn't Finds City Well Prepared. A city truck spread salt on 18th Avenue in Brooklyn on Friday.}
{Brooklyn U.S. parking regulation sign $\rightarrow$ Mexico City Metrobus-style Spanish transit-lane sign.}
{A Mexican transit-corridor sign is inconsistent with a caption placing the scene on 18th Avenue in Brooklyn, New York City.}\\
\midrule
\typeexample{technology}{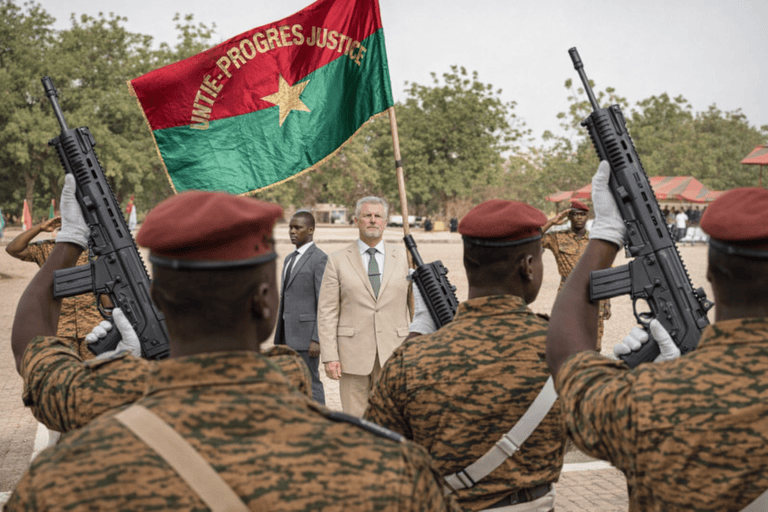}
{When the Face of America Falls Ill: A Virus's Toll on Diplomats. Andrew Young, the American ambassador to Burkina Faso, during the opening ceremony of a joint military exercise last year. Mr. Young tested positive for the coronavirus.}
{Ceremonial rifles with fixed bayonets $\rightarrow$ Italian Beretta ARX160 assault rifles.}
{The caption describes the opening ceremony of a joint military exercise in Burkina Faso, but equipping the ceremonial guard with modern Italian Beretta ARX160 rifles implies a different military sourcing pattern and ceremonial practice than the Burkinab\'e setting shown.}\\
\midrule
\typeexample{branding}{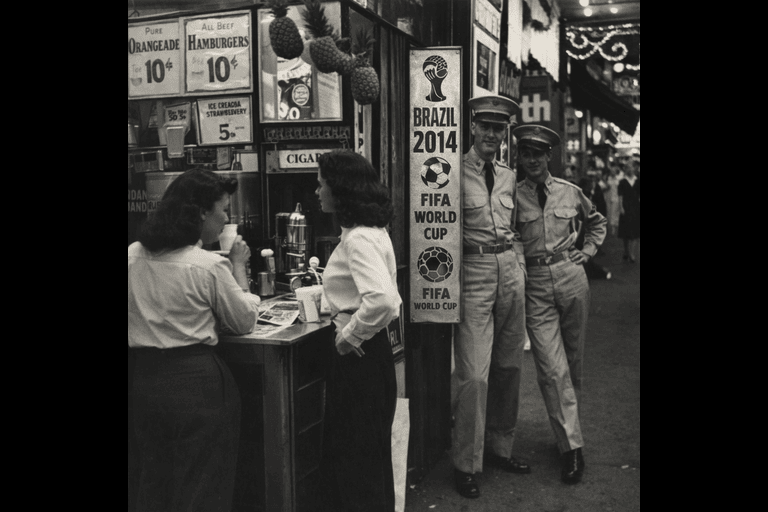}
{The Radical Empathy of Dan Weiner. Broadway, New York City, 1951.}
{Vertical ``PIZZA'' food sign $\rightarrow$ 2014 FIFA World Cup ``BRAZIL 2014'' promotional poster.}
{A 2014 FIFA World Cup advertisement is impossible in a street scene captioned as Broadway, New York City, 1951.}\\
\midrule
\typeexample{environment}{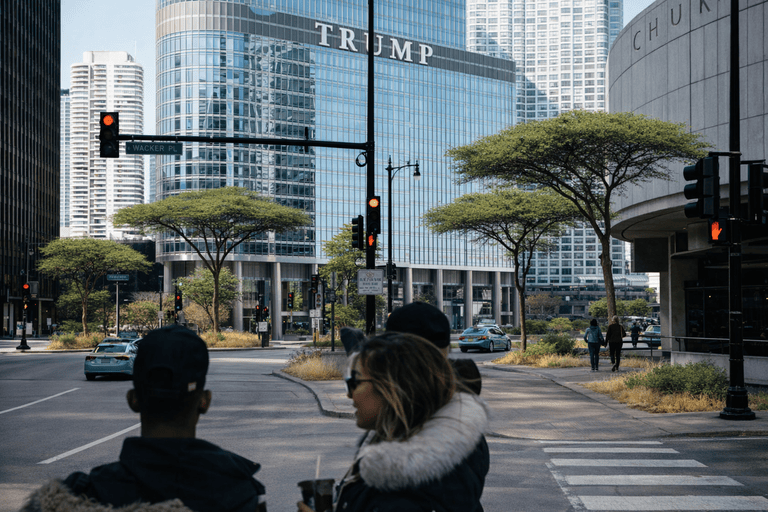}
{Trump's Financial Disclosure Form: Five Takeaways. The Trump International Hotel and Tower in Chicago, where profits declined last year.}
{Leafless Chicago winter street trees and dormant planting $\rightarrow$ Sub-Saharan savanna acacia trees with dry golden grass and scrub.}
{The caption identifies the scene as the Trump International Hotel and Tower in Chicago, but savanna acacia landscaping and dry tropical vegetation imply a Sub-Saharan African climate rather than a cold-season Midwestern U.S. city.}\\
\bottomrule
\end{tabular}%
}
\caption{Representative visual examples for all nine inconsistency types in \benchmark{}.}
\label{tab:type_visual_examples}
\end{table*}

\section{Experimental Setup}
\label{sec:experimental_setup}

\subsection{Baseline Hyperparameters}
\label{subsec:baseline_hyperparameters}

This section summarizes the hyperparameter settings used for all baseline methods.
Unless otherwise specified, all prompted VLM baselines use the same zero-shot Chain-of-Thought (CoT) instruction~\cite{wei2022chain} as \ours{}.

\paragraph{CLIP.}~\citep{radford2021learning}
We use \texttt{openai/clip-vit-large-patch14} to encode image--claim pairs and compute image--text cosine similarity scores. Verdicts are predicted using a threshold selected on the validation split to maximize F1, where pairs with similarity scores below the threshold are classified as \textsc{inconsistent}.

\paragraph{SNIFFER.}~\citep{qi2024sniffer}
We use the official SNIFFER checkpoint with Vicuna-13B-v1.1 as the language backbone.

\paragraph{MiRAGe.}~\citep{huang2024miragenews}
We use the official MiRAGe-Img checkpoint, which combines a frozen BLIP-2-Flan-T5-XL vision encoder with a learned CBM-style concept layer concatenated with a CLIP-based linear logit, yielding a 301-dimensional feature passed to a linear classification head with sigmoid activation. 
We adopt threshold of $0.639$, classifying images with scores above the threshold as \textsc{inconsistent}.

\paragraph{GPT-5.4-mini.}
For GPT-5.4-mini, we set the maximum completion length to 4,096 tokens. Since GPT-5.4-mini is a reasoning-oriented model, no decoding temperature is specified.

\paragraph{Qwen and Other Open-Source VLMs.}
For Qwen2.5-VL-3B-Instruct~\citep{bai1others}, Qwen3-VL-4B-Instruct, Qwen3-VL-8B-Instruct, Qwen3-VL-30B-A3B~\citep{bai2025qwen3}, LLaVA-OneVision-7B~\citep{li2024llava}, and InternVL3-8B~\citep{zhu2025internvl3},
we use greedy decoding with a maximum generation length of 8,192 tokens.

For Qwen3-VL-8B-Thinking~\citep{bai2025qwen3}, we use a maximum context length of 65,536 tokens and a maximum generation length of 32,768 tokens. Decoding hyperparameters are set to temperature $1.0$, top-$p$ $0.95$, top-$k$ $20$, and presence penalty $1.5$.

\subsection{Implementation Details}
\label{app:experimental_setup}

All training is conducted in PyTorch~\citep{paszke2019pytorch} on $8\times$ AMD MI210 GPUs, each equipped with 64\,GB of VRAM. We use Qwen3-VL-4B-Instruct~\citep{bai2025qwen3} as the backbone model.
LoRA adapters are inserted into the linear layers of the vision encoder, the vision--language projector, and the language model, while all pretrained model weights remain frozen. The adapters use rank $r=64$, scaling factor $\alpha=128$, and dropout rate $0.05$.
For SFT, we train for 3 epochs using AdamW with a peak learning rate of $2\times10^{-4}$, cosine learning-rate scheduling, and a warmup ratio of $0.10$. Training uses a batch size of 8.
For GRPO training, optimization is initialized from the merged SFT checkpoint. We train for 3 epochs with a learning rate of $1\times10^{-5}$, training batch size 8, and 8 sampled responses per prompt. 
The reward function follows Section~\ref{subsec:reward_function_design} and includes rewards for format validity, verdict exact match, inconsistency type exact match, visual-description similarity, and explanation similarity. The corresponding weights are set to $\lambda_{\text{fmt}}=0.10$, $\lambda_{\text{ver}}=0.35$, $\lambda_{\text{type}}=0.25$, $\lambda_{\text{desc}}=0.15$, and $\lambda_{\text{exp}}=0.15$. The semantic similarity rewards are computed using Qwen3-Embedding-0.6B.
The KL regularization coefficient is set to $0.01$. 
DPO is initialized from the same SFT checkpoint as GRPO, with a frozen copy serving as the reference model.
Using the same reward, we pair the highest- and lowest-scoring of eight SFT responses per instance, retaining 3,000 preference pairs after filtering.
We update all language-model parameters while freezing the vision encoder and vision--language projector, using sigmoid DPO with $\beta=0.1$, AdamW with learning rate $10^{-6}$, and an effective batch size of 8.
For FLUX.2-klein-9B image generation, we use 28 inference steps.

\section{Additional Experimental Results}
\label{sec:additional_experimental_results}

\begin{figure}[!t]
    \centering
    \includegraphics[width=0.99\linewidth]{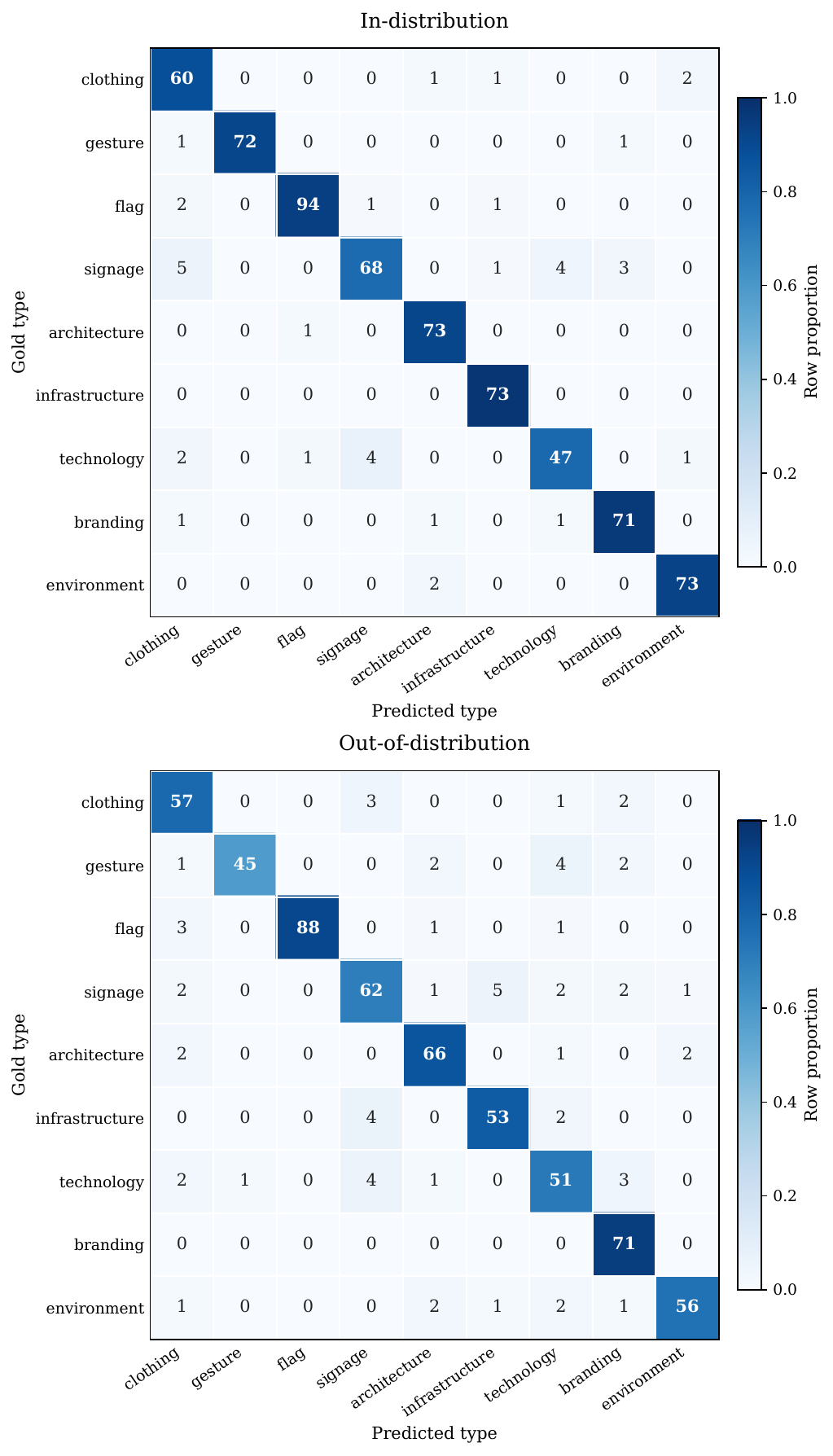}
    \caption{Type-level confusion matrices of \ours{} on the in-distribution and out-of-distribution test splits.}
    \label{fig:type_confusion}
\end{figure}

\subsection{Fine-Grained Type Confusion Analysis}
\label{subsec:typeacc_confusion_matrix}

To assess whether the nine inconsistency categories are reliably distinguishable, we report type-level confusion matrices for both test splits in Figure~\ref{fig:type_confusion}. Both matrices exhibit strong diagonal dominance, indicating that \ours{} can accurately discriminate among the nine inconsistency types in most cases.

Importantly, boundary cases discussed in Section~\ref{subsec:source_data_and_taxonomy} are well separated. For example, confusion between \textit{signage} and \textit{branding} occurs only 3 times in the in-distribution split and 2 times in the out-of-distribution split, suggesting that our taxonomy design and disambiguation rules produce clean and semantically coherent category boundaries.

The remaining off-diagonal errors are sparse rather than systematic, and are mainly concentrated in context-dependent categories such as \textit{signage} and \textit{technology}. This behavior is expected, as these categories often require fine-grained contextual knowledge and are therefore more sensitive to unseen edited-in visual elements. Overall, these results demonstrate that \ours{} captures fine-grained distinctions across all nine inconsistency categories, providing empirical support for the effectiveness of our inconsistency taxonomy.

\begin{figure}[!t]
    \centering
    \includegraphics[width=0.99\linewidth]{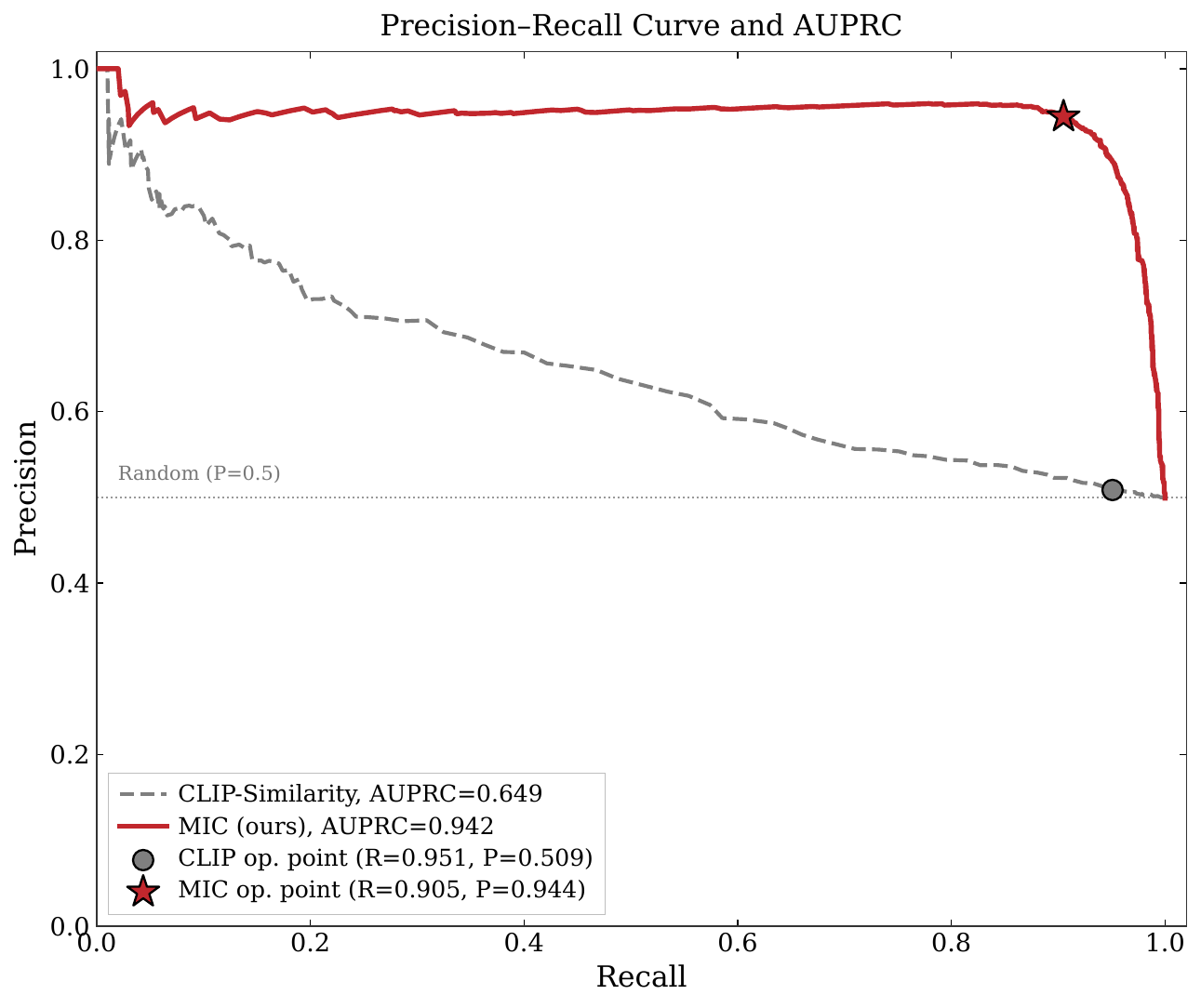}
        \caption{Precision-Recall curves and AUPRC of \ours{} and CLIP on the combined in-distribution and out-of-distribution test set.}
    \label{fig:auprc}
\end{figure}

\begin{table}[!t]
  \centering
  \resizebox{0.99\columnwidth}{!}{%
  \begin{tabular}{l rr rr}
  \toprule
  & \multicolumn{2}{c}{\textbf{In-Distribution}} & \multicolumn{2}{c}{\textbf{Out-of-Distribution}} \\
  \cmidrule(lr){2-3} \cmidrule(lr){4-5}
  \textbf{Model} & \textbf{VisualSim} & \textbf{ExplSim} & \textbf{VisualSim} & \textbf{ExplSim} \\
  \midrule
  \multicolumn{5}{l}{\textit{/* Qwen3-Embedding-0.6B */}} \\
  GPT-5.4-mini   & 20.14 & 17.88 & 19.43 & 17.02 \\
  Qwen3-VL-4B-Instruct    & 23.16 & 25.24 & 24.18 & 27.18 \\
  \ours{}        & \textbf{83.04} & \textbf{82.54} & \textbf{72.09} & \textbf{75.78} \\
  \midrule
  \multicolumn{5}{l}{\textit{/* all-mpnet-base-v2 */}} \\
  GPT-5.4-mini   & 14.06 & 15.73 & 12.97 & 14.60 \\
  Qwen3-VL-4B-Instruct    & 15.44 & 23.42 & 15.54 & 24.81 \\
  \ours{}        & \textbf{69.70} & \textbf{74.70} & \textbf{58.99} & \textbf{67.39} \\
  \midrule
  \multicolumn{5}{l}{\textit{/* e5-mistral-7b-instruct */}} \\
  GPT-5.4-mini   & 18.88 & 19.44 & 17.59 & 18.19 \\
  Qwen3-VL-4B-Instruct    & 25.38 & 30.89 & 26.65 & 32.63 \\
  \ours{}        & \textbf{81.97} & \textbf{82.67} & \textbf{71.38} & \textbf{72.67} \\
  \bottomrule
  \end{tabular}%
  }
  \caption{Robustness of \emph{VisualSim} and \emph{ExplSim} to the choice of evaluation embedding model.
  The best performance per dataset per metric is marked in boldface.}
  \label{tab:embedding_robustness}
\end{table}

\subsection{AUPRC Analysis}
\label{subsec:auprc}

To assess whether the verdict performance of \ours{} depends on a specific decision threshold, we report the area under the precision--recall curve (AUPRC) on the combined in-distribution and out-of-distribution test sets for the two methods for which we extract calibrated continuous confidence scores: CLIP and \ours{}. 
For CLIP, we use the image--text cosine similarity score as the prediction confidence. For \ours{}, we use the softmax probability assigned to the \textsc{inconsistent} label token at the \texttt{<verdict>} position, normalized over the \textsc{consistent} and \textsc{inconsistent} label tokens.
As shown in Figure~\ref{fig:auprc}, \ours{} achieves an AUPRC of $94.20\%$, substantially outperforming the $64.90\%$ obtained by CLIP. This result indicates that the precision--recall advantage of \ours{} is maintained consistently across a wide range of decision thresholds, rather than depending on a single operating point.
At their default operating thresholds, \ours{} achieves a precision of $94.40\%$ and a recall of $90.50\%$, whereas CLIP attains a recall of $95.10\%$ at a substantially lower precision of only $50.90\%$. Overall, these findings demonstrate that the verdict improvements of \ours{} reflect robust threshold-invariant gains in classification quality.

\begin{figure}[!t]
    \centering
    \includegraphics[width=0.99\linewidth]{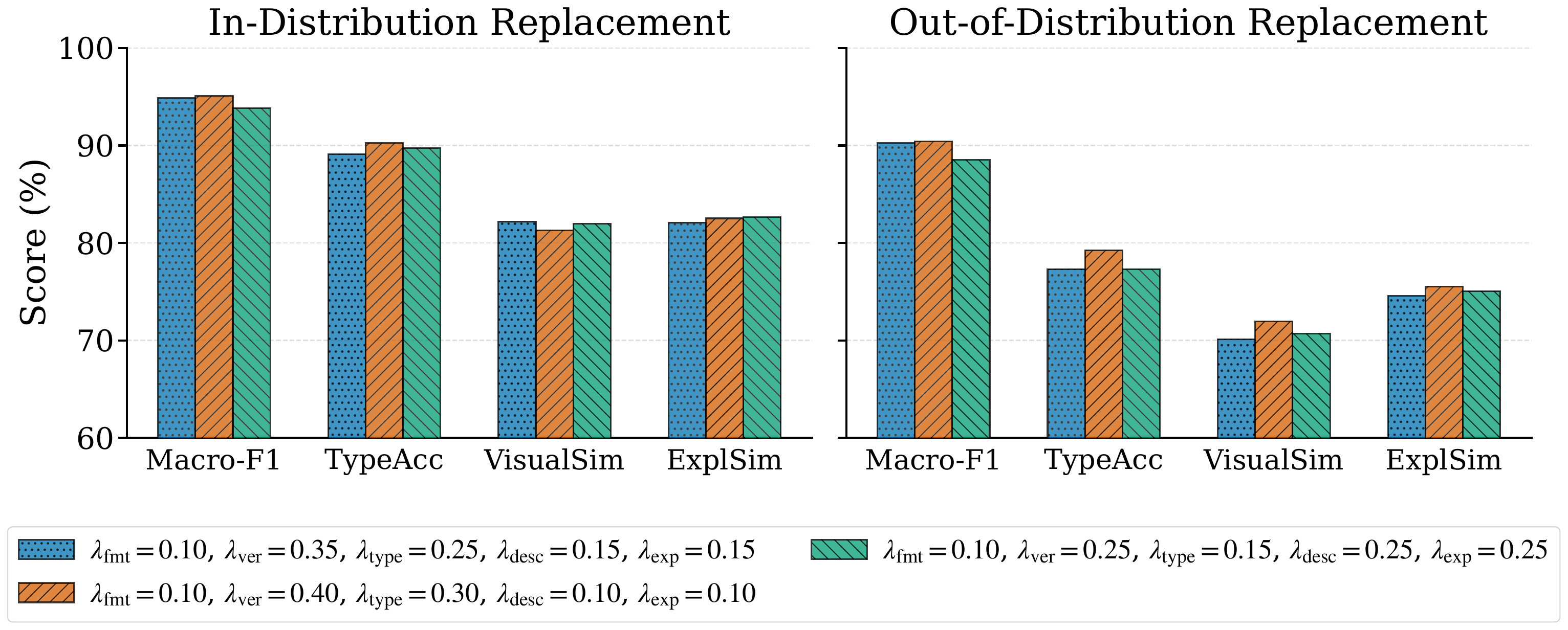}
    \caption{Impact of reward weight on \ours{}.}
    \label{fig:reward_sensitivity}
\end{figure}

\subsection{Reward Weight Sensitivity Analysis}
\label{subsec:lambda_sensitivity}

To investigate the sensitivity of \ours{} to the GRPO reward coefficients, we train the model from the same SFT checkpoint under three reward-weight configurations, while keeping the backbone, training data, and optimization schedule fixed.
The first configuration follows the default reward weights reported in Appendix~\ref{app:experimental_setup}.
The second configuration assigns larger weights to the accuracy-based rewards, namely verdict prediction and inconsistency type classification.
The third configuration assigns larger weights to the semantic rewards, namely visual evidence description and world-knowledge explanation.
As shown in Figure~\ref{fig:reward_sensitivity}, \ours{} exhibits a similar overall performance pattern across the three configurations on both in-distribution and out-of-distribution replacement splits.
Although individual metrics fluctuate as the reward weights change, no single configuration consistently outperforms the others across all evaluation metrics.
These results suggest that the performance of \ours{} is relatively robust to moderate changes in the reward-weight configuration.

\subsection{Robustness Across Embedding Models}
\label{subsec:alt_embedding}

Since \ours{} uses Qwen3-Embedding-0.6B both as the semantic reward signal during GRPO training and as the evaluation model for \emph{VisualSim} and \emph{ExplSim}, we investigate whether the observed gains generalize across different embedding families. To this end, we recompute \emph{VisualSim} and \emph{ExplSim} using two additional embedding models: \texttt{all-mpnet-base-v2}~\cite{ReimersG19} and \texttt{e5-mistral-7b-instruct}~\cite{WangYHYMW24}.
As shown in Table~\ref{tab:embedding_robustness}, \ours{} consistently outperforms representative baselines across all three embedding models on both test splits.
These results show that the gains in reference-based semantic similarity persist across the three evaluated embedding models.

\begin{table*}[!t]
    \centering
    \resizebox{0.85\textwidth}{!}{%
    \begin{tabular}{l rrrr rrrr}
    \toprule
    & \multicolumn{4}{c}{\textbf{In-Distribution Replacement}}
    & \multicolumn{4}{c}{\textbf{Out-of-Distribution Replacement}} \\
    \cmidrule(lr){2-5} \cmidrule(lr){6-9}
    \textbf{Model}
    & \textbf{VisualSim} & \textbf{ExplSim}
    & \textbf{VisualJudge} & \textbf{ExplJudge}
    & \textbf{VisualSim} & \textbf{ExplSim}
    & \textbf{VisualJudge} & \textbf{ExplJudge} \\
    \midrule

    SNIFFER
    & -- & 10.66 & -- & 3.50
    & -- & 8.49 & -- & 2.32 \\

    GPT-5.4-mini
    & 20.14 & 17.88 & 21.05 & 20.74
    & 19.43 & 17.02 & 19.89 & 19.71 \\

    Qwen2.5-VL-3B-Instruct
    & 13.83 & 29.11 & 10.87 & 9.87
    & 11.83 & 28.08 & 8.60 & 8.22 \\

    Qwen3-VL-4B-Instruct
    & 23.16 & 25.24 & 23.53 & 22.84
    & 24.18 & 27.18 & 25.42 & 24.21 \\

    Qwen3-VL-8B-Instruct
    & 26.74 & 28.55 & 27.68 & 26.60
    & 25.13 & 26.88 & 25.16 & 25.21 \\

    Qwen3-VL-30B-A3B
    & 37.95 & 37.42 & 38.35 & 37.58
    & 38.45 & 36.61 & 38.40 & 36.48 \\

    LLaVA-OneVision-7B
    & 13.97 & 18.65 & 8.73 & 8.39
    & 10.43 & 16.45 & 7.39 & 6.25 \\

    InternVL3-8B
    & 18.24 & 22.65 & 17.38 & 17.75
    & 17.82 & 22.51 & 17.05 & 17.02 \\

    Qwen3-VL-8B-Thinking
    & 43.41 & 46.19 & 42.30 & 41.42
    & 42.64 & 46.28 & 41.72 & 40.89 \\

    \midrule

    \ours{}
    & \textbf{83.04} & \textbf{82.54}
    & \textbf{81.02} & \textbf{85.29}
    & \textbf{72.09} & \textbf{75.78}
    & \textbf{69.20} & \textbf{74.64} \\

    \bottomrule
    \end{tabular}%
    }
    \caption{
    Reference-based evaluation of \ours{} on \benchmark{}.
    The best performance per dataset per metric is marked in boldface.
    }
    \label{tab:llm_judge_results}
\end{table*}

\subsection{Reference-Based Evaluation with an LLM Judge}
\label{subsec:llm_judge_evaluation}

To test whether the explanation-quality gains are specific to embedding-based evaluation, we complement the \emph{VisualSim} and \emph{ExplSim} metrics in Table~\ref{tab:main_results} with reference-based evaluation using Qwen3-32B~\cite{yang2025qwen3} as an LLM judge. \emph{VisualJudge} evaluates the predicted visual evidence description against its reference annotation, while \emph{ExplJudge} evaluates the predicted world-knowledge explanation against its reference.
Table~\ref{tab:llm_judge_results} shows that the advantage of \ours{} is preserved under LLM-based evaluation. 
On the in-distribution split, \ours{} achieves $81.02$ \emph{VisualJudge} and $85.29$ \emph{ExplJudge}, compared with $42.30$ and $41.42$ for the strongest baseline on these metrics. The same pattern holds out of distribution, where \ours{} achieves $69.20$ and $74.64$, respectively. These results show that the advantage of \ours{} in reference-based evaluation is consistent across embedding-based similarity and LLM-based scoring.

\begin{table}[!t]
\centering
\resizebox{0.9\columnwidth}{!}{%
\begin{tabular}{l rrr}
\toprule
\textbf{Model} & \textbf{$F_1^{\mathrm{inc}}$} & \textbf{$F_1^{\mathrm{con}}$} & \textbf{Macro-F1} \\
\midrule
CLIP            & 65.31 &  3.77 & 34.54 \\
GPT-5.4-mini    & 38.24 & 68.18 & 53.21 \\
Qwen2.5-VL-3B-Instruct   & 50.56 & 59.55 & 55.05 \\
Qwen3-VL-8B-Instruct     & 53.12 & 68.75 & 60.94 \\
InternVL3-8B    & 57.82 & 68.98 & 63.40 \\
\midrule
\ours{}         & \textbf{85.98} & \textbf{83.87} & \textbf{84.93} \\
\bottomrule
\end{tabular}}
\caption{Robustness to the AI-generation shortcut.}
\label{tab:ai_shortcut_control}
\end{table}

\newcommand{\consexample}[4]{%
\includegraphics[width=\linewidth]{figs/control_examples/#1.png} &
\includegraphics[width=\linewidth]{figs/control_examples/#1_consistent.png} &
#2 &
#3 &
#4
}

\begin{table*}[!t]
\centering
\small
\renewcommand{\arraystretch}{1.12}
\resizebox{\textwidth}{!}{%
\begin{tabular}{
>{\centering\arraybackslash}m{3.2cm}
>{\centering\arraybackslash}m{3.2cm}
>{\raggedright\arraybackslash}m{5.0cm}
>{\raggedright\arraybackslash}m{3.1cm}
>{\raggedright\arraybackslash}m{5.6cm}
}
\toprule
\textbf{Original Image} & \textbf{AI-Generated Consistent Image} & \textbf{Claim} & \textbf{Benign Addition} & \textbf{Why It Remains Consistent} \\
\midrule
\consexample{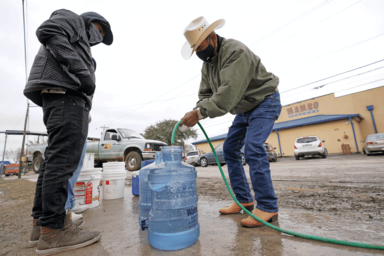}
{Gov.\ Greg Abbott of Texas said on Thursday that access to clean water remained a problem even as power was restored.}
{a weathered pickup truck}
{Pickup trucks are common in Texas, so the addition is plausible and does not alter the depicted event.}\\
\midrule
\consexample{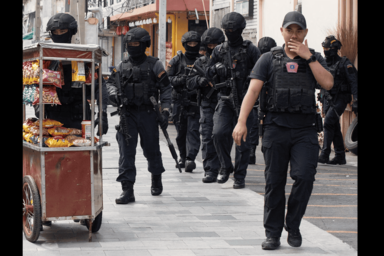}
{Police officers in tactical gear left the site of a shooting in Bangkok on Friday.}
{a street vendor cart with snacks}
{Street vendor carts are common in Bangkok and add a local element without changing the depicted event.}\\
\midrule
\consexample{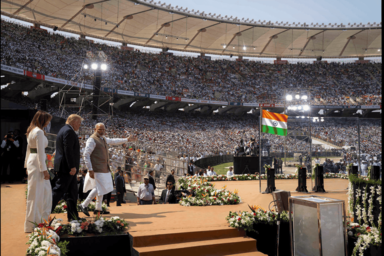}
{President Trump with Prime Minister Narendra Modi at Motera Stadium in Ahmedabad, India.}
{an Indian flag on a pole}
{An Indian flag is expected at an event held in India, introducing no contradiction with the claim.}\\
\midrule
\consexample{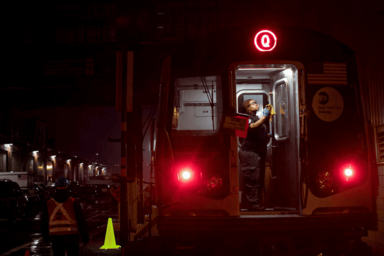}
{An M.T.A.\ worker disinfected a subway car at the Coney Island Yard in Brooklyn on Tuesday.}
{a small yellow traffic cone}
{Traffic cones are routine in a subway maintenance yard, matching the claimed context.}\\
\midrule
\consexample{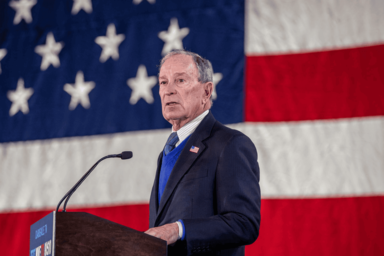}
{Michael R.\ Bloomberg speaking in Clarksville, Tenn., on Friday.}
{a small American flag pin}
{Flag pins are a common accessory for U.S.\ politicians at public speaking events.}\\
\midrule
\consexample{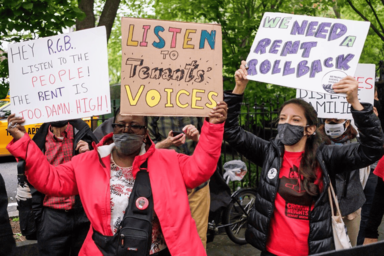}
{Activists gathered outside City Hall calling for an extension of the eviction moratorium in New York on Monday.}
{a yellow taxi cab}
{Yellow taxis are ubiquitous in New York City, so the addition fits the setting.}\\
\midrule
\consexample{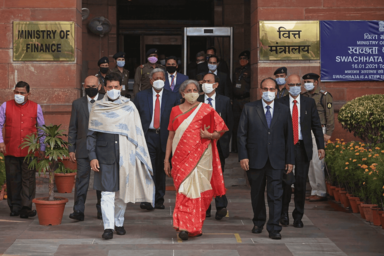}
{Finance Minister Nirmala Sitharaman of India before presenting the annual budget to Parliament in New Delhi.}
{a potted green plant}
{Potted plants are common decor in Indian public buildings and fit the Parliament setting.}\\
\bottomrule
\end{tabular}%
}
\caption{Representative AI-generated consistent examples used in the AI-generation shortcut control. Each edited image contains a benign addition that preserves consistency with the claim.}
\label{tab:ai_shortcut_examples}
\end{table*}

\rh{\begin{table}[!t]
\centering
\resizebox{0.99\columnwidth}{!}{%
\begin{tabular}{l rrrr}
\toprule
\textbf{Model} &
\textbf{Macro-F1} &
\textbf{TypeAcc} &
\textbf{VisualSim} &
\textbf{ExplSim} \\
\midrule
Qwen3-VL-4B-Instruct  & 65.20 & 43.00 & 39.90 & 43.50 \\
GPT-5.4-mini          & 50.70 &  9.00 &  9.70 & 11.60 \\
Qwen3-VL-8B-Instruct  & 54.00 & 54.00 & 50.90 & 54.80 \\
Qwen3-VL-8B-Thinking  & 68.50 & 21.00 & 17.50 & 18.80 \\
\midrule
\ours{}               & \textbf{95.50} & \textbf{65.00} &
                         \textbf{52.50} & \textbf{59.90} \\
\bottomrule
\end{tabular}%
}
\caption{Performance (in \%) on the two-inconsistency evaluation set. The best performance per metric is marked in boldface.}
\label{tab:multi_inconsistency}
\end{table}}

\begin{table}[!t]
\centering
\resizebox{0.9\columnwidth}{!}{%
\begin{tabular}{l rrrr}
\toprule
\textbf{Method}
& \textbf{Macro-F1}
& \textbf{TypeAcc}
& \textbf{VisualSim}
& \textbf{ExplSim} \\
\midrule
\multicolumn{5}{l}{\textit{/* In-Distribution Replacement */}} \\
Zero-shot CoT
& 58.90 & 19.06 & 23.16 & 25.24 \\
SFT
& 90.21 & 86.77 & 77.52 & 78.98 \\
SFT + DPO
& 75.08 & 84.64 & 80.40 & \textbf{83.91} \\
SFT + GRPO (\ours{})
& \textbf{94.88} & \textbf{89.76} & \textbf{83.04} & 82.54 \\
\midrule
\multicolumn{5}{l}{\textit{/* Out-of-Distribution Replacement */}} \\
Zero-shot CoT
& 59.44 & 19.63 & 24.18 & 27.18 \\
SFT
& 86.15 & 74.50 & 69.03 & 72.72 \\
SFT + DPO
& 74.09 & 75.07 & \textbf{72.89} & 73.89 \\
SFT + GRPO (\ours{})
& \textbf{90.26} & \textbf{78.65} & 72.09 & \textbf{75.78} \\
\bottomrule
\end{tabular}}
\caption{Comparison of training strategies. DPO and GRPO are initialized from the same SFT checkpoint.}
\label{tab:training_strategy_comparison}
\end{table}

\rh{\begin{table}[!t]
\centering
\resizebox{0.9\columnwidth}{!}{%
\begin{tabular}{l rrr}
\toprule
\textbf{Model}
& \textbf{$F_1^{\mathrm{con}}$}
& \textbf{$F_1^{\mathrm{inc}}$}
& \textbf{Macro-F1} \\
\midrule
Qwen2.5-VL-3B-Instruct &  7.70 & 67.60 & 37.60 \\
InternVL3-8B           & 51.20 & 66.10 & 58.70 \\
Qwen3-VL-4B-Instruct   & 79.10 & 82.60 & 80.80 \\
Qwen3-VL-8B-Instruct   & 83.90 & 86.00 & 84.90 \\
GPT-5.4-mini           & 93.50 & 92.50 & 93.00 \\
\midrule
\ours{}                 & \textbf{95.20} & \textbf{94.70} & \textbf{95.00} \\
\bottomrule
\end{tabular}}
\caption{Cross-benchmark consistency evaluation on a balanced, human-verified subset of MiRAGeNews. Results are reported in \%.}
\label{tab:external_benchmark_evaluation}
\end{table}}

\subsection{AI-Generation Shortcut Control}
\label{subsec:ai_generated_shortcut}
To test whether \ours{} relies on whether an image appears AI-generated rather than on claim-grounded inconsistencies, we construct a control set of 200 test claims approximately balanced across the nine inconsistency types.
For each claim, we use GPT-4o-mini to generate a benign, context-preserving edit prompt, and use FLUX.2-klein-9B to synthesize a claim-consistent AI-generated image, with examples shown in Table~\ref{tab:ai_shortcut_examples}.
We pair each AI-consistent image with an AI-generated inconsistent counterpart produced by the same generator for the same claim, so both classes contain AI-generated images.
As shown in Table~\ref{tab:ai_shortcut_control}, \ours{} achieves balanced performance, with $85.98\%$ $F_1^{\mathrm{inc}}$, $83.87\%$ $F_1^{\mathrm{con}}$, and $84.93\%$ Macro-F1.
In contrast, CLIP obtains only $3.77\%$ $F_1^{\mathrm{con}}$, suggesting that it largely predicts AI-generated images as \textsc{inconsistent} regardless of claim consistency. 
Overall, this control suggests that \ours{} is not driven solely by image provenance cues, but instead captures claim-grounded visual consistency.

\subsection{Multiple-Inconsistency Evaluation}
\label{subsec:multi_inconsistency_generalization}
To evaluate generalization beyond single-inconsistency training, we construct a test set of 100 claims, with 50 in-distribution and 50 out-of-distribution cases. Each claim is paired with an authentic image and a human-verified counterpart containing two inconsistencies, yielding 200 image--claim instances. 
For each two-inconsistency image, models are prompted to identify both inconsistencies and provide their corresponding types, visual evidence descriptions, and world-knowledge explanations in a single response. 
\emph{TypeAcc} requires both types to be correct, while \emph{VisualSim} and \emph{ExplSim} are averaged over the two predicted findings.
As shown in Table~\ref{tab:multi_inconsistency}, \ours{} performs best across all metrics. 
Despite being trained only on single-inconsistency examples, it achieves $95.50\%$ Macro-F1 and $65.00\%$ strict \emph{TypeAcc}, indicating that its verification ability transfers to multiple simultaneous inconsistencies.

\subsection{Training Strategy Comparison}
\label{subsec:training_strategy_comparison}
To evaluate the contribution of GRPO, we compare zero-shot CoT, SFT, SFT followed by DPO, and SFT followed by GRPO using the same Qwen3-VL-4B-Instruct backbone.
DPO and GRPO are initialized from the same SFT checkpoint, and the DPO preference pairs are constructed by ranking sampled responses using the same component-level reward as GRPO.
As shown in Table~\ref{tab:training_strategy_comparison}, SFT provides most of the task adaptation, while GRPO further improves SFT across all eight metrics, including gains of 4.67 and 4.11 Macro-F1 points on the in-distribution and out-of-distribution splits.
DPO improves selected semantic-similarity metrics but substantially reduces Macro-F1 on both splits.
Overall, these results support the use of GRPO for jointly improving consistency prediction and fine-grained explanation quality.

\subsection{Cross-Dataset Evaluation}
\label{subsec:cross_dataset_evaluation}
To investigate whether \ours{} transfers beyond \benchmark{}, we evaluate \ours{} and five baselines on a balanced, human-verified subset of MiRAGeNews~\cite{huang2024miragenews} without further training or adaptation.
We randomly sample candidate image--claim pairs from MiRAGeNews and manually screen them for claim consistency.
The resulting evaluation subset contains 50 generated image--claim pairs verified to exhibit a world-knowledge inconsistency, such as an incorrect national flag, a police uniform from the wrong country, or a public figure depicted with the wrong identity, and 50 consistent real-image pairs.
As shown in Table~\ref{tab:external_benchmark_evaluation}, \ours{} achieves $95.20\%$ $F_1^{\mathrm{con}}$, $94.70\%$ $F_1^{\mathrm{inc}}$, and $95.00\%$ Macro-F1, obtaining the highest Macro-F1 among the evaluated models.
These results provide evidence that \ours{} transfers to a task-relevant subset of an independently constructed dataset.

\begin{table}[!t]
\centering
\resizebox{0.99\columnwidth}{!}{%
\begin{tabular}{l rrrr}
\toprule
\textbf{Model}
& \textbf{Macro-F1}
& \textbf{TypeAcc}
& \textbf{VisualSim}
& \textbf{ExplSim} \\
\midrule
Qwen2.5-VL-3B-Instruct & 36.52 & 12.50 & 10.66 & 13.70 \\
LLaVA-OneVision-7B     & 53.43 & 13.75 & 24.07 & 27.93 \\
InternVL3-8B           & 60.03 & 31.75 & 23.31 & 27.67 \\
GPT-5.4-mini           & 59.94 & 29.00 & 21.53 & 23.67 \\
Qwen3-VL-8B-Instruct   & 63.42 & 38.25 & 32.33 & 34.48 \\
Qwen3-VL-4B-Instruct   & 59.53 & 36.00 & 24.01 & 29.80 \\
\midrule
\ours{}                 & \textbf{83.86} & \textbf{46.00}
                        & \textbf{63.91} & \textbf{68.59} \\
\bottomrule
\end{tabular}}
\caption{Performance (in \%) on the held-out flag and signage types.
The best performance per metric is marked in boldface.}
\label{tab:held_out_type_evaluation}
\end{table}

\subsection{Held-Out Inconsistency Type Evaluation}
\label{subsec:held_out_type_evaluation}
To evaluate whether \ours{} generalizes to inconsistency types absent from task-specific training, we train a variant of \ours{} on seven of the nine categories, excluding all flag and signage examples, and evaluate it on these two held-out types.
As shown in Table~\ref{tab:held_out_type_evaluation}, \ours{} outperforms all baselines across all metrics, achieving $83.86\%$ Macro-F1, $46.00\%$ \emph{TypeAcc}, $63.91\%$ \emph{VisualSim}, and $68.59\%$ \emph{ExplSim}.
These results indicate that its verification ability transfers to inconsistency types not observed during task-specific training.

\section{Error Examples}
\label{sec:error_examples}

Table~\ref{tab:failure_examples} presents representative failure cases of \ours{} for qualitative analysis. These examples complement the aggregate error statistics by illustrating different failure modes encountered during inconsistency detection and explanation generation. In some cases, \ours{} fails to identify the manipulated visual evidence altogether. In others, the model predicts the correct verdict but justifies it using hallucinated or unedited visual features. We also observe cases where the model attends to the manipulated region but assigns it the wrong real-world identity or contextual origin.
For each example, we provide the ai-generated image, the associated claim, the model output, and a brief diagnosis explaining why the given image and the claim are inconsistent.

\begin{table*}[!t]
\centering
\small
\renewcommand{\arraystretch}{1.15}

\resizebox{\textwidth}{!}{%
\begin{tabular}{
m{2.2cm}
m{4.6cm}
m{3.6cm}
m{4.4cm}
m{5.4cm}
}

\toprule
\textbf{Failure Mode}
&
\textbf{AI-Generated Image}
&
\textbf{Claim}
&
\textbf{MIC Output}
&
\textbf{Why This Is Incorrect}
\\
\midrule


Missed cultural edit
&
\includegraphics[width=4.8cm]{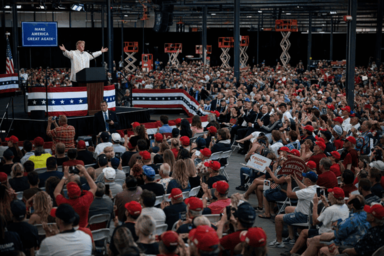}
&
President Trump defied the governor of Nevada by holding an indoor rally near Las Vegas.
&
\texttt{<verdict>}: CONSISTENT \newline
\texttt{<type>}: None \newline
\texttt{<visual>}: None \newline
\texttt{<explanation>}: None
&
The candidate's dark suit has been replaced by a Filipino \emph{barong Tagalog}, an attire that does not fit a U.S.\ campaign rally. MIC fails to register the cultural mismatch and incorrectly predicts \texttt{CONSISTENT}.
\\

\midrule


Missed object or scene edit
&
\includegraphics[width=4.8cm]{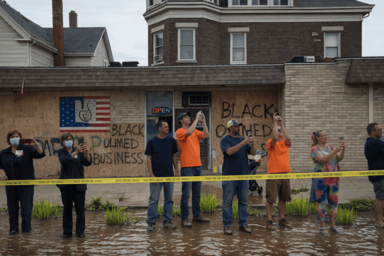}
&
Residents of Kenosha, Wis., watched President Trump's motorcade this month.
&
\texttt{<verdict>}: CONSISTENT \newline
\texttt{<type>}: None \newline
\texttt{<visual>}: None \newline
\texttt{<explanation>}: None
&
The sidewalk in a Wisconsin street scene has been replaced with shallow muddy water and rice-paddy seedlings, an environmental cue that does not fit an urban Midwestern setting. MIC overlooks the scene-level edit entirely and predicts \texttt{CONSISTENT}.
\\

\midrule


Evidence describes feature not in image
&
\includegraphics[width=4.8cm]{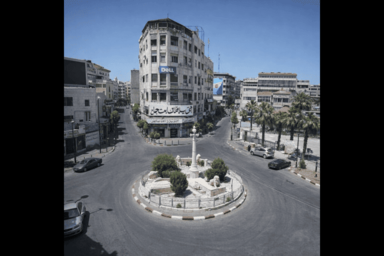}
&
Activists were arrested for planning an anti-corruption demonstration in Manara Square, Ramallah.
&
\texttt{<verdict>}: INCONSISTENT \newline
\texttt{<type>}: branding \newline
\texttt{<visual>}: Dell-branded storefront/logo \newline
\texttt{<explanation>}: A Dell storefront logo introduces a distinctly American consumer-tech brand into a Palestinian protest scene, creating a cultural mismatch.
&
MIC reaches the correct verdict but justifies it using hallucinated evidence: the cited Dell storefront does not exist in the image. The actual edit, Arabic storefront signs replaced by Urdu Nastaliq script, is never identified.
\\

\midrule


Evidence describes unedited element
&
\includegraphics[width=4.8cm]{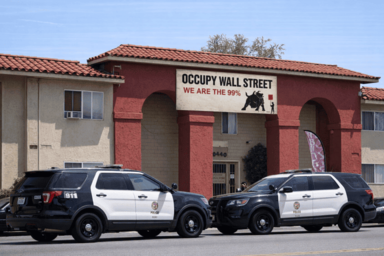}
&
Firefighters pronounced three children dead at an apartment complex in Los Angeles.
&
\texttt{<verdict>}: INCONSISTENT \newline
\texttt{<type>}: architecture \newline
\texttt{<visual>}: Spanish colonial-style arched stucco facade with terracotta tile roofline \newline
\texttt{<explanation>}: The edited architecture suggests a Latin American Spanish colonial setting rather than a typical California urban residential building.
&
MIC bases its verdict on an unedited architectural feature already present in the original image. The actual edit, an apartment leasing sign replaced by an Occupy Wall Street banner, is missed.
\\

\midrule


Wrong country/brand attribution
&
\includegraphics[width=4.8cm]{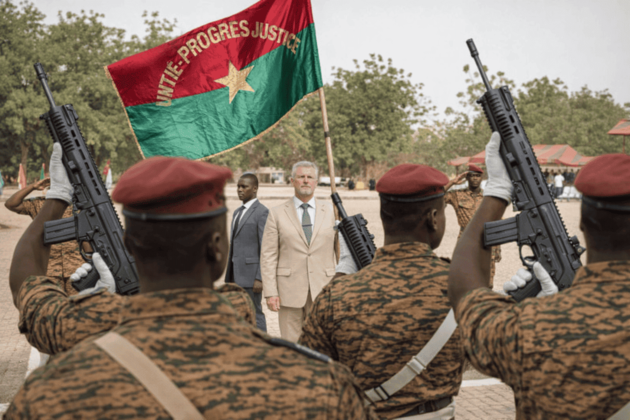}
&
Andrew Young, the American ambassador to Burkina Faso, during the opening ceremony of a joint military exercise.
&
\texttt{<verdict>}: INCONSISTENT \newline
\texttt{<type>}: technology \newline
\texttt{<visual>}: Russian AK-12 assault rifles \newline
\texttt{<explanation>}: Equipping the ceremonial honor guard with Russian-service AK-12 rifles shifts the implied military equipment and alliance context away from the Burkinab\'e setting in the caption.
&
MIC attends to the correct edited object, the ceremonial rifles, and reaches the correct verdict, but misattributes their identity: the actual edit introduces Italian Beretta~ARX160 rifles, not Russian AK-12s. The verdict and visual region are correct, but the entity-to-country link fails.
\\

\bottomrule
\end{tabular}%
}

\caption{
Representative MIC failure cases across different error modes.
MIC may miss edited evidence, hallucinate unsupported evidence, attend to unedited regions, or misattribute edited entities.
}

\label{tab:failure_examples}

\end{table*}

\section{Recent News Examples}
\label{sec:recent_news_examples}

Table~\ref{tab:recent_news_examples} presents representative examples from the recent-news benchmark introduced in Section~\ref{subsec:recent_news_generalization}. The AI-generated images are created using GPT-Image-1.5. 
Each example includes the original news image, the edited counterpart, the associated claim, the manipulated visual element, and the world-knowledge explanation. These examples illustrate temporally grounded and context-dependent inconsistencies in recent news events, where successful verification requires up-to-date geopolitical, cultural, or event-specific knowledge.

\begin{table*}[!t]
\centering
\small
\renewcommand{\arraystretch}{1.15}

\resizebox{\textwidth}{!}{%
\begin{tabular}{
>{\raggedright\arraybackslash}m{2.1cm}
>{\centering\arraybackslash}m{3.1cm}
>{\centering\arraybackslash}m{3.1cm}
>{\raggedright\arraybackslash}m{4.7cm}
>{\raggedright\arraybackslash}m{4.7cm}
>{\raggedright\arraybackslash}m{5.6cm}
}

\toprule
\textbf{Inconsistency Type}
&
\textbf{Original Image}
&
\textbf{Edited Image}
&
\textbf{Claim}
&
\textbf{What Changed}
&
\textbf{Why Contradicts}
\\
\midrule
clothing
&
\includegraphics[width=\linewidth]{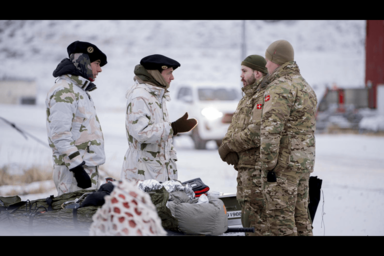}
&
\includegraphics[width=\linewidth]{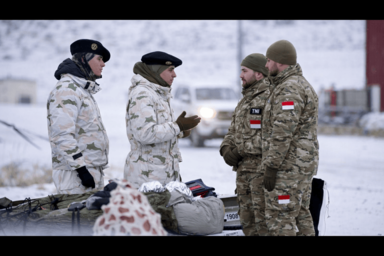}
&
Meeting of French mountain troops and Danish soldiers as part of the Arctic endurance mission.
&
Danish-style red patch with white cross $\rightarrow$ Indonesian TNI red-over-white flag patches and ``TNI'' identifiers.
&
The caption frames the scene as a meeting between French and Danish forces. Replacing the visible uniform identifiers with Indonesian TNI patches changes the represented military affiliation.
\\

\midrule

flag
&
\includegraphics[width=\linewidth]{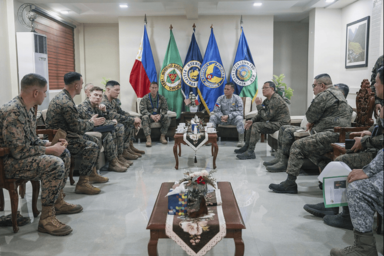}
&
\includegraphics[width=\linewidth]{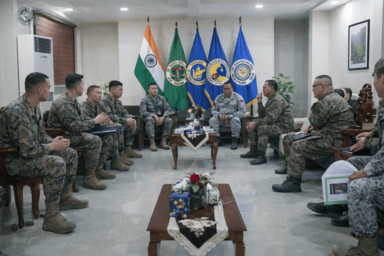}
&
U.S. Marine Corps Maj. Sean Gunn speaks to Philippine Air Force leadership.
&
Philippine flag $\rightarrow$ Indian tricolor flag with navy Ashoka Chakra.
&
The claim anchors the scene in a Philippine Air Force meeting. Replacing the Philippine flag with India's national flag misidentifies the host-country context.
\\

\midrule

gesture
&
\includegraphics[width=\linewidth]{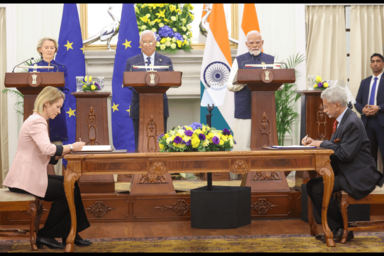}
&
\includegraphics[width=\linewidth]{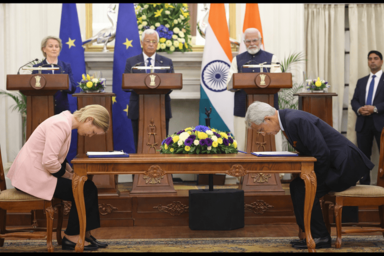}
&
Kaja Kallas and S. Jaishankar signing a Security and Defence Partnership in New Delhi.
&
Two officials signing documents $\rightarrow$ two officials performing a mutual Japanese deep bow while seated.
&
The claim describes a diplomatic signing ceremony in India. The edited body language changes the action into a Japanese-style bowing greeting, so the depicted interaction no longer matches the claimed event.
\\

\midrule

signage
&
\includegraphics[width=\linewidth]{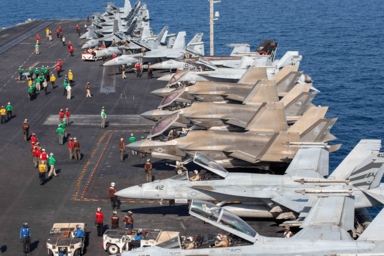}
&
\includegraphics[width=\linewidth]{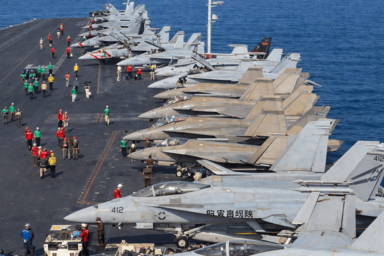}
&
Aircraft attached to Carrier Air Wing (CVW) 9 sit on the flight deck of Nimitz-class aircraft carrier USS Abraham Lincoln (CVN 72) in support of Operation Epic Fury.
&
Aircraft fuselage text ``VFA-41'' $\rightarrow$ Japanese Air Self-Defense Force text.
&
The claim describes U.S.\ Navy aircraft aboard the USS Abraham Lincoln. The edited Japanese military marking assigns the aircraft to a different national force, creating a direct mismatch with the carrier-air-wing context.
\\

\midrule

architecture
&
\includegraphics[width=\linewidth]{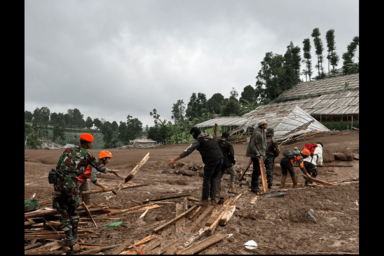}
&
\includegraphics[width=\linewidth]{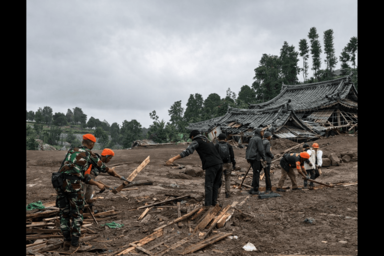}
&
Korpaskhas evacuates victims of the Cisarua landslide in West Bandung.
&
Damaged slanted greenhouse roofs $\rightarrow$ damaged Japanese shrine/temple-style curved kawara tile roofs.
&
The claim places the landslide site in West Bandung, Indonesia, but the edited roof forms suggest Japanese religious or traditional architecture rather than local agricultural structures.
\\

\midrule

infrastructure
&
\includegraphics[width=\linewidth]{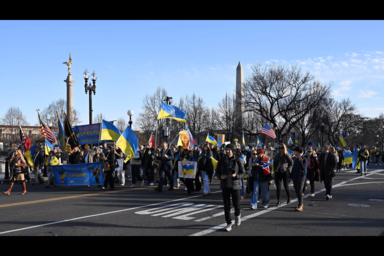}
&
\includegraphics[width=\linewidth]{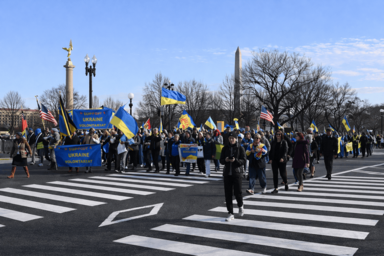}
&
The front of the march in support of Ukraine with flags and banners; behind the trees is the Washington Monument on 17th Street NW, Washington, DC.
&
Washington, DC road markings $\rightarrow$ Japanese Tokyo-style zebra crosswalk and diamond warning marking.
&
The claim identifies the scene as 17th Street NW in Washington, DC, while the edited road markings are characteristic of Japanese urban traffic infrastructure.
\\

\midrule

technology
&
\includegraphics[width=\linewidth]{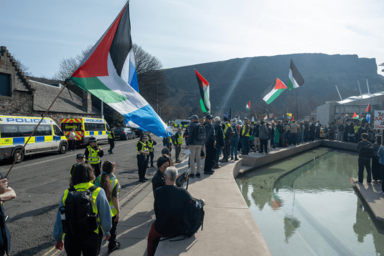}
&
\includegraphics[width=\linewidth]{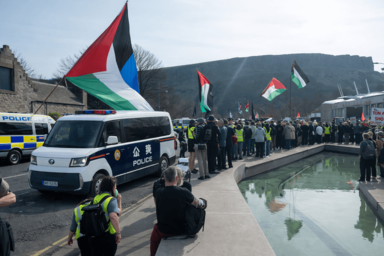}
&
On 21 March 2026, a protest was held outside the Scottish Parliament against immigration to the country.
&
UK police van $\rightarrow$ Chinese public-security police van with Chinese markings.
&
The claim places the protest outside the Scottish Parliament, where policing would be associated with Scottish or UK authorities. The edited Chinese police van introduces the wrong national security context.
\\

\midrule

branding
&
\includegraphics[width=\linewidth]{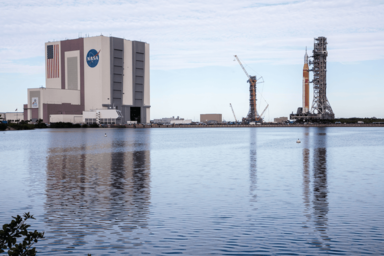}
&
\includegraphics[width=\linewidth]{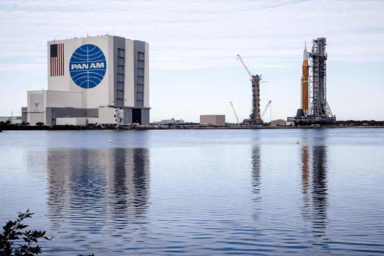}
&
The Artemis II stack being rolled out of the VAB before its February launch attempt.
&
NASA logo on the Vehicle Assembly Building $\rightarrow$ vintage Pan Am globe logo.
&
The caption refers to a contemporary NASA Artemis II rollout at the Vehicle Assembly Building. The edited Pan Am branding introduces a defunct airline identity in place of the expected NASA visual context.
\\

\midrule

environment
&
\includegraphics[width=\linewidth]{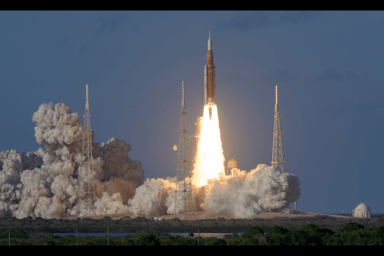}
&
\includegraphics[width=\linewidth]{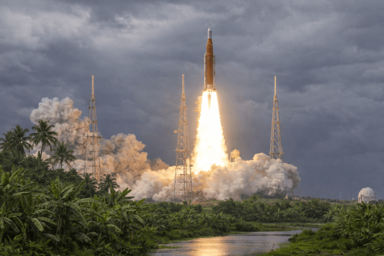}
&
NASA's Artemis II mission launched four astronauts aboard the SLS rocket and Orion spacecraft from Kennedy Space Center on April 1, 2026.
&
Kennedy Space Center coastal vegetation $\rightarrow$ tropical rainforest vegetation with banana plants and coconut palms.
&
The claim places the launch at Kennedy Space Center in Florida, but the edited vegetation suggests a Southeast Asian tropical rainforest rather than the Florida Space Coast environment.
\\

\bottomrule
\end{tabular}%
}

\caption{
Representative edited examples from recent real-world news images from 2026.
}
\label{tab:recent_news_examples}
\end{table*}


\section{Prompts}
\label{appendix:prompts}

This section presents the prompt templates used during the construction of \benchmark{}. The prompts correspond to the automated stages described in Section~\ref{subsec:dataset_construction_pipeline}.

\FloatBarrier
\begin{tcolorbox}[
enhanced,
adjusted title,
breakable,
float*=htbp,
width=\textwidth,
title={\textsf{Prompt Template for Sample Screening for Editing}},
fonttitle=\small\bfseries\sffamily,
fontupper=\small\sffamily,
boxsep=4pt,
left=6pt,
right=6pt,
top=2pt,
bottom=3.5pt,
arc=1.5pt,
drop shadow=black!25,
colback=gray!1!white,
colframe=blue!30!black,
coltitle=black,
colbacktitle=blue!7!white,
overlay={
  \node[fill=white, draw=gray!40, rounded corners=1pt, inner sep=1pt]
  at (frame.south east) [anchor=south east, xshift=-3pt, yshift=2.5pt] {};
},
]

\begin{lstlisting}[basicstyle=\ttfamily\small, breaklines=true, columns=fullflexible]
You are helping to build a dataset for studying image-caption inconsistency detection.

Given a NEWS IMAGE and its associated METADATA, determine:
1. Which inconsistency types are FEASIBLE to create by editing this image
2. Whether this image-caption pair is SUITABLE overall for the task

### Metadata ###
- Caption: {caption}
- Headline: {headline}
- Location: {location}
- Time: {time}
- Keywords: {keywords}

### Inconsistency types ###
(Mark as feasible ONLY if the visual element is clearly visible in the image AND the caption provides a factual anchor to contradict.)

- clothing: {type_descs[clothing]}
- flag: {type_descs[flag]}
- gesture: {type_descs[gesture]}
- signage: {type_descs[signage]}
- architecture: {type_descs[architecture]}
- infrastructure: {type_descs[infrastructure]}
- technology: {type_descs[technology]}
- branding: {type_descs[branding]}
- environment: {type_descs[environment]}

### Rules ###
- An inconsistency type is feasible ONLY if BOTH conditions are met:
  (a) The image contains the relevant visual element (e.g., a flag is visible).
  (b) The caption/metadata provides enough context to create a meaningful inconsistency (e.g., caption says "Australian", so changing to a German flag creates an inconsistency).
- Mark "suitable" as true only if at least 1 type is feasible.
- Be strict: if you are not sure a visual element is present, mark it as not feasible.

### Output (JSON only) ###
{
    "suitable": true/false,
    "feasible_types": ["type1", "type2"],
    "visual_elements": "Brief description of key visual elements in the image (1-2 sentences).",
    "reasoning": "Brief explanation of why the feasible types were selected (1-2 sentences)."
}
\end{lstlisting}
\end{tcolorbox}

\begin{tcolorbox}[
enhanced,
adjusted title,
breakable,
float*=htbp,
width=\textwidth,
title={\textsf{Prompt Template for Visual Element Visibility Check}},
fonttitle=\small\bfseries\sffamily,
fontupper=\small\sffamily,
boxsep=4pt,
left=6pt,
right=6pt,
top=2pt,
bottom=3.5pt,
arc=1.5pt,
drop shadow=black!25,
colback=gray!1!white,
colframe=blue!30!black,
coltitle=black,
colbacktitle=blue!7!white,
overlay={
  \node[fill=white, draw=gray!40, rounded corners=1pt, inner sep=1pt]
  at (frame.south east) [anchor=south east, xshift=-3pt, yshift=2.5pt] {};
},
]

\begin{lstlisting}[basicstyle=\ttfamily\small, breaklines=true, columns=fullflexible]
You are a strict visual inspector. Look at this NEWS IMAGE carefully.

### Task ###
For each inconsistency type below, determine if the required visual element is CLEARLY VISIBLE in the image.

CRITICAL RULE: You must ONLY mark a type as feasible if you can literally SEE the element in the image. Do NOT infer or guess based on the caption. If you cannot clearly see it, mark it as NOT feasible.

### Caption context (for reference only -- do NOT use this to guess what is in the image) ###
- Caption: {caption}
- Location: {location}
- Time: {time}

### Types to verify -- for each, state what you SEE ###

1. clothing: Can you clearly see identifiable uniforms (color, badges, insignia, patches) OR traditional/cultural clothing (sari, thobe, kimono, chador, kilt, etc.)?
2. flag: Can you clearly see a flag or national banner? Describe its colors/pattern.
3. gesture: Can you clearly see a specific social interaction or gesture (handshake, salute, protest gesture, ceremony, greeting ritual)?
4. signage: Can you clearly READ text on signs/banners/placards/nameplates? What does it say, and in what language?
5. architecture: Can you clearly see distinctive architectural features (dome shapes, minaret towers, pagoda roofs, Gothic arches, colonial facades, traditional house styles)?
6. infrastructure: Can you clearly see transportation infrastructure, road signs, traffic systems, rail stations, or utility infrastructure with country-specific features?
7. technology: Can you clearly see identifiable vehicles, aircraft, weapons, or devices? What specific type?
8. branding: Can you clearly see advertisements, brand logos, event marks, campaign materials, or product displays?
9. environment: Can you clearly see natural environment indicators (vegetation type, terrain, climate markers like snow/desert/tropical foliage, seasonal cues)?

### Quality rating ###
Rate the overall suitability of this image for creating contextual inconsistencies:
- high: Multiple clearly visible elements, sharp image, specific caption with verifiable facts.
- medium: At least one clearly visible element, decent image quality.
- low: Elements are small/blurry/ambiguous, or caption is too vague.

### Output (JSON only, no other text) ###
{
    "quality": "high/medium/low",
    "visual_description": "A thorough description of everything you see in the image. Describe the scene, people (appearance, clothing, posture, expressions), objects, buildings, signs, vehicles, natural environment, lighting, and any other notable visual details. Be as detailed as possible. Do NOT copy the caption -- describe only what is VISIBLE.",
    "feasible_types": {
        "clothing":         {"visible": true/false, "evidence": "what exactly you see, or 'not visible'"},
        "flag":             {"visible": true/false, "evidence": "..."},
        "gesture":          {"visible": true/false, "evidence": "..."},
        "signage":          {"visible": true/false, "evidence": "..."},
        "architecture":     {"visible": true/false, "evidence": "..."},
        "infrastructure":   {"visible": true/false, "evidence": "..."},
        "technology":       {"visible": true/false, "evidence": "..."},
        "branding":         {"visible": true/false, "evidence": "..."},
        "environment":      {"visible": true/false, "evidence": "..."}
    }
}
\end{lstlisting}
\end{tcolorbox}

\begin{tcolorbox}[
enhanced,
adjusted title,
breakable,
float*=htbp,
width=\textwidth,
title={\textsf{Prompt Template for Edit Prompt Generation}},
fonttitle=\small\bfseries\sffamily,
fontupper=\small\sffamily,
boxsep=4pt,
left=6pt,
right=6pt,
top=2pt,
bottom=3.5pt,
arc=1.5pt,
drop shadow=black!25,
colback=gray!1!white,
colframe=blue!30!black,
coltitle=black,
colbacktitle=blue!7!white,
overlay={
  \node[fill=white, draw=gray!40, rounded corners=1pt, inner sep=1pt]
  at (frame.south east) [anchor=south east, xshift=-3pt, yshift=2.5pt] {};
},
]

\begin{lstlisting}[basicstyle=\ttfamily\small, breaklines=true, columns=fullflexible]
You are a researcher creating a dataset to study image-caption inconsistency detection.

### Goal ###
Generate a DETAILED image editing instruction that modifies a specific visual element so that:
1. The edited image still looks REALISTIC and NATURAL on its own -- it should not look obviously fake or absurd.
2. But the edited element is INCONSISTENT with the caption's context -- the inconsistency can only be detected by someone who has world knowledge about the event, location, culture, or time period described in the caption.
3. The rest of the image must remain UNCHANGED -- only modify the target element.

### News context ###
- Caption: {caption}
- Headline: {headline}
- Keywords: {keywords}

### Detailed visual description of the image ###
{visual_description}

### Target element to modify ###
- Type: {edit_type}
- What is currently visible: {evidence}

### How to create the inconsistency ###
{inconsistency_def}

### Key Principles ###
1. Contextual relevance: The replacement element must be chosen based on the specific news story. Ask yourself: "What would be a PLAUSIBLE but WRONG alternative in this context?" Pick something geopolitically, historically, or culturally related -- not random.
2. Visual realism: The edit should look natural. The replacement element should fit the scene's lighting, style, and composition.
3. Subtle inconsistency: The inconsistency should require WORLD KNOWLEDGE to detect. Someone unfamiliar with the context might not notice anything wrong.
4. Extreme specificity: Your editing prompt must be so detailed that an image editor can follow it WITHOUT seeing the original image. Include:
   - For signage: the EXACT replacement text string AND its language.
   - For flag: the EXACT flag design.
   - For clothing: the EXACT uniform colors, badge design, insignia placement, OR the EXACT garment name, fabric pattern, colors, and how it is worn.
   - For technology: the EXACT make, model, color, and distinguishing features of the replacement.
   - For gesture: the EXACT gesture, posture, and body language to change to.
   - For branding: the EXACT brand name, logo design, colors, event mark, or product shown.
   - For architecture: the EXACT architectural element to modify (dome shape, arch style, decorative pattern, roof type) and what to replace it with.
   - For infrastructure: the EXACT sign/marking/station feature to modify and the country-specific replacement style.
   - For environment: the EXACT vegetation/terrain/climate element to modify and what climate zone to shift it toward.
5. Preserve everything else: Only change the target element. Do NOT alter the background, other people, lighting, or composition.

### Contextual inconsistency ###
The edit must create a meaningful inconsistency with the caption that requires WORLD KNOWLEDGE to detect:
- Ask yourself: "What would be PLAUSIBLE but WRONG in this specific context?"
- The replacement should be from a country/culture/era that is RELATED to the story but INCORRECT.
- The edited image should look perfectly normal on its own -- only someone who reads the caption and has relevant knowledge would notice the inconsistency.
- Avoid extremes: not so obvious it is immediately spotted, not so obscure nobody would notice.

### Mandatory region constraint ###
For this specific edit, you MUST choose your replacement from the following region:
{region_hint}

Find the most contextually relevant replacement FROM THIS REGION that creates a meaningful inconsistency with the caption. If this region has no direct connection to the story, find an INDIRECT but still meaningful connection (e.g., same political bloc, similar conflict, historical parallel, trade relationship).

### Examples of GOOD edits (contextually meaningful + diverse) ###
- Caption about Australian military -> New Zealand DPCU pattern with silver fern badge (close ally, similar but distinct) [clothing]
- Caption about Greek austerity protests -> Spanish red-yellow flags (both EU debt crisis countries) [flag]
- Caption about a diplomatic signing ceremony -> Japanese deep bow posture (changes the social action) [gesture]
- Caption about Iranian nuclear talks -> Arabic text in Naskh script (neighboring language, shifts location) [signage]
- Caption about Istanbul summit -> Persian-style pointed onion dome with blue mosaic tilework (shifts Turkey -> Iran) [architecture]
- Caption about Mumbai traffic -> left-hand-drive road with Autobahn-style blue signs (shifts India -> Germany) [infrastructure]
- Caption about Pakistani military -> Indian BMP-2 IFVs with Indian Army markings (rival neighbor) [technology]
- Caption about 2006 Iraq War -> Blockbuster Video rental ad (Blockbuster was declining by 2006 -- temporal marker) [branding]
- Caption about Norwegian fjord -> Mediterranean olive grove with dry golden grass and cypress trees (shifts Nordic -> Southern Europe) [environment]

### Examples of BAD edits (avoid these) ###
- Unrelated changes: Mozambique flag in a European story (no narrative connection).
- Visually absurd: Hawaiian shirts at a funeral.
- Vague: "change text to Japanese" without specifying EXACT text string.
- Generic: "change to a different uniform" without colors, badges, insignia.
- Always picking the same replacement (e.g., always Russia for flags, always Japanese for clothing).

### Output (respond in JSON only, no other text) ###
{
    "editing_prompt": "A hyper-specific instruction for image editing. Start with 'Edit the image to replace ONLY...' and describe EXACTLY what to change: specific colors, exact text strings in foreign scripts, precise insignia designs, specific garment names. The instruction must be detailed enough for someone who has never seen the image. MUST end with 'Keep all other elements unchanged.'",
    "what_changed": "Brief: 'Original element -> Replacement element'.",
    "why_contradicts": "One sentence: why does this replacement DIRECTLY CONTRADICT the caption's narrative?",
    "world_knowledge_needed": "One sentence: what specific world knowledge is needed to detect this inconsistency?",
    "difficulty": "easy/medium/hard -- how difficult is it for a human to detect this inconsistency without reading the caption."
}
\end{lstlisting}
\end{tcolorbox}

\end{document}